%% file: main.tex
\pdfoutput=1

\documentclass[11pt]{article}

\usepackage[final]{acl}

\usepackage{times}
\usepackage{latexsym}
\usepackage{amsmath,amsfonts}
\usepackage{array}
\usepackage{graphicx}
\usepackage{multirow}
\usepackage{floatrow}
\usepackage{multicol}
\usepackage{booktabs}
\usepackage{multirow}
\usepackage[normalem]{ulem} 
\useunder{\uline}{\ul}{}
\newcommand{\ie}{\emph{i.e., }}
\newcommand{\eg}{\emph{e.g., }}

\newcommand{\sectionref}[1]{\S~\ref{#1}}
\usepackage[T1]{fontenc}

\usepackage[utf8]{inputenc}

\usepackage{microtype}

\usepackage{inconsolata}
\usepackage{caption}
\usepackage{graphicx}
\usepackage{float} 
\usepackage{subcaption}
\usepackage{svg}
\usepackage{makecell}

\title{PunchBench: Benchmarking MLLMs in Multimodal Punchline Comprehension}

\author{\textbf{Kun Ouyang}$^\dag$$^\ddag$, \textbf{Yuanxin Liu}$^\dag$, \textbf{Shicheng Li}$^\dag$,\\ \textbf{Yi Liu}$^\dag$, \textbf{Hao Zhou}$^\ddag$, \textbf{Fandong Meng}$^\ddag$, \textbf{Jie Zhou}$^\ddag$, \textbf{Xu Sun}$^\dag$\thanks{~~Xu Sun is the corresponding author.}
\\   $^\dag$ State Key Laboratory of Multimedia Information Processing,\\School of Computer Science, Peking University \\$^\ddag$ WeChat AI, Tencent Inc., China\\
kunouyang10@gmail.com, liuyuanxin@stu.pku.edu.cn, \{lisc99, imliuyi \}@pku.edu.cn, \\       \{tuxzhou, fandongmeng, withtomzhou\}@tencent.com, xusun@pku.edu.cn\\}

\begin{document}
\maketitle
\begin{abstract}
Multimodal punchlines, which involve humor or sarcasm conveyed in image-caption pairs, are a popular way of communication on online multimedia platforms. With the rapid development of multimodal large language models (MLLMs), it is essential to assess their ability to effectively comprehend these punchlines. However, existing benchmarks on punchline comprehension suffer from three major limitations: 1) language shortcuts that allow models to solely rely on text, 2) lack of question diversity, and 3) narrow focus on a specific domain of multimodal content (e.g., cartoon).
To address these limitations, we introduce a multimodal \textbf{Punch}line comprehension \textbf{Bench}mark, named \textbf{PunchBench}, which is tailored for accurate and comprehensive evaluation of punchline comprehension. 
To enhance the evaluation accuracy, we generate synonymous and antonymous captions by modifying original captions, which mitigates the impact of shortcuts in the captions. 
To provide a comprehensive evaluation, PunchBench incorporates diverse question formats and image-captions from various domains.
On this basis, we conduct extensive evaluations and reveal a significant gap between state-of-the-art MLLMs and humans in punchline comprehension. 
To improve punchline comprehension, we propose Simple-to-Complex Chain-of-Question (SC-CoQ) strategy, enabling the models to incrementally address complicated questions by first mastering simple ones. SC-CoQ effectively enhances the performance of various MLLMs on PunchBench, surpassing in-context learning and chain-of-thought. Datasets, codes are publicly available at \url{https://github.com/OuyangKun10/PunchBench}.
\end{abstract}

\input{Chapter/1_introduction}
\input{Chapter/2_related}
\input{Chapter/3_dataset_task}
\input{Chapter/4_evaluation}

\input{Chapter/5_conclusion}
\section*{Limitations}
In this work, we focus on multimodal punchline comprehension for the image-caption pairs, which only consist of static content. According to the evaluation results, MLLMs struggle with the punchline comprehension and fall behind humans.
Extending this challenge to videos, where punchlines are often embedded in dynamic flows of information, poses even greater complexity. Unlike static images, videos require models to process temporal dynamics and integrate contextual cues across frames, demanding more advanced comprehension capabilities.
Given the added challenges of punchline comprehension in video content, such as comedy, this area presents a meaningful avenue for further exploration. In future work, we aim to evaluate MLLMs' ability to understand punchlines within videos, advancing their capability to process and interpret dynamic multimodal content.
\section*{Acknowledgements}
We thank all the anonymous reviewers for their constructive comments.
This research was partially supported by the National Natural Science Foundation of China under Grant No. 92470205 and No. 62176002.
Xu Sun is the corresponding author of this paper.

\bibliography{reference}

\input{Chapter/6_appendix}

\end{document}

%% file: Chapter/1_introduction.tex
\section{Introduction}
\begin{figure}
    \centering
    \includegraphics[width=\textwidth]{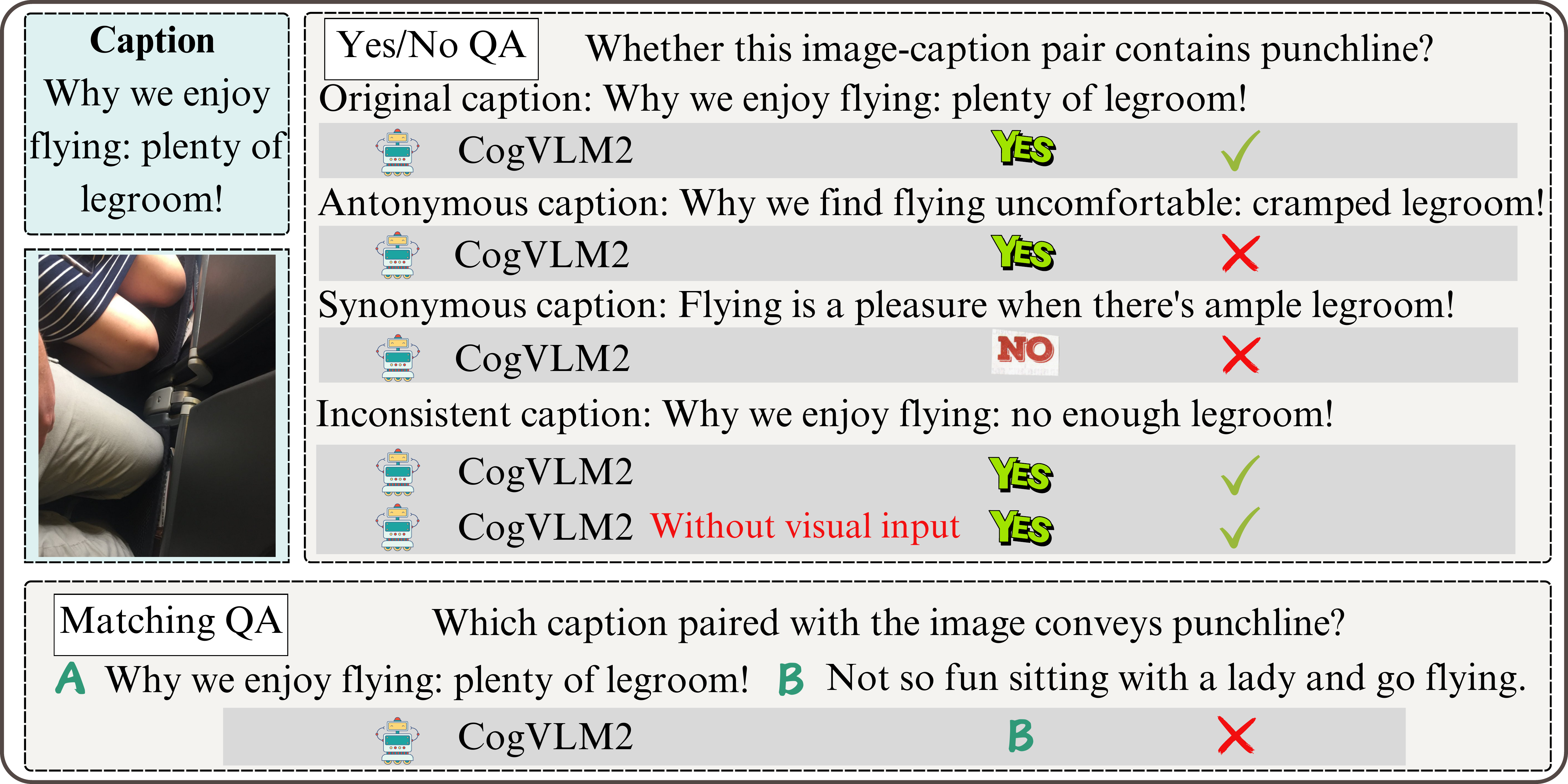}
    \caption{An example of multimodal punchline comprehension. We illustrate the response of CogVLM2 when provided with different captions and question formats.}

\label{Fig:Intro}
\end{figure}
Recent research on Multimodal Large Language Models (MLLMs)~\cite{Qwen2VL,GPT-4o} has made rapid progress in vision-language tasks such as visual question answering~\cite{vqaiccv}, dense image captioning~\cite{DBLP:conf/cvpr/JohnsonKF16} and optical character recognition~\cite{Islam2017ASO}. Despite the advanced capabilities of modern MLLMs in comprehending factual information from visual content, whether they can effectively grasp punchlines within the multimodal context remains an open question.

As illustrated in Figure \ref{Fig:Intro}, multimodal punchlines are typically presented as image-caption pairs \cite{DBLP:conf/acl/CaiCW19}, where humor or sarcasm is elicited through a striking contrast or alignment between visual and textual elements. Understanding these punchlines is important yet challenging for the development of MLLMs. \textbf{On the one hand}, multimodal punchlines are an essential way of communication on online multimedia platforms. Improving comprehension of punchlines is crucial for many real-world applications, including Human-AI interaction \cite{DBLP:conf/hci/HempelmannP15} and sentiment analysis \cite{DBLP:conf/wanlp/MahdaouyMEMBK21}. \textbf{On the other hand}, unlike conventional visual question answering and captioning tasks, multimodal punchline understanding necessitates a nuanced perception of visual content, a strong grasp of language prior knowledge, as well as a deep understanding of the interplay between visual and textual information \cite{DBLP:conf/acl/JingSOJN23}.

There are some prior studies on multimodal punchline comprehension, attempting to evaluate sarcasm explanation~\cite{DBLP:conf/aaai/Desai0A22} and humor comprehension~\cite{DBLP:conf/acl/HesselMHLDZM023}, respectively. However, despite the valuable benchmarks presented by these studies, they suffer from three major limitations that hinder an accurate and comprehensive assessment of multimodal punchline comprehension. \textbf{First, existing benchmarks overlook the potential shortcuts in the captions.} As shown in the \textit{Yes/No QA} task from Figure~\ref{Fig:Intro}, CogVLM2~\cite{hong2024cogvlm2} can correctly identify that the original caption conveys a punchline regarding the image but fails when some words in the original caption are replaced with antonymous or synonymous ones. Additionally, the model can correctly answer \textit{Yes/No QA} solely based on an inconsistent caption without visual input. This suggests that the model may exploit biased words (e.g., "enjoy," "plenty of") or text-only inconsistencies (e.g., "enjoy flying" versus "not enough legroom") to arrive at the correct answer rather than genuinely understanding the multimodal punchline. \textbf{Second, most previous benchmarks are constrained to a single question format}~\cite{DBLP:conf/acl/CaiCW19,DBLP:conf/aaai/Desai0A22}, limiting their ability to assess the robustness of MLLMs across various user question formats. As depicted in Figure~\ref{Fig:Intro}, the model can answer the \textit{Yes/No QA} correctly but struggle with the \textit{Matching QA}, highlighting performance variations across question formats. \textbf{Third, prior works}~\cite{DBLP:conf/aaai/QiaoJSCZN23,DBLP:conf/acl/HesselMHLDZM023} \textbf{solely focus on humor or sarcasm within a narrow domain} (\eg cartoon). This limits their applicability to broader real-world scenarios that convey punchlines, and hence causes insufficient evaluations.
\\ \indent
In light of the above limitations, we introduce a novel multimodal \textbf{Punch}line comprehension \textbf{Bench}mark, \textbf{PunchBench} for short, designed to provide an accurate and comprehensive evaluation of this task. To enhance \textbf{evaluation accuracy}, we modify captions to mitigate the impact of potential shortcuts. Specifically, we apply context consistency adaptation to eliminate inconsistent captions, and then use word substitution and inversion to generate synonymous and antonymous captions with the help of ChatGPT~\cite{chatgpt}. Regarding \textbf{evaluation comprehensiveness}, PunchBench features diversity across multiple dimensions. For punchline types, it includes both humor and sarcasm. For task types, it involves two levels of punchline understanding: shallow-level punchline perception and deep-level punchline reasoning. Each task employs diverse question formats: \textit{Yes/No QA}, \textit{Matching QA}, \textit{Multi-option QA} and \textit{Generation QA}. Furthermore, PunchBench spans a wide range of multimodal content domains, including posts, cartoons, comments, and memes. In total, PunchBench comprises $6,000$ image-caption pairs and $54,000$ question-answer pairs, allowing a comprehensive evaluation.

Leveraging PunchBench, we evaluate a range of state-of-the-art MLLMs. The results reveal a significant gap between MLLMs and humans in punchline comprehension. Additionally, the performance of MLLMs varies across different question formats, and shows notable degradation when faced with synonymous or antonymous captions. These observations emphasize the importance of incorporating diverse question formats, synonymous and antonymous captions in the evaluation process.
\\ \indent
To improve the punchline understanding ability of MLLMs, we propose a strategy called \textbf{Simple-to-Complex Chain-of-Question} (SC-CoQ), inspired by the simple-to-complex progression for solving complicated problems.
SC-CoQ structures questions from simple to complex within and across tasks, enabling the models to incrementally develop the capability to address complex questions by first mastering simple ones. Compared to in-context learning~\cite{DBLP:conf/nips/BrownMRSKDNSSAA20} and chain-of-thought~\cite{DBLP:conf/nips/Wei0SBIXCLZ22} methods, SC-CoQ demonstrates superior performance, further validating its effectiveness in promoting punchline comprehension.

In a nutshell, our contributions can be summarized as follows.
\begin{itemize}
    \item We introduce PunchBench, which, to the best of our knowledge, is the first benchmark for accurate and comprehensive evaluation of multimodal punchline comprehension.
    \item Extensive evaluations on PunchBench reveal a significant gap between MLLMs and humans in punchline comprehension, and highlights the performance variations across question formats in each task.
    \item We propose Simple-to-Complex Chain-of-Question (SC-CoQ), which follows a progression from simple to complex questions to effectively improve punchline comprehension.
\end{itemize}

%% file: Chapter/2_related.tex
\section{Related Works}
\subsection{Multimodal Large Language Models}
Large Language Models (LLMs) for pure text like ChatGPT~\cite{chatgpt}, GPT-4~\cite{openai2024gpt4}, and LLaMA~\cite{DBLP:journals/corr/abs-2302-13971} have proved impressive comprehension capabilities of text. Following this success and to expand it on multimodal tasks, many efforts~\cite{DBLP:conf/icml/0008LSH23,DBLP:conf/nips/LiuLWL23a} have been made to integrate visual comprehension capability into LLMs, and lead to a blowout of Multimodal Large Language Models (MLLMs), both closed-source models (\eg GPT-4V~\cite{GPT-4V} and GPT-4o~\cite{GPT-4o}) and open-source models (\eg LLaVA series~\cite{DBLP:conf/nips/LiuLWL23a,DBLP:conf/cvpr/LiuLLL24,liu2024llavanext}, CogVLM series~\cite{wang2023cogvlm,hong2024cogvlm2}, Qwen-VL family~\cite{Qwen-VL,Qwen2VL} and GLM-4V~\cite{glm2024chatglm}). They demonstrate unprecedented and surprising multimodal understanding capabilities in vision-language tasks such as visual question answering~\cite{vqaiccv}, dense image captioning~\cite{DBLP:conf/cvpr/JohnsonKF16} and optical character recognition~\cite{Islam2017ASO}.
 
\subsection{Punchline Comprehension}
Despite significant progress of MLLMs in understanding factual information from visual content~\cite{DBLP:conf/cvpr/LongWHXRZZL23,DBLP:conf/emnlp/JianY024}, the punchline comprehension capabilities~\cite{DBLP:conf/acl/CaiCW19,DBLP:journals/corr/abs-2402-03658} of MLLMs still lack sufficient evaluations.
Prior works~\cite{DBLP:conf/aaai/Desai0A22,DBLP:conf/acl/KumarKA022,DBLP:conf/acl/HesselMHLDZM023} related to multimodal punchline comprehension have concentrated on sarcasm or humor. For example, \citeauthor{DBLP:conf/aaai/Desai0A22}
curated the MORE dataset for multimodal sarcasm explanation, which aims to explain the ironic semantics of multimodal post.
Furthermore, previous benchmarks overlooked potential shortcuts in captions that MLLMs may exploit to answer questions, undermining true comprehension of punchlines.
Noticing these concerns, our benchmark is introduced to provide an accurate and comprehensive evaluation of multimodal punchline comprehension.

%% file: Chapter/3_dataset_task.tex
\begin{figure*}
    \centering
    \includegraphics[width=0.9\textwidth]{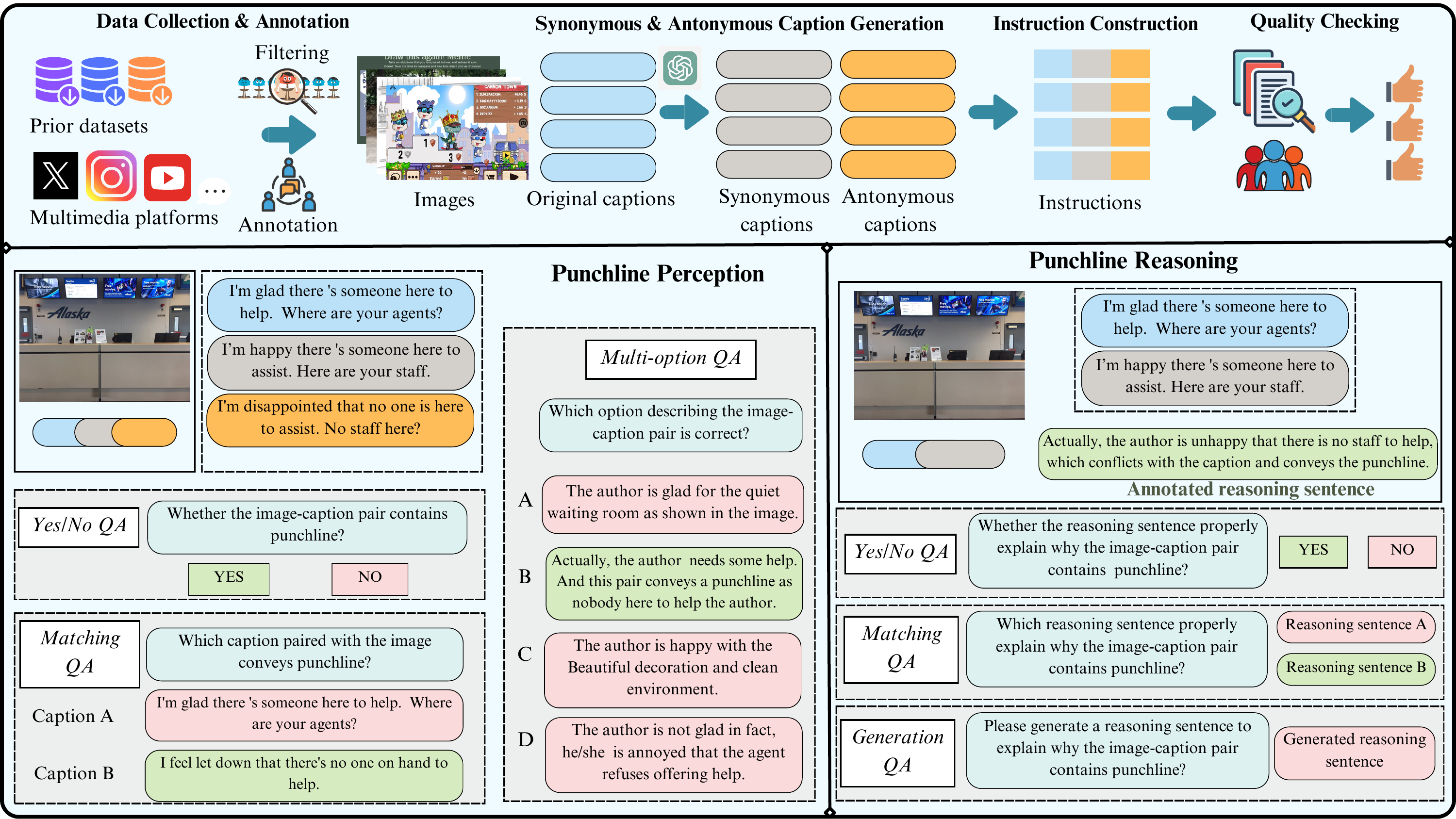}
     \caption{\textbf{Upper}: Data collection workflow for PunchBench. We first collect image-caption pairs from prior datasets and multimedia platforms with meticulous filtering, conduct human annotation to obtain the corresponding labels and reasoning sentences for the pairs. And we then utilize \texttt{gpt-3.5-turbo-0125} to generate synonymous and antonymous captions corresponding to the original captions. Based on these image-caption pairs, we construct corresponding instructions for punchline perception and reasoning. Finally, we perform quality checking to ensure the reliability of our PunchBench. \textbf{Lower}: Data examples for Punchline Perception and Punchline Reasoning.}
    \label{Fig:MOCK}
\end{figure*}
\section{PunchBench}
As illustrated in Figure~\ref{Fig:MOCK}, our PunchBench is constructed in four steps: Source Data Collection \& Annotation (\sectionref{sec:data}), Synonymous \& Antonymous Caption Generation (\sectionref{sec:caption}), Instruction Construction (\sectionref{sec:instruction}), Quality Checking (\sectionref{sec:checking}). In this section, we elaborate on the construction process as well as the data statistics (\sectionref{sec:statistics}).

\subsection{Source Data Collection \& Annotation}
\label{sec:data}
The image-caption pairs in our dataset are obtained from two sources.
1) Prior datasets.~Recognizing the wealth of resources in prior datasets that contribute to punchline comprehension, we select three relevant datasets, \ie MTSD~\cite{castro-etal-2019-towards}, MORE~\cite{DBLP:conf/acl/KumarKA022} and HUB~\cite{DBLP:conf/acl/HesselMHLDZM023}. Then, we meticulously filter the high-quality image-caption pairs using a hybrid approach that combines both manual and automatic filtering, as detailed in Appendix~\ref{Appendix:daa_collection}.
2) Multimedia platforms. To ensure up-to-date of our dataset, we gather image-caption pairs from the social media platforms, such as X, Instagram, and YouTube. Additionally, we include image-caption pairs from the cartoon websites like CartoonMovement and CartoonStock.
The information about these multimeida platforms is provided in Appendix~\ref{Appendix:Platform}.
\\ \indent
After obtaining the raw set of image-caption pairs, we implement a crowd voting process, which is outlined in Appendix~\ref{Appendix:daa_collection}, to identify a label indicating whether the image-caption pair contains punchline. Ultimately, we compile a collection of $6,000$ image-caption pairs spanning diverse scenarios (\eg cartoon, post, comment, and meme), half of which are identified as containing punchline. 
To explain why the particular pair contains punchline, we employ three human annotators to handcraft reasoning sentence for it, which is detailed in Appendix~\ref{Appendix:daa_collection}.
Finally, we acquire $6,000$ image-caption pairs along with their corresponding labels and reasoning sentences.
To emphasize the superiority of PunchBench, we provide a comparison between our PunchBench and prior datasets in Table~\ref{Tab:Benchmark}.
\subsection{Synonymous \& Antonymous Caption Generation}\label{sec:caption}
As aforementioned, MLLMs may exploit shortcuts in the captions, such as word bias and context inconsistency, to answer the question without truly understanding the image-caption pair. To prevent these shortcuts, we generate \textbf{synonymous caption} and \textbf{antonymous caption} for each image-caption pair through following methods. 
1) \textit{Word substitution and inversion}. Assisted by \texttt{gpt-3.5-turbo-0125}, we substitute the sentiment, action, object and other words with synonymous words to generate synonymous caption, and we invert the semantics by replacing these words with their antonyms to obtain antonymous caption. 
2) \textit{Context consistency adaption}. To adapt the consistency of captions containing semantically conflicting components, \eg ``I am so glad today! What a disgusting rainy day!'', we first leverage \texttt{gpt-3.5-turbo-0125} to identify and isolate the two conflicting parts, ``I am so glad today'' contradicts ``What a disgusting rainy day''. And we then employ word substitution and inversion for the two parts to generate synonymous and antonymous caption.
We supplement additional implementation details in Appendix~\ref{Appendix:caption}.
\begin{figure*}
    \centering
    \includegraphics[width=0.95\textwidth]{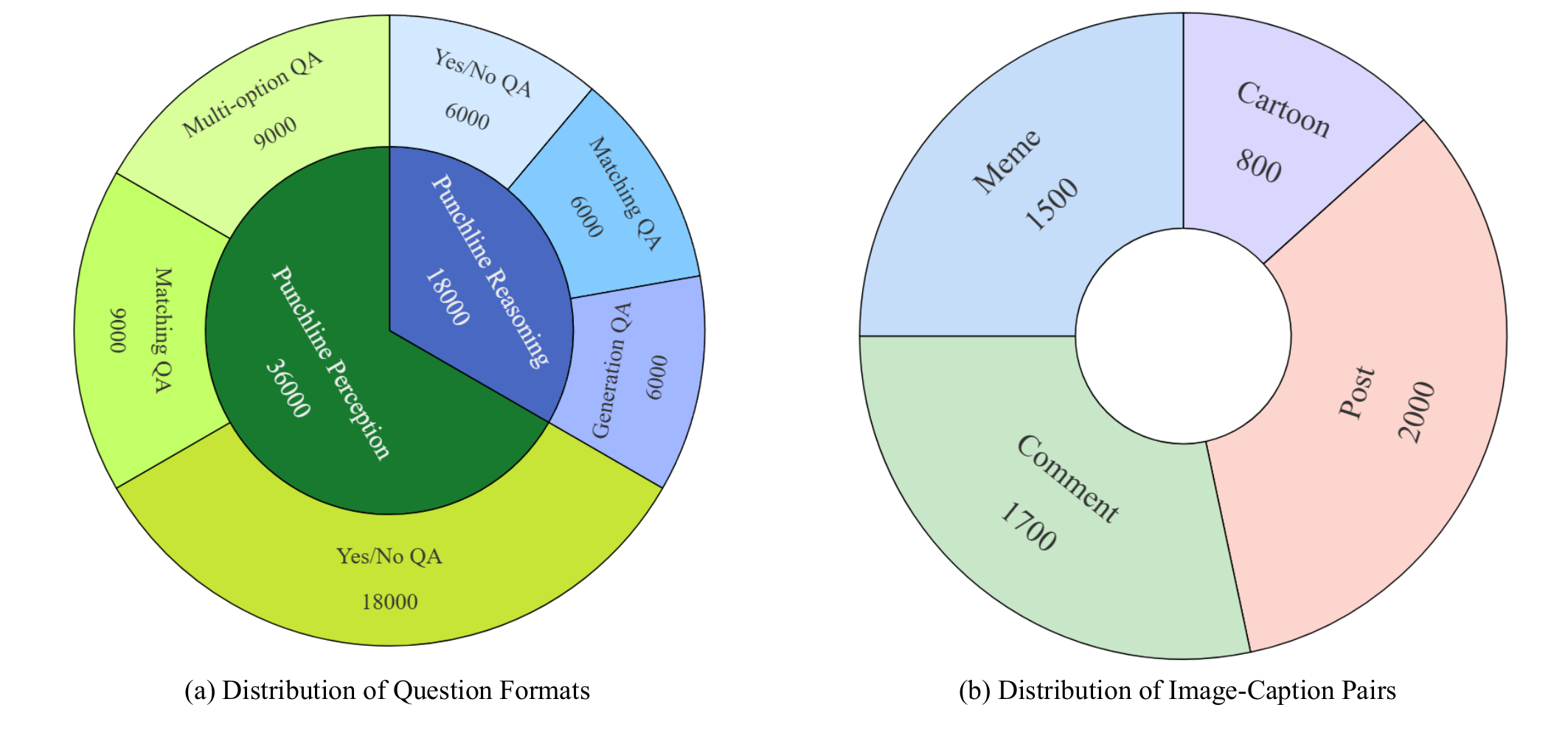}
     \caption{The overall data statistics of our PunchBench.}
    \label{Fig:data}
\end{figure*}
\subsection{Instruction Construction}
\label{sec:instruction}

Based on the collected image-caption pairs and corresponding annotations, we now construct instructions for two types of tasks: \textbf{Punchline Perception}, which assesses whether an MLLM can identify the existence of punchline in image-caption pairs, and \textbf{Punchline Reasoning}, which requires the model to understand the reason why a particular image-caption pair contains punchline. Figure~\ref{Fig:MOCK} illustrates some examples of the instructions. Before delving into the details, we first clarify some notations.

\paragraph{Notations.} Each image-caption pair $P^x_i=\text{<}I_i, C^x_i\text{>}$ consists of an image $I_i$ and a caption $C^x_i$, where $x\in\{o, s, a\}$ denotes the original ($C^o$), synonymous ($C^s$) and antonymous ($C^a$) caption. And each pair is assigned a label $L^x_i\in\{0,1\}$, where $1$ indicates that the pair contains punchline while $0$ is opposite. Notably, $P^s_i$ shares the same label as $P^o_i$, while $P^a_i$ serves as the contrast. We detail instruction construction process as follows, temporally omitting the subscript $i$ that indexes the samples for simplicity.

\subsubsection{Punchline Perception}
\label{subsec:perception}
\paragraph{Yes/No QA.} 
The model is required to answer whether the given image-caption pair $P^x$ contains punchline. The instruction is derived based on various instruction templates, with the answer ``Yes'' or ``No'' being determined by the label $L^x$. 
To attain a balance, the number of negative answers is equal to that of positive answers. 
\paragraph{Matching QA.} 
The model is asked to select between two captions, recognizing which one effectively conveys punchline with the given image. For pair $P^x$ containing punchline, we utilize \texttt{gpt-4o-2024-05-13}\footnote{\url{https://platform.openai.com/docs/models}.} to generate a distractor caption $C^d$ for the image $I$. The distractor caption $C^d$ just describes the content of image $I$ without conveying the punchline.
Finally, the image-caption pair $P^x$, as well as $C_d$ are subsequently integrated into several templates to obtain the instructions. 
To prevent bias associated with the position of captions, we randomize the order in which the two captions are displayed for each instruction. 
\paragraph{Multi-option QA.}
The model aims to discern the correct one from four options \ie $O_1, O_2, O_3, O_4$ describing the image-caption pair $P^x$. The four options are generated by \texttt{gpt-3.5-turbo-0125} based on the caption $C^x$ and former distractor caption $C^d$, with only one being correct. These options, along with $P^x$ are incorporated into the instruction templates.
The sequence of the four options are shuffled to avoid the positional bias.

\subsubsection{Punchline Reasoning}
\label{subsec:reasoning}
We utilize the $3,000$ pairs $P^o$ containing punchline, their synonymous captions $C^s$ and annotated reasoning sentences $R^a$ to construct instructions for punchline reasoning. 
\paragraph{Yes/No QA.}
Presented with an image-caption pair and a reasoning sentence, the model is asked to identify whether the reasoning sentence succeeds in explaining why the pair contains punchline. Specifically, we first resort to \texttt{gpt-3.5-turbo-0125} to generate distractor reasoning sentence $R^d$ based on our annotated reasoning sentence $R^a$. 
And we then randomly assign half of the image-caption pairs to annotated reasoning sentences $R^o$, while the other part is linked to the distractor ones $R^d$, incorporate them into instruction templates.
The answer to instruction using $R^a$ is ``Yes'' and using $R^d$ is ``No''.
Finally, we have an equal number of positive and negative instructions.
\paragraph{Matching QA.} 
Given an image-caption pair and two reasoning sentences, \ie $R^a$ (correct) and $R^d$ (distractor), only one of which appropriately interprets the punchline in the pair, the model is required to select the correct reasoning sentence. Specifically, $R^a$ and $R^d$ are paired with $P^o$ or $P^s$ in several templates to construct the instructions, with the order of $R^d$ and $R^a$ being randomly shuffled.
\paragraph{Generation QA.} 
In this task, the image-caption pair is utilized in various instruction templates to prompt the model to generate a reasoning sentence to explain the punchline, with $R^a$ serving as the reference answer.
\\
The above instructions undergo a thorough review and refinement process by human annotators. 
The instruction templates and more details of this construction process are supplied in Appendix~\ref{Appendix:instruction}.

\subsection{Quality Checking}
\label{sec:checking}
To ensure the quality of PunchBench, we randomly sample $100$ instructions for each question format, excluding \textit{Generation QA}, for quality checking process.
Three human annotators are employed to answer the questions guided by the sampled instructions. Human annotators have an extra option ``CBA'' that means ``Cannot Be Answered'' for each question. Among $500$ instructions, only $1$ is labeled by ``CBA'', which verifies the high quality of the instructions. 
Moreover, they answer the questions with high accuracy as results reported in Table~\ref{tab:main_res}, which further demonstrates the superior quality of our dataset.
\subsection{Dataset Statistics}
\label{sec:statistics}
We illustrate Figure~\ref{Fig:data} to exhibit the dataset statistics of our PunchBench. PunchBench consists of $6,000$ image-caption pairs, spanning cartoon, post, comment and meme. Each image has three types of captions: original, synonymous, and antonymous captions. Our question formats include \textit{Yes/No QA}, \textit{Matching QA}, and \textit{Multi-option QA} for punchline perception, and \textit{Yes/No QA}, \textit{Matching QA}, and \textit{Generation QA} for punchline reasoning.  
Above all, our PunchBench covers a diverse question formats and domains, which can provide a comprehensive evaluation.
We also compare our PunchBench with previous Benchmarks in Appendix~\ref{Appendix:statistics}.

\input{Table/main_results}
\section{Simple-to-Complex Chain-of-Question}
\label{sec:SC-CoQ}
In our initial evaluation (the ``Zero-shot'' results in Table~\ref{tab:main_res}), we observe that different question formats present varying levels of difficulty for the MLLMs. The general trend for punchline perception is \textit{Yes/No QA} < \textit{Matching QA} < \textit{Multi-option QA}, and for punchline reasoning, it is \textit{Yes/No QA} < \textit{Matching QA} < \textit{Generation QA}, where < indicates easier than. Inspired by these observations, we propose a Simple-to-Complex Chain-of-Question (SC-CoQ) strategy, which prompts MLLMs to answer the simpler questions before solving the most complex questions. Specifically, we introduce two variations of SC-CoQ, Intra-task and Inter-task:

\noindent\textbf{Intra-task SC-CoQ} integrates the various formats of questions within the same task to improve performance on the most challenging question (\ie \textit{Multi-option QA and Generation QA}). We sequence the questions in a specific order mirroring simple to complex, \ie <\textit{Yes/No QA}, \textit{Matching QA}, \textit{Multi-option QA} or \textit{Generation QA}>. 

\noindent\textbf{Inter-task SC-CoQ} incorporates similar question formats (\ie \textit{Yes/No QA} and \textit{Matching QA}) across different tasks to enhance punchline comprehension.~For \textit{Yes/No QA}, we sequentially link the questions from the two tasks, \ie <$\textit{Yes/No QA}_m$, $\textit{Yes/No QA}_n$> or <$\textit{Yes/No QA}_n$, $\textit{Yes/No QA}_m$>, where $m$ refers to punchline perception task and $n$ denotes punchline reasoning task. For \textit{Matching QA}, this chain utilizes both \textit{Yes/No QA} and \textit{Matching QA} to reinforce punchline comprehension across tasks, \ie <$\textit{Yes/No QA}_m$, $\textit{Yes/No QA}_n$, $\textit{Matching QA}_m$, $\textit{Matching QA}_n$> or <$\textit{Yes/No QA}_n$, $\textit{Yes/No QA}_m$, $\textit{Matching QA}_n$, $\textit{Matching QA}_m$>. More details of SC-CoQ and specific prompting examples can be found in Appendix~\ref{Appendix:SC-CoQ}.

%% file: Table/main_results.tex
\begin{table*}[]
\caption{Evaluation results on PunchBench. The best results among the MLLMs are in \textbf{boldface}, while the second best are {\ul underlined}. $\star$ denotes the best results among the prompting methods. The results are the average of four replicates. And the P-value between SC-CoQ performance and other prompting method results is consistently less than $0.01$.}
\label{tab:main_res}
\resizebox{\textwidth}{!}{
\begin{tabular}{lccccccccccccc}
\hline
\multicolumn{1}{l|}{\multirow{2}{*}{\textbf{Model}}} &
  \multicolumn{1}{l|}{\multirow{2}{*}{\textbf{\#Params}}} &
  \multicolumn{4}{c|}{\textbf{Yes/No QA}} &
  \multicolumn{4}{c|}{\textbf{Matching QA}} &
  \multicolumn{4}{c}{\textbf{Multi-choice QA}} \\ \cline{3-14} 
\multicolumn{1}{c|}{} &
  \multicolumn{1}{c|}{} &
  \multicolumn{1}{c|}{\textbf{Zero-shot}} &
    \multicolumn{1}{c|}{\textbf{CoT}} &
  \multicolumn{1}{c|}{\textbf{3 shot}} &
  \multicolumn{1}{c|}{\textbf{SC-CoQ}} &
  \multicolumn{1}{c|}{\textbf{Zero-shot}} &
    \multicolumn{1}{c|}{\textbf{CoT}} &
  \multicolumn{1}{c|}{\textbf{3 shot}} &
  \multicolumn{1}{c|}{\textbf{SC-CoQ}} &
  \multicolumn{1}{c|}{\textbf{Zero-shot}} &
      \multicolumn{1}{c|}{\textbf{CoT}} &
    \multicolumn{1}{c|}{\textbf{3 shot}} &
  \multicolumn{1}{c}{\textbf{SC-CoQ}} 
 \\
 \hline
\multicolumn{1}{l|}{\textbf{LLaVA}} &
  \multicolumn{1}{c|}{7B} &
  \multicolumn{1}{c|}{62.7} &
    \multicolumn{1}{c|}{61.5} &
  \multicolumn{1}{c|}{63.5} &
  \multicolumn{1}{c|}{$64.8\star$} &
  \multicolumn{1}{c|}{54.2} &
  \multicolumn{1}{c|}{54.9} &
    \multicolumn{1}{c|}{55.8} &
  \multicolumn{1}{c|}{$57.1\star$} &
  \multicolumn{1}{c|}{36.4} &
  \multicolumn{1}{c|}{37.5} &
    \multicolumn{1}{c|}{37.2} &
  $39.1\star$
  \\
\multicolumn{1}{l|}{\textbf{GLM-4V}} &
  \multicolumn{1}{c|}{9B} &
  \multicolumn{1}{c|}{61.4} &
  \multicolumn{1}{c|}{61.8} &
    \multicolumn{1}{c|}{62.2} &
  \multicolumn{1}{c|}{$63.7\star$} &
  
  \multicolumn{1}{c|}{55.3} &
    \multicolumn{1}{c|}{53.1} &
  \multicolumn{1}{c|}{56.9} &
  \multicolumn{1}{c|}{$57.7\star$} &
  
  \multicolumn{1}{c|}{38.2} &
    \multicolumn{1}{c|}{38.8} &
  \multicolumn{1}{c|}{39.5} &
  $40.6\star$ 
   \\
\multicolumn{1}{l|}{\textbf{Qwen2-VL-2B-Instruct}} &
  \multicolumn{1}{c|}{2B} &
  \multicolumn{1}{c|}{56.9} &
  \multicolumn{1}{c|}{57.2} &
    \multicolumn{1}{c|}{57.4} &
  \multicolumn{1}{c|}{58.0$\star$} &
  
\multicolumn{1}{c|}{52.3} &
  \multicolumn{1}{c|}{52.0} &
  \multicolumn{1}{c|}{51.8} &
  \multicolumn{1}{c|}{53.2$\star$} &
  
  \multicolumn{1}{c|}{33.1} &
    \multicolumn{1}{c|}{33.5} &
  \multicolumn{1}{c|}{33.4} &
  34.1$\star$
  \\
\multicolumn{1}{l|}{\textbf{Qwen2-VL-7B-Instruct}} &
  \multicolumn{1}{c|}{7B} &
  \multicolumn{1}{c|}{70.1} &
    \multicolumn{1}{c|}{71.9} &
  \multicolumn{1}{c|}{72.4} &
  \multicolumn{1}{c|}{$73.2\star$} &
  
    \multicolumn{1}{c|}{58.0} &
      \multicolumn{1}{c|}{58.4} &
  \multicolumn{1}{c|}{59.2} &
  \multicolumn{1}{c|}{$61.3\star$} &
  
  \multicolumn{1}{c|}{41.7} &
  \multicolumn{1}{c|}{43.0} &
    \multicolumn{1}{c|}{42.4} &
  $44.1\star$ 
  \\
  \multicolumn{1}{l|}{\textbf{Qwen2-VL-72B-Instruct}} &
  \multicolumn{1}{c|}{72B} &
  \multicolumn{1}{c|}{73.7} &
    \multicolumn{1}{c|}{{\ul74.8}} &
  \multicolumn{1}{c|}{ 74.5} &
  \multicolumn{1}{c|}{$76.1\star$} &
  
    \multicolumn{1}{c|}{60.2} &
      \multicolumn{1}{c|}{61.5} &
  \multicolumn{1}{c|}{61.7} &
  \multicolumn{1}{c|}{$62.9\star$} &
  
  \multicolumn{1}{c|}{48.8} &
  \multicolumn{1}{c|}{49.7} &
    \multicolumn{1}{c|}{50.1} &
  $51.7\star$ 
  \\
  \multicolumn{1}{l|}{\textbf{CogVLM2}} &
  \multicolumn{1}{c|}{19B} &
  \multicolumn{1}{c|}{68.2} &
    \multicolumn{1}{c|}{67.6} &
  \multicolumn{1}{c|}{69.5} &
  \multicolumn{1}{c|}{$71.3\star$} &
  
    \multicolumn{1}{c|}{57.3} &
  \multicolumn{1}{c|}{58.9} &
    \multicolumn{1}{c|}{58.6} &
  \multicolumn{1}{c|}{$60.8\star$} &
  
  \multicolumn{1}{c|}{43.4} &
    \multicolumn{1}{c|}{44.2} &
  \multicolumn{1}{c|}{44.7} &
  $46.3\star$ 
   \\
\multicolumn{1}{l|}{\textbf{LLaVA-OneVision}}&
\multicolumn{1}{c|}{7B} &
  \multicolumn{1}{c|}{64.3} &
    \multicolumn{1}{c|}{65.8} &
  \multicolumn{1}{c|}{66.0} &
  \multicolumn{1}{c|}{$67.2\star$} &
  
    \multicolumn{1}{c|}{55.9} &
  \multicolumn{1}{c|}{56.4} &
    \multicolumn{1}{c|}{56.8} &
  \multicolumn{1}{c|}{$57.9\star$} &
  
  \multicolumn{1}{c|}{39.7} &
    \multicolumn{1}{c|}{41.1} &
  \multicolumn{1}{c|}{40.3} &
  $42.4\star$ 
  \\
  \multicolumn{1}{l|}{\textbf{InternVL2.5}}&
  \multicolumn{1}{c|}{8B}&
   \multicolumn{1}{c|}{69.5} &
    \multicolumn{1}{c|}{70.1} &
  \multicolumn{1}{c|}{70.7} &
  \multicolumn{1}{c|}{$71.4\star$} &
  
    \multicolumn{1}{c|}{58.4} &
  \multicolumn{1}{c|}{59.0} &
    \multicolumn{1}{c|}{59.2} &
  \multicolumn{1}{c|}{$60.0\star$} &
  
  \multicolumn{1}{c|}{42.0} &
    \multicolumn{1}{c|}{42.9} &
  \multicolumn{1}{c|}{43.1} &
  $44.3\star$ 
  \\
  \multicolumn{1}{l|}{\textbf{MiniCPM-o 2.6}}&
  \multicolumn{1}{c|}{8B}&  
  \multicolumn{1}{c|}{70.8} &
    \multicolumn{1}{c|}{71.7} &
  \multicolumn{1}{c|}{71.4} &
  \multicolumn{1}{c|}{$72.3\star$} &
  
    \multicolumn{1}{c|}{59.1} &
  \multicolumn{1}{c|}{59.6} &
    \multicolumn{1}{c|}{60.1} &
  \multicolumn{1}{c|}{$61.2\star$} &
  
  \multicolumn{1}{c|}{43.1} &
    \multicolumn{1}{c|}{43.7} &
  \multicolumn{1}{c|}{43.5} &
  $45.4\star$ 
  \\
  \multicolumn{1}{l|}{\textbf{Aria}}&
  \multicolumn{1}{c|}{3.5B$\times$8}&  
  \multicolumn{1}{c|}{72.1} &
    \multicolumn{1}{c|}{72.9} &
  \multicolumn{1}{c|}{73.2} &
  \multicolumn{1}{c|}{$74.5\star$} &
  
    \multicolumn{1}{c|}{61.8} &
  \multicolumn{1}{c|}{62.7} &
    \multicolumn{1}{c|}{62.3} &
  \multicolumn{1}{c|}{$63.6\star$} &
  
  \multicolumn{1}{c|}{47.9} &
    \multicolumn{1}{c|}{49.0} &
  \multicolumn{1}{c|}{48.6} &
  $50.8\star$ 
   \\
\multicolumn{1}{l|}{\textbf{GPT-4V}} &
  \multicolumn{1}{c|}{-} &
  \multicolumn{1}{c|}{75.0} &
    \multicolumn{1}{c|}{ 74.2} &
  \multicolumn{1}{c|}{{\ul 76.2}} &
  \multicolumn{1}{c|}{{\ul 78.1}$\star$} &
  
  \multicolumn{1}{c|}{62.1} &
    \multicolumn{1}{c|}{{\ul 63.2}} &
  \multicolumn{1}{c|}{{\ul 63.9}} &
  \multicolumn{1}{c|}{{\ul 65.0}$\star$} &
  
  \multicolumn{1}{c|}{48.1} & 
    \multicolumn{1}{c|}{{\ul50.5}} &
  \multicolumn{1}{c|}{{\ul 50.3}} &
  {\ul51.9}$\star$
\\
\multicolumn{1}{l|}{\textbf{GPT-4o}} &
  \multicolumn{1}{c|}{-} &
  \multicolumn{1}{c|}{{\ul 77.5}} &
  \multicolumn{1}{c|}{\textbf{78.6}} &
    \multicolumn{1}{c|}{\textbf{79.2}} &
  \multicolumn{1}{c|}{\textbf{80.7}$\star$} &
  
  \multicolumn{1}{c|}{{\ul 64.2}} &
    \multicolumn{1}{c|}{\textbf{66.3}} &
  \multicolumn{1}{c|}{\textbf{65.4}} &
  \multicolumn{1}{c|}{\textbf{67.9}$\star$} &
  
  \multicolumn{1}{c|}{{\ul 50.8}} &
    \multicolumn{1}{c|}{\textbf{51.4}} &
  \multicolumn{1}{c|}{\textbf{52.0}} &
  \textbf{53.1}$\star$
 \\ \hline
\multicolumn{1}{l|}{\textbf{Human}} &
 \multicolumn{1}{c|}{-} &
  \multicolumn{1}{c|}{\textbf{98.3}} &
  \multicolumn{1}{c|}{-} &
  \multicolumn{1}{c|}{-} &
  \multicolumn{1}{c|}{-} &
  \multicolumn{1}{c|}{\textbf{97.7}} &
  \multicolumn{1}{c|}{-} &
  \multicolumn{1}{c|}{-} &
  \multicolumn{1}{c|}{-} &
  \multicolumn{1}{c|}{\textbf{90.7}} &
  \multicolumn{1}{c|}{-} &
  \multicolumn{1}{c|}{-} &
  -
  \\ \hline
\multicolumn{14}{c}{(a) Punchline Perception} \\ \hline
\multicolumn{1}{l|}{\multirow{2}{*}{\textbf{Model}}} &
  \multicolumn{1}{l|}{\multirow{2}{*}{\textbf{\#Params}}} &
  \multicolumn{4}{c|}{\textbf{Yes/No QA}} &
  \multicolumn{4}{c|}{\textbf{Matching QA}} &
  \multicolumn{4}{c}{\textbf{Generation QA}} \\ \cline{3-14} 
\multicolumn{1}{c|}{} &
  \multicolumn{1}{c|}{} &
  \multicolumn{1}{c|}{\textbf{Zero-shot}} &
    \multicolumn{1}{c|}{\textbf{CoT}} &
  \multicolumn{1}{c|}{\textbf{3 shot}} &
  \multicolumn{1}{c|}{\textbf{SC-CoQ}} &
  \multicolumn{1}{c|}{\textbf{Zero-shot}} &
    \multicolumn{1}{c|}{\textbf{CoT}} &
  \multicolumn{1}{c|}{\textbf{3 shot}} &
  \multicolumn{1}{c|}{\textbf{SC-CoQ}} &
  \multicolumn{1}{c|}{\textbf{Zero-shot}} &
      \multicolumn{1}{c|}{\textbf{CoT}} &
    \multicolumn{1}{c|}{\textbf{3 shot}} &
  \multicolumn{1}{c}{\textbf{SC-CoQ}} 
 \\
 \hline
\multicolumn{1}{l|}{\textbf{LLaVA}} &
  \multicolumn{1}{c|}{7B} &
  \multicolumn{1}{c|}{60.1} &
     \multicolumn{1}{c|}{61.7} &
  \multicolumn{1}{c|}{61.3} &
  \multicolumn{1}{c|}{62.6$\star$}&
  
  \multicolumn{1}{c|}{50.7} &
   \multicolumn{1}{c|}{51.3} &
     \multicolumn{1}{c|}{51.9} &
  \multicolumn{1}{c|}{53.0$\star$}&
  
   \multicolumn{1}{c|}{35.2} &
       \multicolumn{1}{c|}{37.1} &
   \multicolumn{1}{c|}{36.6} &
  38.7$\star$
  \\
\multicolumn{1}{l|}{\textbf{GLM-4V}} &
  \multicolumn{1}{c|}{9B} &
  \multicolumn{1}{c|}{59.7} &
     \multicolumn{1}{c|}{60.8} &
  \multicolumn{1}{c|}{61.3} &
  \multicolumn{1}{c|}{$62.9\star$} &
  
  \multicolumn{1}{c|}{53.1} &
     \multicolumn{1}{c|}{52.2} &
  \multicolumn{1}{c|}{54.8} &
  \multicolumn{1}{c|}{$55.9\star$} &
  
   \multicolumn{1}{c|}{37.1} &
      \multicolumn{1}{c|}{38.5} &
  \multicolumn{1}{c|}{38.2} &
  39.8$\star$
   \\
\multicolumn{1}{l|}{\textbf{Qwen2-VL-2B-Instruct}} &
  \multicolumn{1}{c|}{2B} &
  \multicolumn{1}{c|}{54.2} &
  \multicolumn{1}{c|}{55.1} &
    \multicolumn{1}{c|}{54.0} &
  \multicolumn{1}{c|}{55.9$\star$} &
  
\multicolumn{1}{c|}{49.5} &
  \multicolumn{1}{c|}{49.0} &
  \multicolumn{1}{c|}{50.6} &
  \multicolumn{1}{c|}{51.4$\star$} &
  
  \multicolumn{1}{c|}{31.7} &
    \multicolumn{1}{c|}{32.1} &
  \multicolumn{1}{c|}{31.5} &
  33.2$\star$
  \\
\multicolumn{1}{l|}{\textbf{Qwen2-VL-7B-Instruct}} &
  \multicolumn{1}{c|}{7B} &
  \multicolumn{1}{c|}{64.5} &
  \multicolumn{1}{c|}{65.3} &
    \multicolumn{1}{c|}{66.0} &
  \multicolumn{1}{c|}{67.4$\star$} &
  
\multicolumn{1}{c|}{55.7} &
  \multicolumn{1}{c|}{56.1} &
  \multicolumn{1}{c|}{57.2} &
  \multicolumn{1}{c|}{58.4$\star$} &
  
  \multicolumn{1}{c|}{40.6} &
    \multicolumn{1}{c|}{41.5} &
  \multicolumn{1}{c|}{41.9} &
  43.7$\star$
   \\
\multicolumn{1}{l|}{\textbf{Qwen2-VL-72B-Instruct}} &
  \multicolumn{1}{c|}{72B} &
  \multicolumn{1}{c|}{72.0} &
  \multicolumn{1}{c|}{72.7} &
    \multicolumn{1}{c|}{73.0} &
  \multicolumn{1}{c|}{74.9$\star$} &
  
\multicolumn{1}{c|}{57.5} &
  \multicolumn{1}{c|}{{\ul59.1}} &
  \multicolumn{1}{c|}{{\ul59.4}} &
  \multicolumn{1}{c|}{60.4$\star$} &
  
  \multicolumn{1}{c|}{45.0} &
    \multicolumn{1}{c|}{46.1} &
  \multicolumn{1}{c|}{{\ul 46.7}} &
  {\ul48.0}$\star$
  \\
\multicolumn{1}{l|}{\textbf{CogVLM2}} &
  \multicolumn{1}{c|}{19B} &
  \multicolumn{1}{c|}{66.3} &
     \multicolumn{1}{c|}{67.2} &
  \multicolumn{1}{c|}{68.0} &
  \multicolumn{1}{c|}{69.6$\star$} &
  
  \multicolumn{1}{c|}{54.2} &
     \multicolumn{1}{c|}{54.9} &
  \multicolumn{1}{c|}{55.4} &
  \multicolumn{1}{c|}{56.3$\star$} &
  
   \multicolumn{1}{c|}{41.8} &
      \multicolumn{1}{c|}{42.7} &
  \multicolumn{1}{c|}{42.5} &
  43.4$\star$
  \\
\multicolumn{1}{l|}{\textbf{LLaVA-OneVision}}&
\multicolumn{1}{c|}{7B}   &
\multicolumn{1}{c|}{61.7} &
    \multicolumn{1}{c|}{61.2} &
  \multicolumn{1}{c|}{62.8} &
  \multicolumn{1}{c|}{$63.9\star$} &
  
    \multicolumn{1}{c|}{52.4} &
  \multicolumn{1}{c|}{53.5} &
    \multicolumn{1}{c|}{53.9} &
  \multicolumn{1}{c|}{$54.7\star$} &
  
  \multicolumn{1}{c|}{37.5} &
    \multicolumn{1}{c|}{38.2} &
  \multicolumn{1}{c|}{38.7} &
  $40.1\star$ 
  \\
  \multicolumn{1}{l|}{\textbf{InternVL2.5}}&
  \multicolumn{1}{c|}{8B}  &
\multicolumn{1}{c|}{63.8} &
    \multicolumn{1}{c|}{64.9} &
  \multicolumn{1}{c|}{64.3} &
  \multicolumn{1}{c|}{$65.8\star$} &
  
    \multicolumn{1}{c|}{54.6} &
  \multicolumn{1}{c|}{55.8} &
    \multicolumn{1}{c|}{55.5} &
  \multicolumn{1}{c|}{$56.9\star$} &
  
  \multicolumn{1}{c|}{40.7} &
    \multicolumn{1}{c|}{41.6} &
  \multicolumn{1}{c|}{41.8} &
  $43.0\star$ 
  \\
  \multicolumn{1}{l|}{\textbf{MiniCPM-o 2.6}}&
  \multicolumn{1}{c|}{8B}&
\multicolumn{1}{c|}{67.2} &
    \multicolumn{1}{c|}{68.0} &
  \multicolumn{1}{c|}{68.4} &
  \multicolumn{1}{c|}{$69.7\star$} &
  
    \multicolumn{1}{c|}{56.0} &
  \multicolumn{1}{c|}{56.9} &
    \multicolumn{1}{c|}{57.1} &
  \multicolumn{1}{c|}{$58.4\star$} &
  
  \multicolumn{1}{c|}{42.5} &
    \multicolumn{1}{c|}{43.9} &
  \multicolumn{1}{c|}{43.1} &
  $45.2\star$ 
  \\
  \multicolumn{1}{l|}{\textbf{Aria}}&
  \multicolumn{1}{c|}{3.5B$\times$8}&
\multicolumn{1}{c|}{70.9} &
    \multicolumn{1}{c|}{72.1} &
  \multicolumn{1}{c|}{72.5} &
  \multicolumn{1}{c|}{$73.8\star$} &
  
    \multicolumn{1}{c|}{57.6} &
  \multicolumn{1}{c|}{58.0} &
    \multicolumn{1}{c|}{58.7} &
  \multicolumn{1}{c|}{$59.8\star$} &
  
  \multicolumn{1}{c|}{43.9} &
    \multicolumn{1}{c|}{45.0} &
  \multicolumn{1}{c|}{44.8} &
  $46.3\star$ 
   \\
\multicolumn{1}{l|}{\textbf{GPT-4V}} &
  \multicolumn{1}{c|}{-} &
  \multicolumn{1}{c|}{73.9} &
     \multicolumn{1}{c|}{{\ul 74.7}} &
  \multicolumn{1}{c|}{{\ul75.4}} &
  \multicolumn{1}{c|}{{\ul76.5}$\star$} &
  
  \multicolumn{1}{c|}{57.1} &
     \multicolumn{1}{c|}{59.0} &
  \multicolumn{1}{c|}{58.2} &
  \multicolumn{1}{c|}{{\ul60.6}$\star$} &
  
    \multicolumn{1}{c|}{ 44.7} &
       \multicolumn{1}{c|}{{\ul46.4}} &
  \multicolumn{1}{c|}{45.9} &
  47.5$\star$
\\
\multicolumn{1}{l|}{\textbf{GPT-4o}} &
  \multicolumn{1}{c|}{-} &
  \multicolumn{1}{c|}{{\ul 75.1}} &
     \multicolumn{1}{c|}{\textbf{75.9}} &
  \multicolumn{1}{c|}{\textbf{76.2}} &
  \multicolumn{1}{c|}{\textbf{77.4}$\star$} &
  
  \multicolumn{1}{c|}{{\ul 59.2}} &
    \multicolumn{1}{c|}{\textbf{61.5}} &
  \multicolumn{1}{c|}{\textbf{61.2}} &
  \multicolumn{1}{c|}{\textbf{62.8}$\star$} &
  
  \multicolumn{1}{c|}{{\ul 47.2}} &
     \multicolumn{1}{c|}{\textbf{47.6}} &
  \multicolumn{1}{c|}{\textbf{48.7}} &
  \textbf{50.1}$\star$
\\ \hline
\multicolumn{1}{l|}{\textbf{Human}} &
  \multicolumn{1}{c|}{-} &
  \multicolumn{1}{c|}{\textbf{96.0}}&
  \multicolumn{1}{c|}{-} &
   \multicolumn{1}{c|}{} &
    \multicolumn{1}{c|}{-} &
 \multicolumn{1}{c|}{\textbf{93.0}} &
\multicolumn{1}{c|}{-} & 
 \multicolumn{1}{c|}{} &
  \multicolumn{1}{c|}{-} &
  \multicolumn{1}{c|}{\textbf{100.0}} &
  \multicolumn{1}{c|}{-} &
   \multicolumn{1}{c|}{} &
  - 
\\ \hline
\multicolumn{14}{c}{(b) Punchline Reasoning} \\ \hline
\end{tabular}}
\end{table*}

%% file: Chapter/4_evaluation.tex
\section{Experiments}
\subsection{Baselines}
We include both MLLMs and human baseline for evaluation as follows. 

\noindent\textbf{Evaluated MLLMs.}
We evaluate eight open-source MLLMs (\ie LLaVA~\cite{liu2024llavanext}, GLM-4V~\cite{glm2024chatglm}, Qwen2-VL~\cite{Qwen2VL}, CogVLM2~\cite{hong2024cogvlm2}), LLaVA-OneVision~\cite{li2024llava},
InternVL2.5~\cite{chen2024expanding},
MiniCPM-o 2.6~\cite{yao2024minicpm}, and Aria~\cite{li2024aria}) and two closed-source MLLMs (\ie GPT-4V~\cite{GPT-4V} and GPT-4o~\cite{GPT-4o}).
And we adopt zero-shot, 3-shot (in-context learning) and Chain-of-Thought (CoT) as the baselines for prompting MLLMs.  
A detailed description of these models, their parameter settings, and introduction for in-context learning~\cite{DBLP:conf/nips/BrownMRSKDNSSAA20} and CoT~\cite{DBLP:conf/nips/Wei0SBIXCLZ22} are provided in Appendix~\ref{Appendix:LLMs}. 

\noindent\textbf{Human Baseline.}
To make a comparison with human performance on punchline comprehension, we introduce a human baseline. Specifically, 1) for punchline perception, we first randomly select $100$ instructions for each question format except \textit{Generation QA},
and we then recruit human annotators (three undergraduates outside of the work) to answer the questions guided by the instructions.
Notably, the manually annotated reasoning sentences serve as the performance of human baseline for the \textit{Generation QA}. 
\subsection{Evaluation Metric}
For \textit{Yes/No QA}, \textit{Matching QA} and \textit{Multi-option QA}, we utilize accuracy as the metric. A response is deemed correct when the candidate option (\eg \textit{Yes/No}, \textit{Option A/Option B}, or \textit{A/B/C/D}) mentioned in the response matches the ground truth option. The accuracy is then calculated as the ratio of correct responses to the total number of questions.
For \textit{Generation QA}, where the responses from MLLMs are free-form, we resort to \texttt{gpt-3.5-turbo-0125}\footnote{\url{https://chatgpt.com/}.} to assess whether the response matches the semantics of the annotated reasoning sentence with a binary judgment ``Yes'' or ``No''. Responses marked by "Yes" are considered correct and their ratio serves as the accuracy metric. To ensure the reliability of automatic evaluation, we analyze the correlation between automatic and human assessments. The details provided in the Appendix~\ref{Appendix:human_auto_correlation} demonstrate that the automatic metrics align well with human judgments.

\subsection{Main Results}
The evaluation results of punchline perception and reasoning are presented in Table~\ref{tab:main_res}, and we conclude the following findings from five aspects. 

\noindent\textbf{Overall Performance.}
The evaluated MLLMs exhibit limited capability of punchline comprehension, with the accuracy across different question formats for both punchline perception and reasoning falling below $80\%$ in zero-shot setting. 
As can be seen, the closed-source models consistently surpass the open-source models, where GPT-4o achieves the leading performance among the evaluated MLLMs. Regrettably, GPT-4o still lags substantially behind human-level performance, revealing a substantial gap in punchline comprehension between MLLMs and humans.

\noindent\textbf{Cross-task Performance.} Comparing performance of MLLMs cross the two tasks, we can see that punchline reasoning poses greater challenges than punchline perception, since MLLMs perform worse in punchline reasoning. This disparity is expected, as punchline reasoning demands a deeper understanding to explain why a particular pair contains punchline, rather than simply identifying its presence.
Consequently, punchline reasoning proves to be a more complex task for MLLMs compared to punchline perception.
\\
\noindent\textbf{Cross-question Performance.}
Comparing the results cross question formats within each task, we can observe that there exists a significant variation in performance. The reasons can be two folds. On the one hand, the complexity of the question formats varies inherently. From simplest to most complex, the question formats can be ranked as follows: \textit{Yes/No QA}, \textit{Matching QA}, \textit{Multi-option QA}/\textit{Generation QA}. MLLMs show a noticeable decline in performance as the complexity of the questions increases.
On the other hand, individual models have varying innate strengths and weaknesses across different question formats. For instance, LLaVA exceeds GLM-4V in \textit{Yes/No QA} but falls behind GLM-4V in \textit{Matching QA} for punchline perception task. 
\\ 
\noindent\textbf{Effectiveness of SC-CoQ.}
Compared to the zero-shot setting, both 3-shot and SC-CoQ methods consistently improve performance across all question formats. While CoT method slightly degrades performance in \textit{Yes/No QA} for punchline perception, it enhances performance in other question formats. Notably, SC-CoQ outperforms both 3-shot and CoT approaches corss various question formats, highlighting its superiority. The effectiveness of SC-CoQ is further validated in Section~\ref{sec:synonymous and antisense caption}, where its performance improvements in synonymous and antonymous caption settings are analyzed.

\begin{figure}[h]
    \centering
    \includegraphics[width=\linewidth]{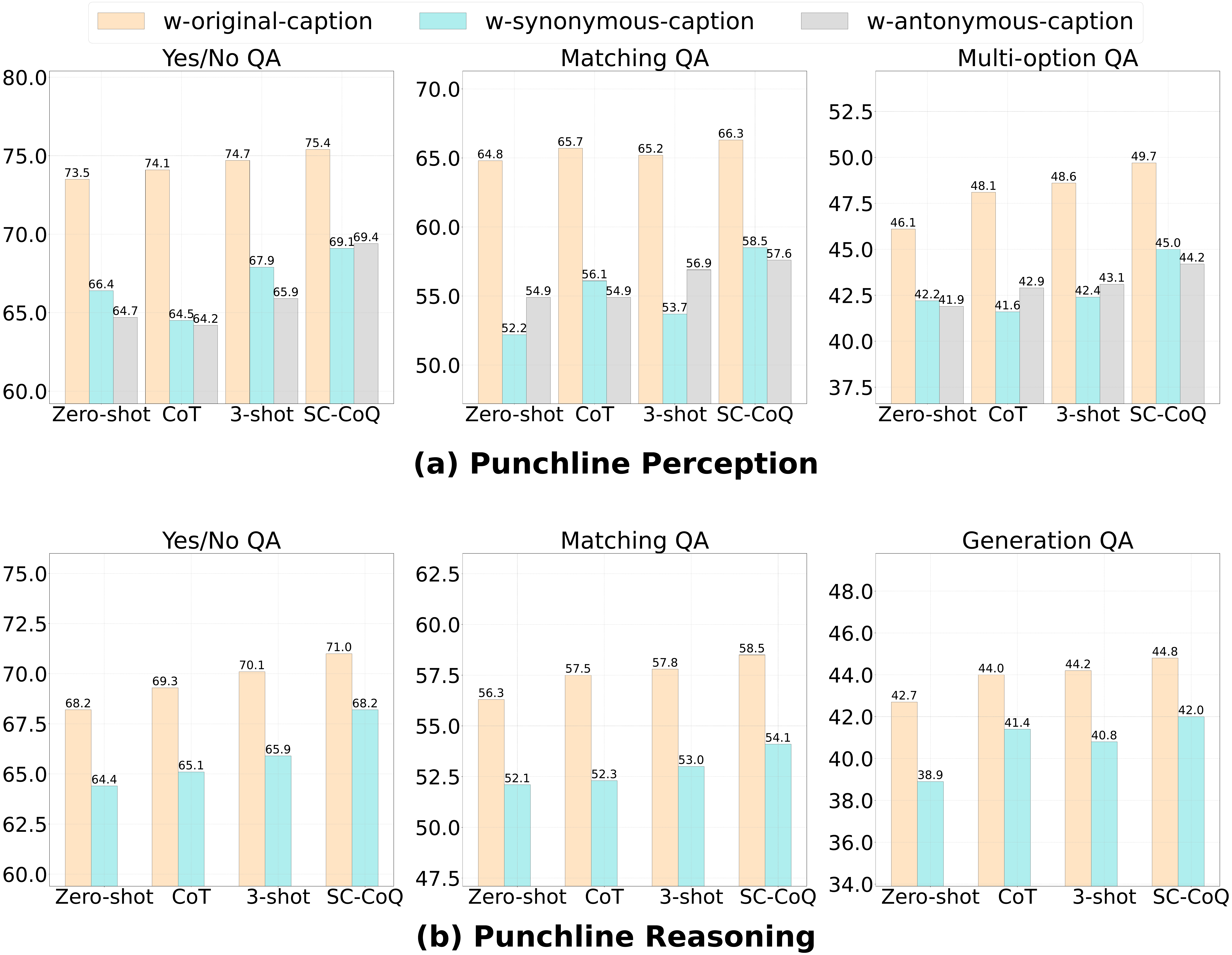}
     \caption{Performance comparison cross original, synonymous and antonymous captions in zero-shot, 3-shot, CoT and our SC-CoQ.}
    \label{Fig:cap}
\end{figure}
\subsection{Effect of Synonymous and Antonymous Captions}
\label{sec:synonymous and antisense caption}
To explore the effect of synonymous and antonymous captions, we compare the performance of CogVLM2 cross the original, synonymous and antonymous captions, as illustrated in Figure~\ref{Fig:cap}.
And the performance comparison for other models are provided in Appendix~\ref{Appendix:Evaluation&Analysis}.
We analyze the results from two perspectives:
1) There is a notable drop in model performance across different question formats when replacing the original caption with synonymous or antonymous captions. It suggests that 
synonymous and antonymous captions effectively successfully eliminate shortcuts found in the original captions and hence challenge models to achieve a thorough comprehension of image-caption pair, which leads to a more comprehensive assessment for punchline comprehension capabilities.
2) When using 3-shot and CoT methods, model performance with synonymous and antonymous captions lags behind that with the original captions. However,
the models show significant improvement across original, synonymous and antonymous captions when applying SC-CoQ. It proves that SC-CoQ can enhance the models' ability to effectively capture the semantics of image-caption pairs and hence achieve better punchline comprehension.

\subsection{Qualitative Analysis}
To provide an intuitive display, we illustrate some testing samples in Figure~\ref{Fig:Q_analysis} for qualitative analysis. Part (a) showcases the responses from two representative models CogVLM2 and GPT-4o in the \textit{Yes/No QA}. Both of them answer correctly when given the original caption, but fail when the original caption is replaced by the synonymous or antonymous caption. This indicates the biases existing in the captions and hence the models may not truly understand the inherent semantics of the image-caption pair to attain the answer. 
And it underscores the significance of introducing synonymous and antonymous captions in assessing punchline comprehension. 
Part (b) exhibits the responses of CogVLM2 with zero-shot, 3-shot, CoT and SC-CoQ for \textit{Multi-option QA}. Notably, with the guidance of SC-CoQ, CogVLM2 successfully answers the question, whereas it fails under the other settings (\ie zero-shot, 3-shot, and CoT). It highlights the effectiveness of SC-CoQ in enhancing punchline comprehension.
More qualitative results for other question formats can be found in Appendix~\ref{subsec:qualitative analysis}.

%% file: Chapter/5_conclusion.tex
\begin{figure}[t]
    \centering
    \includegraphics[width=\textwidth]{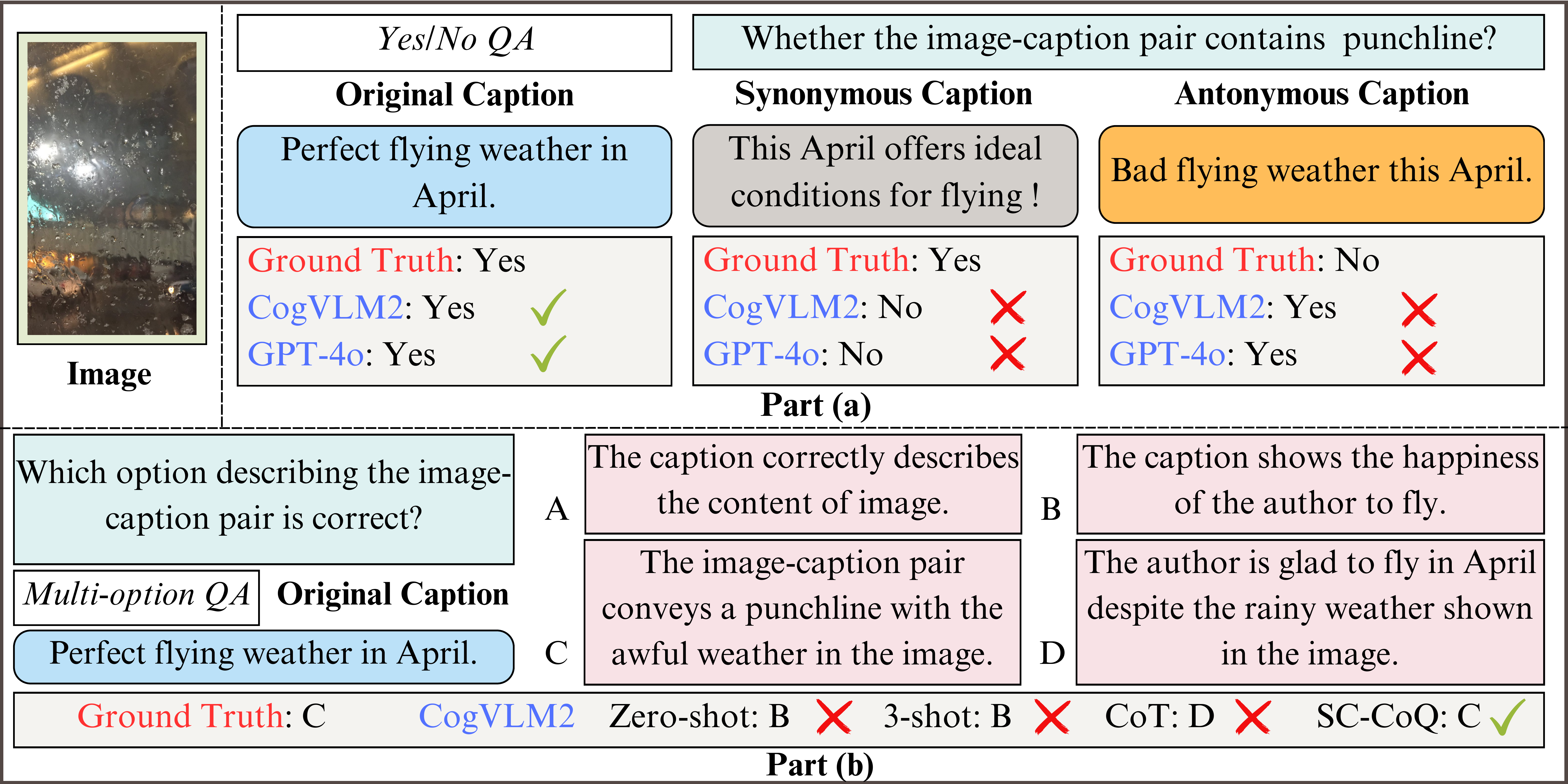}
     \caption{Example responses from CogVLM2 and GPT-4o to the \textit{Yes/No QA} with zero-shot prompts. Responses from CogVLM2 to \textit{Multi-option QA} with different prompting methods are also presented.}
    \label{Fig:Q_analysis}
\end{figure}
\section{Conclusions}
We introduce PunchBench, a benchmark designed to evaluate the ability of MLLMs to comprehend multimodal punchlines. PunchBench distinguish itself from existing benchmarks in two key ways: First, it incorporates synonymous and antonymous captions to mitigate the risk of models relying on shortcuts in the original captions, achieving a more accurate assessment of their capabilities. Second, PunchBench includes a diverse range of punchline types, evaluation tasks, question formats, and multimodal content domains, ensuring a comprehensive evaluation. Our evaluation results highlight a significant gap between the performance of state-of-the-art MLLMs and human capabilities in understanding multimodal punchlines. To address this, we design the Simple-to-Complex Chain-of-Question (SC-CoQ), which effectively enhances the punchline comprehension ability of MLLMs and outperforms widely-used inference-time techniques such as in-context learning and chain-of-thought.

%% file: Chapter/6_appendix.tex
\appendix

\section{More Details for PunchBench}
Here we provide more details for the dataset construction for both punchline perception and reasoning.
\subsection{Source Data Collection \& Annotation}
\label{Appendix:daa_collection}
We detail the data collection process.

\noindent\textbf{Data Collection}.
1) \textbf{Data filtering}.
To reduce time-consuming and labor-cost, we introduce MLLM-based filtering method to answer the above questions to help filter the image-caption pairs. To prevent biases from MLLM, we randomly select a model from the set of evaluated MLLMs as the judge.
It then is required to assess the quality of image-caption pairs by responding the following questions.
Q1: ``Whether it contains possible ethics conflict?'' If No, go to the next question.
Q2: ``Whether the content of image is clearly visible?'' If Yes, go to the next question.
Q3: ``Whether the caption is well-written from the aspects of fluency, length and readability?'' If Yes, this image-caption pair passes the filtering process.
To make sure the filtering quality, we randomly sample $500$ image-caption pairs and then employ three undergraduates outside of this work to answer the above questions. Only $1$ pairs of $500$ fail to pass the manual filtering process, which verifies the reliability of automatic filtering process.
2) \textbf{Crowd voting}. To determine whether a collected image-caption pair contains a punchline, we conducted a crowd voting process using a questionnaire. Participants were asked, “Does the given image-caption pair make you laugh?” and could choose between “Yes” and “No.” Each questionnaire was considered valid if it received more than 10 votes. If one option garnered over 80\% of the votes, it was assigned as the label for the corresponding pair. Notably, for the pairs collected from the prior datasets, we adopted the original labels. Specifically, if the pair is identified as humorous or sarcastic in previous datasets, we regarded it as containing punchline.

\noindent\textbf{Data Annotation}.
To acquire reasoning sentences for particular pairs containing punchline, we employ three human annotators to write reasoning sentence based on the content of image and caption. Specifically, we provide the annotated sarcasm or humor explanations for the pairs existing in the previous datasets, which can be referred to write reasoning sentence. Reasoning sentence must cover the key components in image and caption that convey punchline, and the annotators should state how the interplay between visual content and textual information conveys punchline.

\subsection{Synonymous \& Antonymous Caption}
\label{Appendix:caption}
\begin{figure*}
    \centering
    \includegraphics[width=0.9\textwidth]{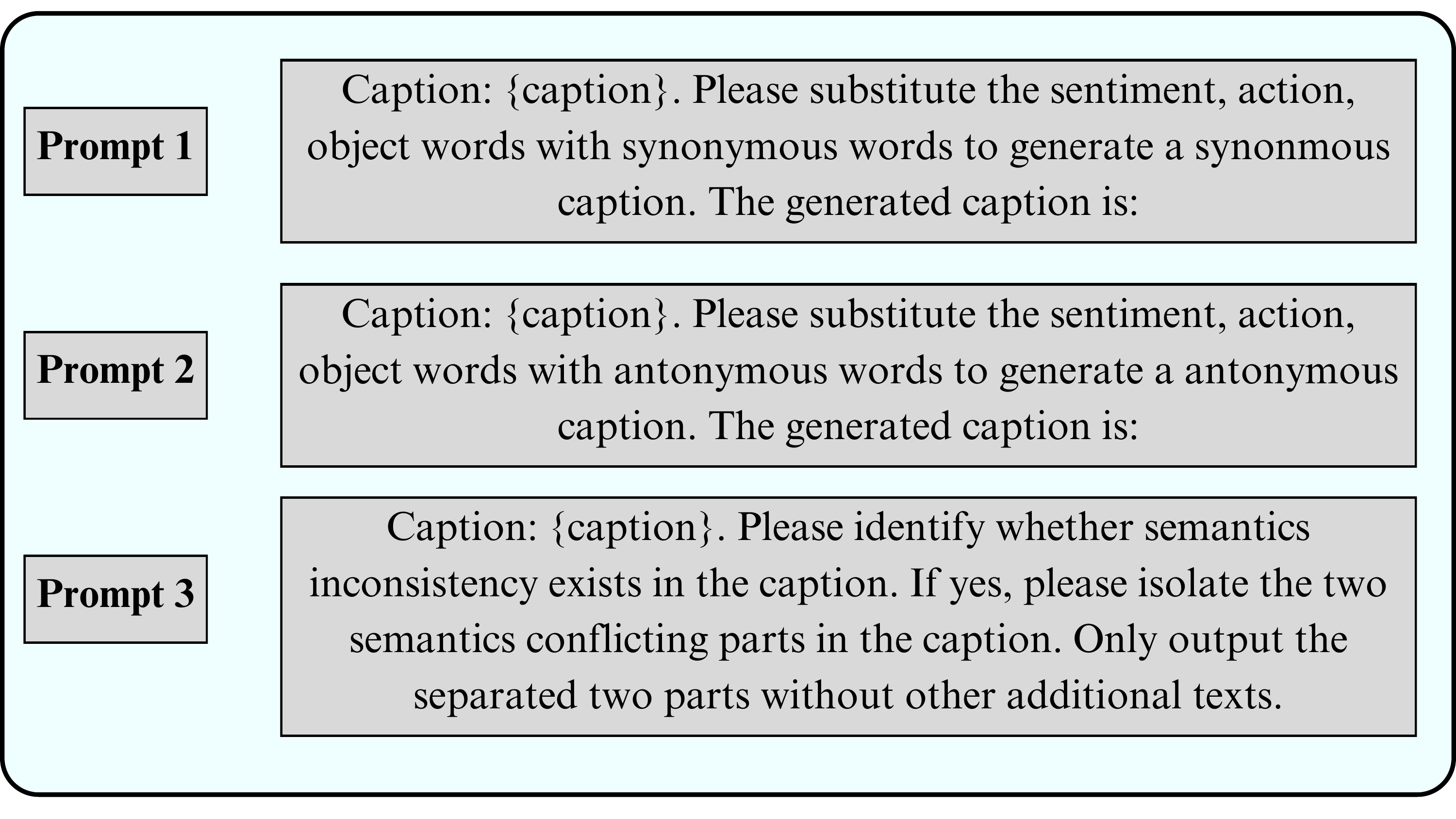}
     \caption{Prompts used to guide \texttt{gpt-3.5-turbo-0125}, where Prompt1 guides the model to generate synonymous caption, Prompt2 guides it to derive antonymous caption, and Prompt3 guides it to identify the context inconsistency.}
    \label{Fig:cap_sa}
\end{figure*}

We illustrate Figure~\ref{Fig:cap_sa} to present the prompts to guide \texttt{gpt-3.5-turbo-0125} to generate synonymous and antonymous captions.
And we provide more implementation details for context consistency adaption as follows. After identifying and isolating the two conflicting parts of inconsistent caption, we adopt word substitution and inversion to derive synonymous and antonymous captions. Specifically, we conduct word substitution for the former part and utilize word inversion for the latter part, if the generated caption maintain the punchline, we regard it as the synonymous caption. And we then conduct word substitution for the latter part and utilize word inversion for the former part, if the generated caption loses the punchline, we regard it as the antonymous caption.

\subsection{Instruction Construction}
\label{Appendix:instruction}
\begin{figure*}
    \centering
    \includegraphics[width=0.9\textwidth]{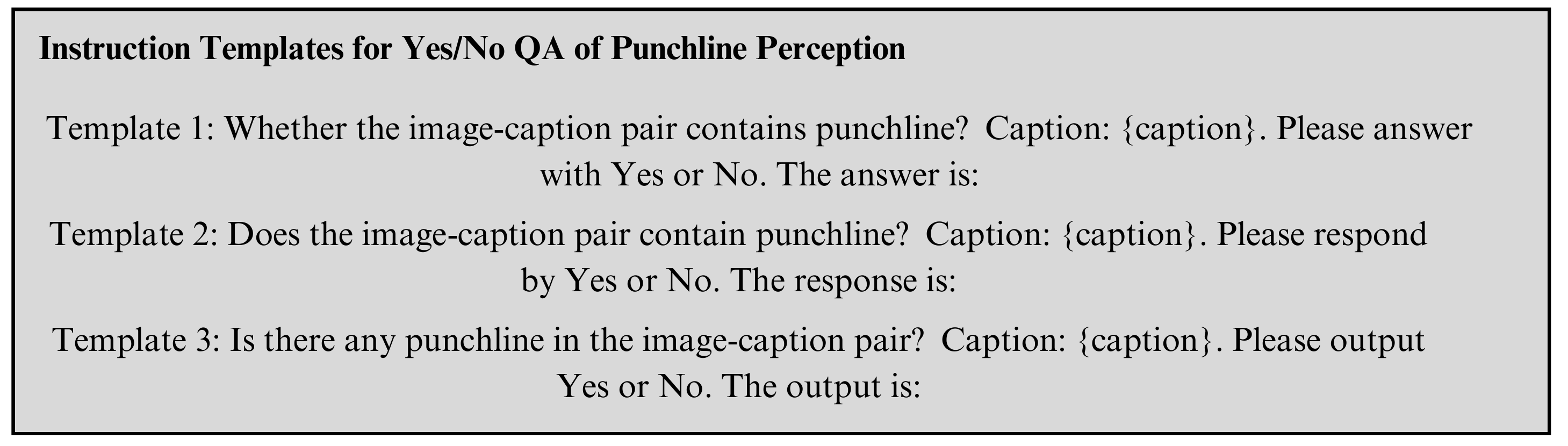}
     \caption{Instruction templates for \textit{Yes/No QA} of punchline perception.}
    \label{Fig:yes_no_1}
\end{figure*}

\begin{figure*}
    \centering
    \includegraphics[width=0.9\textwidth]{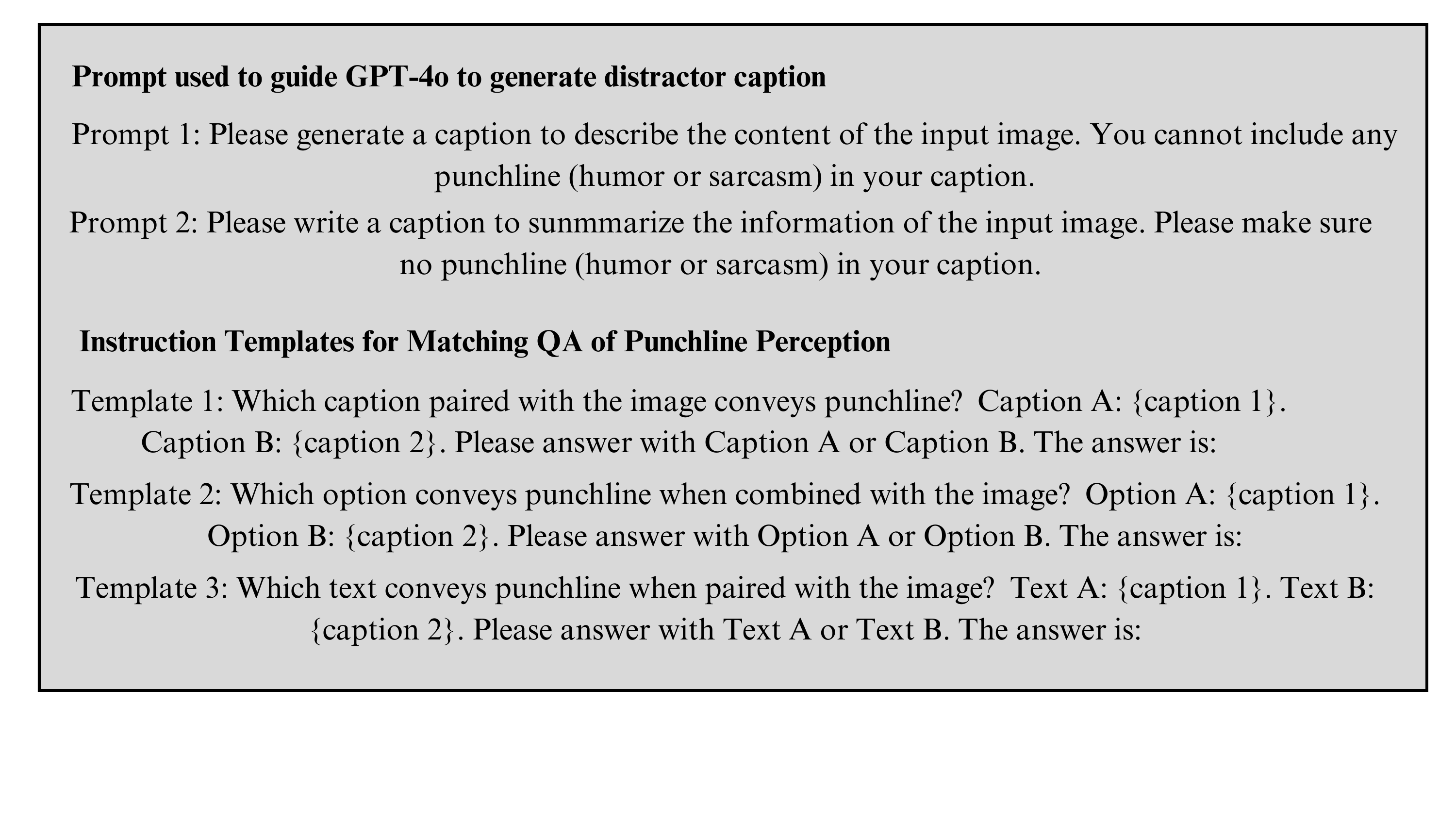}
     \caption{Prompts used to guide GPT-4o to generate distractor caption and instruction templates for \textit{Matching QA} of punchline perception.}
    \label{Fig:matching_1}
\end{figure*}

\begin{figure*}
    \centering
    \includegraphics[width=0.9\textwidth]{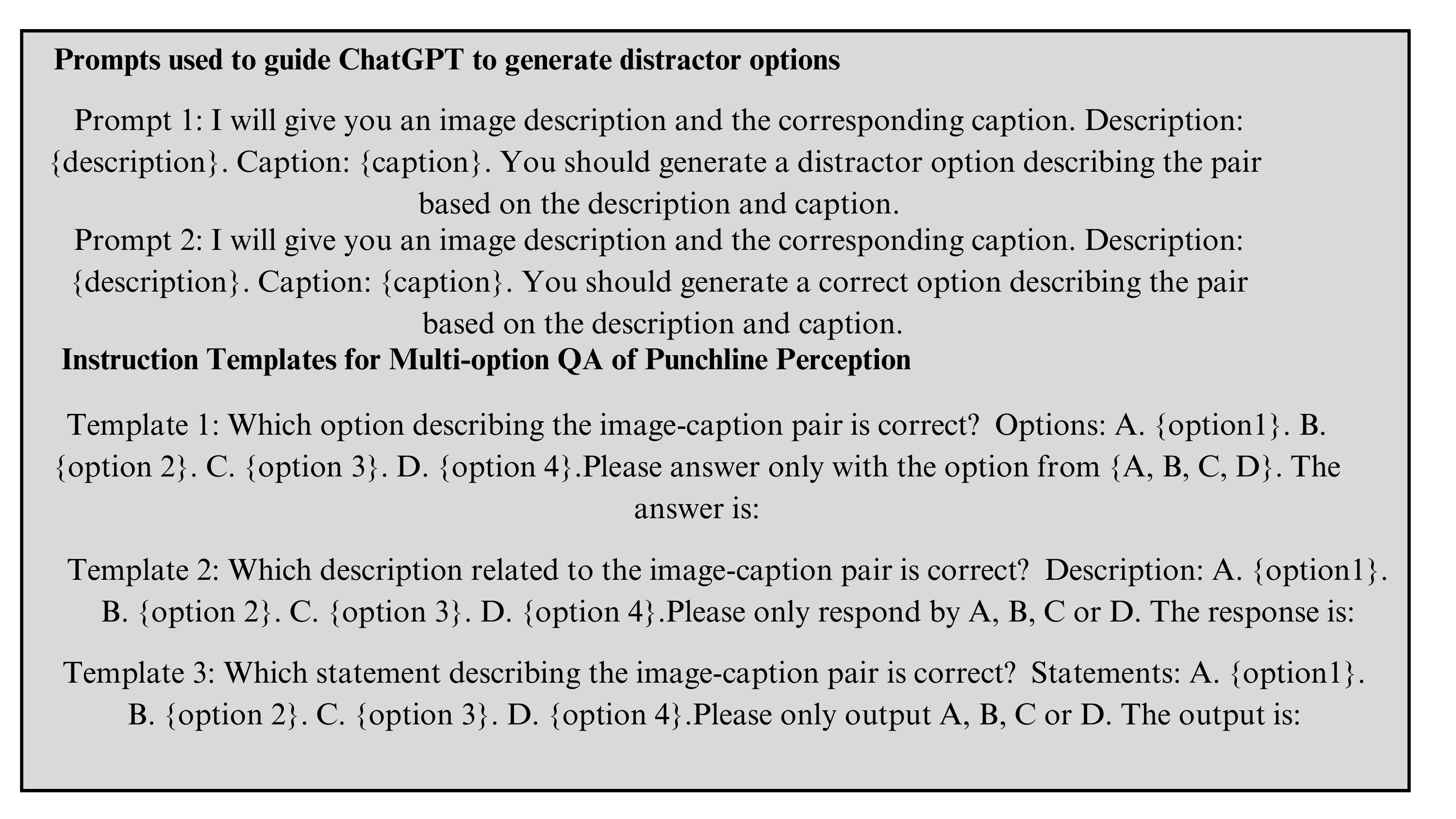}
     \caption{Prompts used to guide GPT-4o to generate distractor options and instruction templates for \textit{Multi-option QA} of punchline perception.}
    \label{Fig:multi_option}
\end{figure*}

\begin{figure*}
    \centering
    \includegraphics[width=0.9\textwidth]{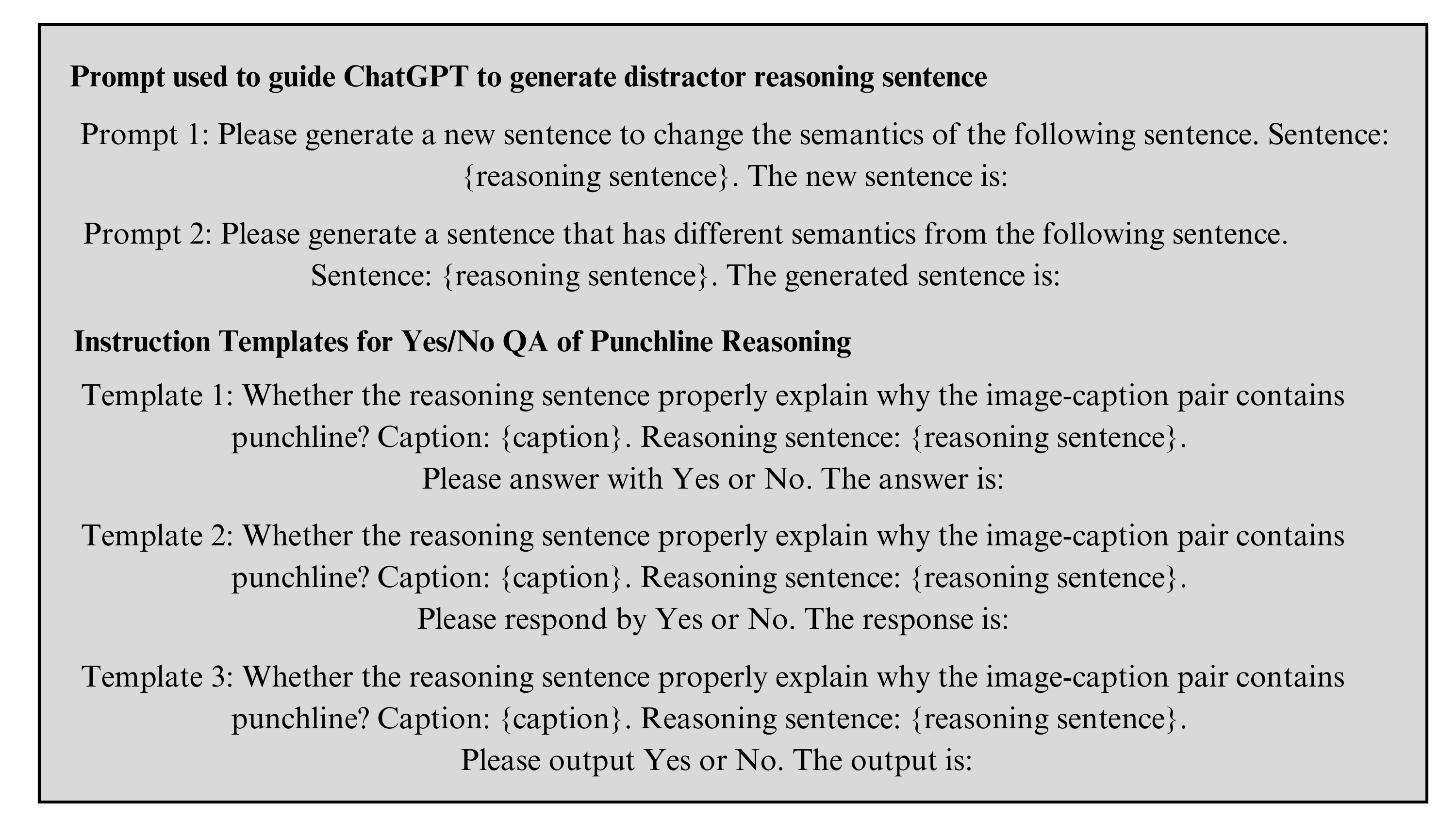}
     \caption{Prompts used to guide ChatGPT to generate distractor reasoning sentence and instruction templates for \textit{Yes/No QA} of punchline reasoning.}
    \label{Fig:yes_no_2}
\end{figure*}

\begin{figure*}
    \centering
    \includegraphics[width=0.9\textwidth]{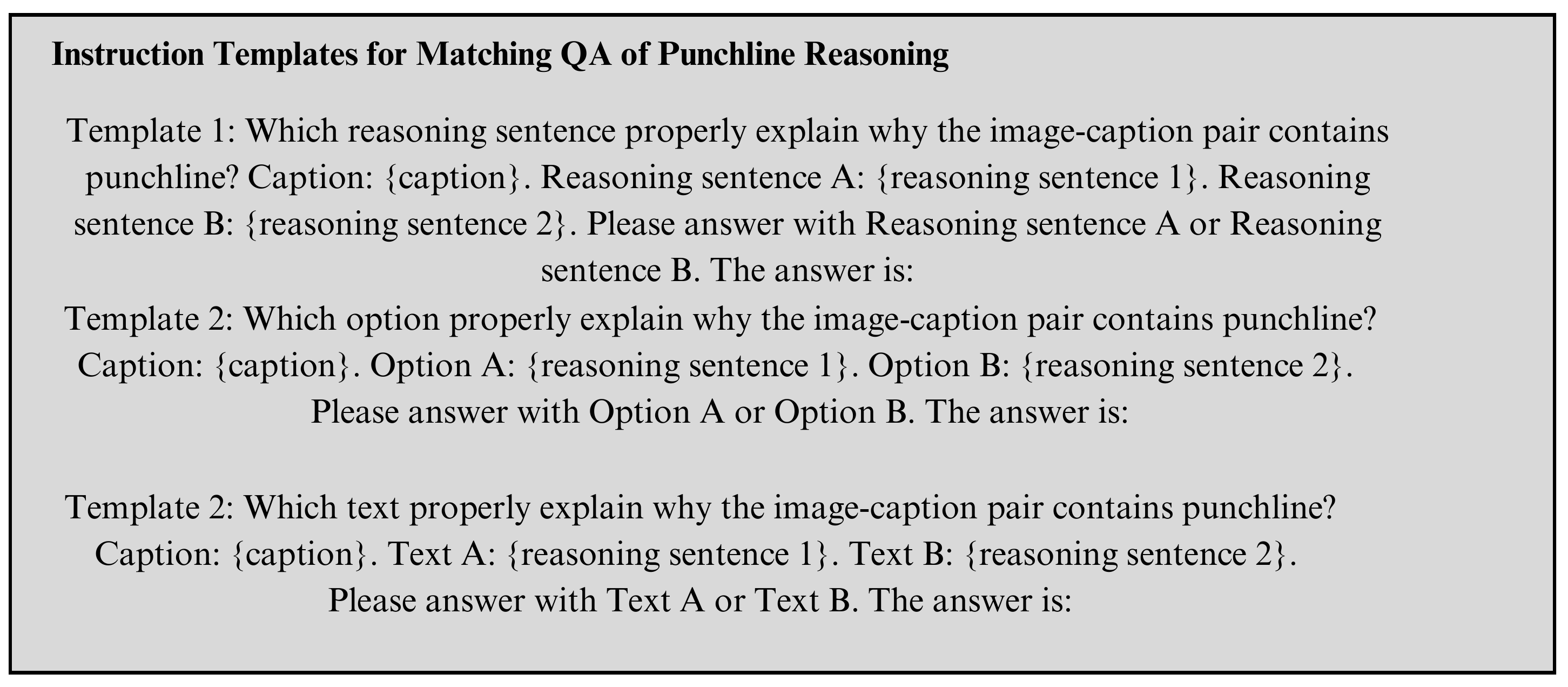}
     \caption{Instruction templates for \textit{Matching QA} of punchline reasoning.}
    \label{Fig:matching_2}
\end{figure*}

\begin{figure*}
    \centering
    \includegraphics[width=0.9\textwidth]{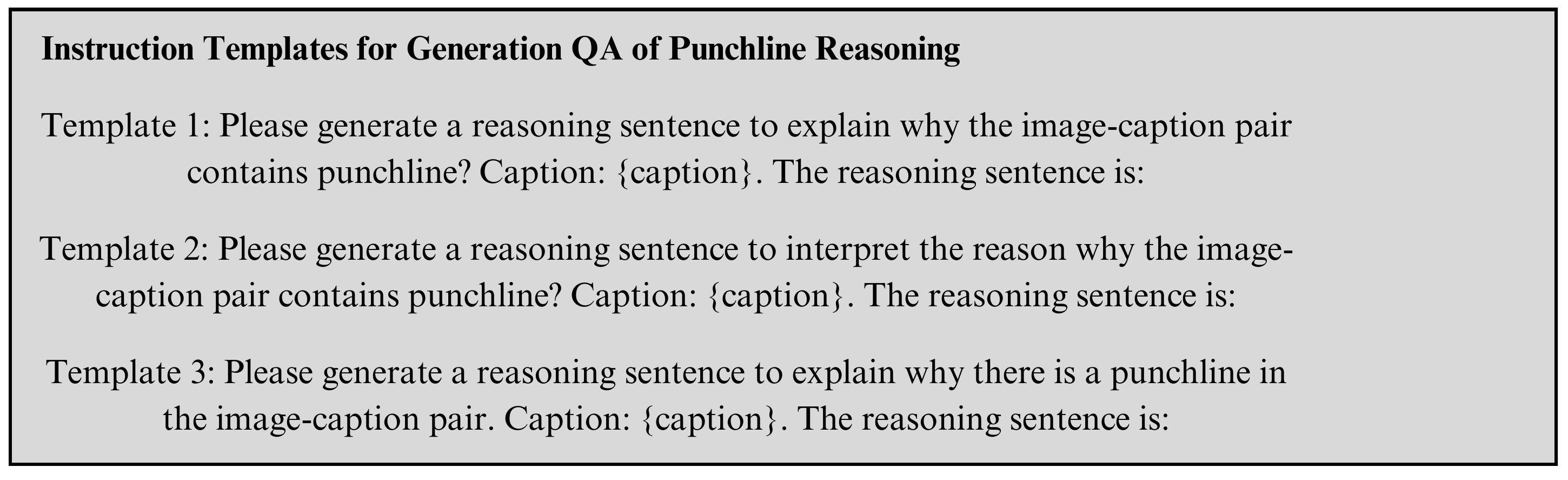}
     \caption{Instruction templates for \textit{Generation QA} of punchline reasoning.}
    \label{Fig:generation}
\end{figure*}

\noindent \textbf{Instruction Template}. We provide various instruction templates for each question format, as follows. For punchline perception, the templates for \textit{Yes/NO QA} are shown in Figure~\ref{Fig:yes_no_1}. The prompts for distractor captions generation and instruction templates for \textit{Matching QA} are exhibited in Figure~\ref{Fig:matching_1}. The prompts for distractor options generation and instruction templates for \textit{Multi-option QA} are exhibited in Figure~\ref{Fig:multi_option}. 
For punchline reasoning, the prompts for distractor reasoning sentence generation and instruction templates for \textit{Yes/No QA} are exhibited in Figure~\ref{Fig:yes_no_2}. The instruction templates for \textit{Matching QA} are exhibited in Figure~\ref{Fig:matching_2}. The instruction templates for \textit{Generation QA} are exhibited in Figure~\ref{Fig:generation}.
\subsection{Benchmark Comparison}
\label{Appendix:statistics}
We compare our PunchBench with the prior benchmarks related to multimodal punchline comprehension in Table~\ref{Tab:Benchmark}.
PunchBench shows superiority in domain, task, question format, punchline type.
\section{More Details for SC-CoQ}
\label{Appendix:SC-CoQ}
\begin{figure*}
    \centering
    \includegraphics[width=0.9\textwidth]{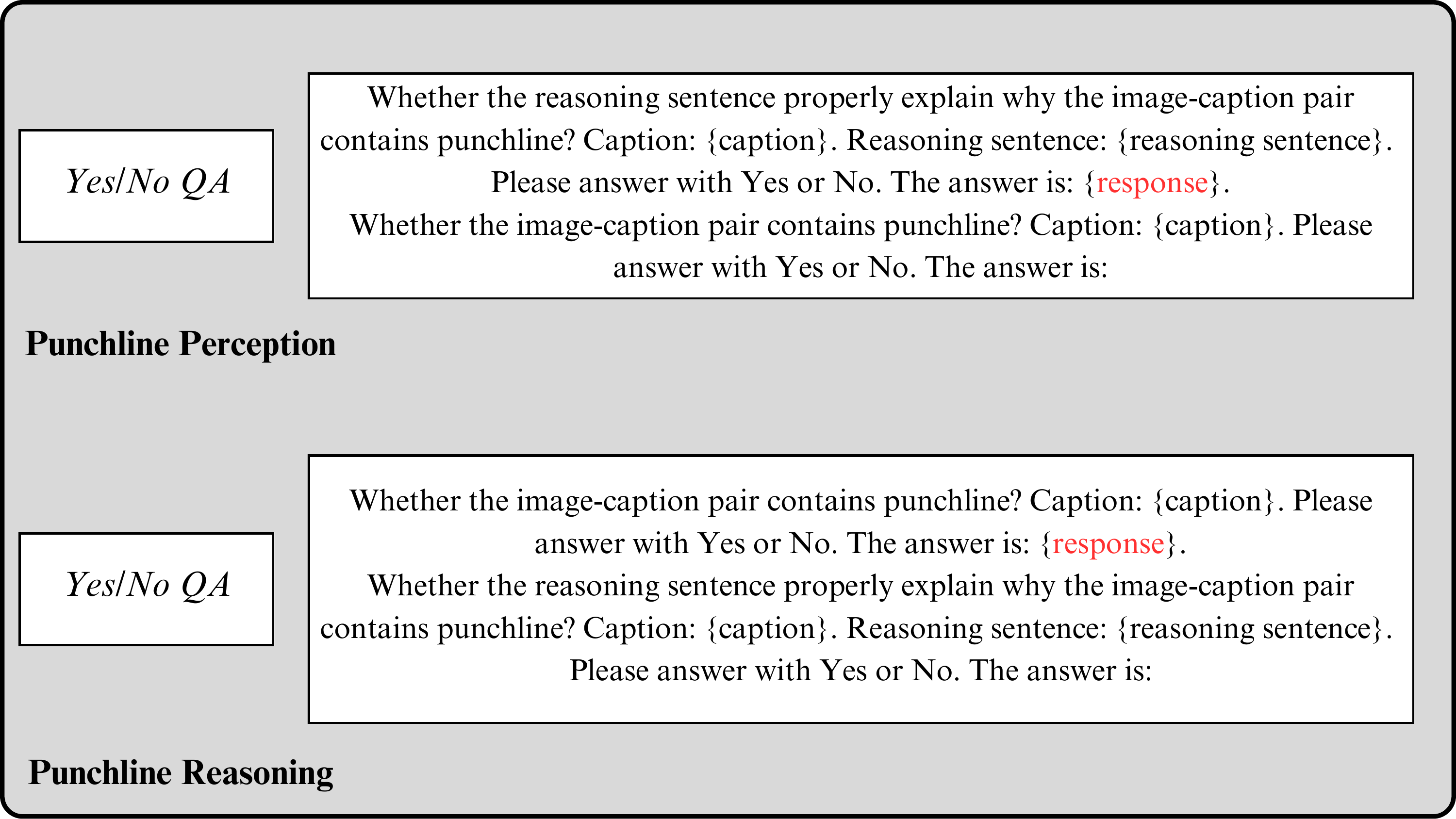}
     \caption{Prompt examples for \textit{Yes/No QA} of punchline perception and reasoning using SC-CoQ.}
    \label{Fig:yes_no_sccoq}
\end{figure*}

\begin{figure*}
    \centering
    \includegraphics[width=0.9\textwidth]{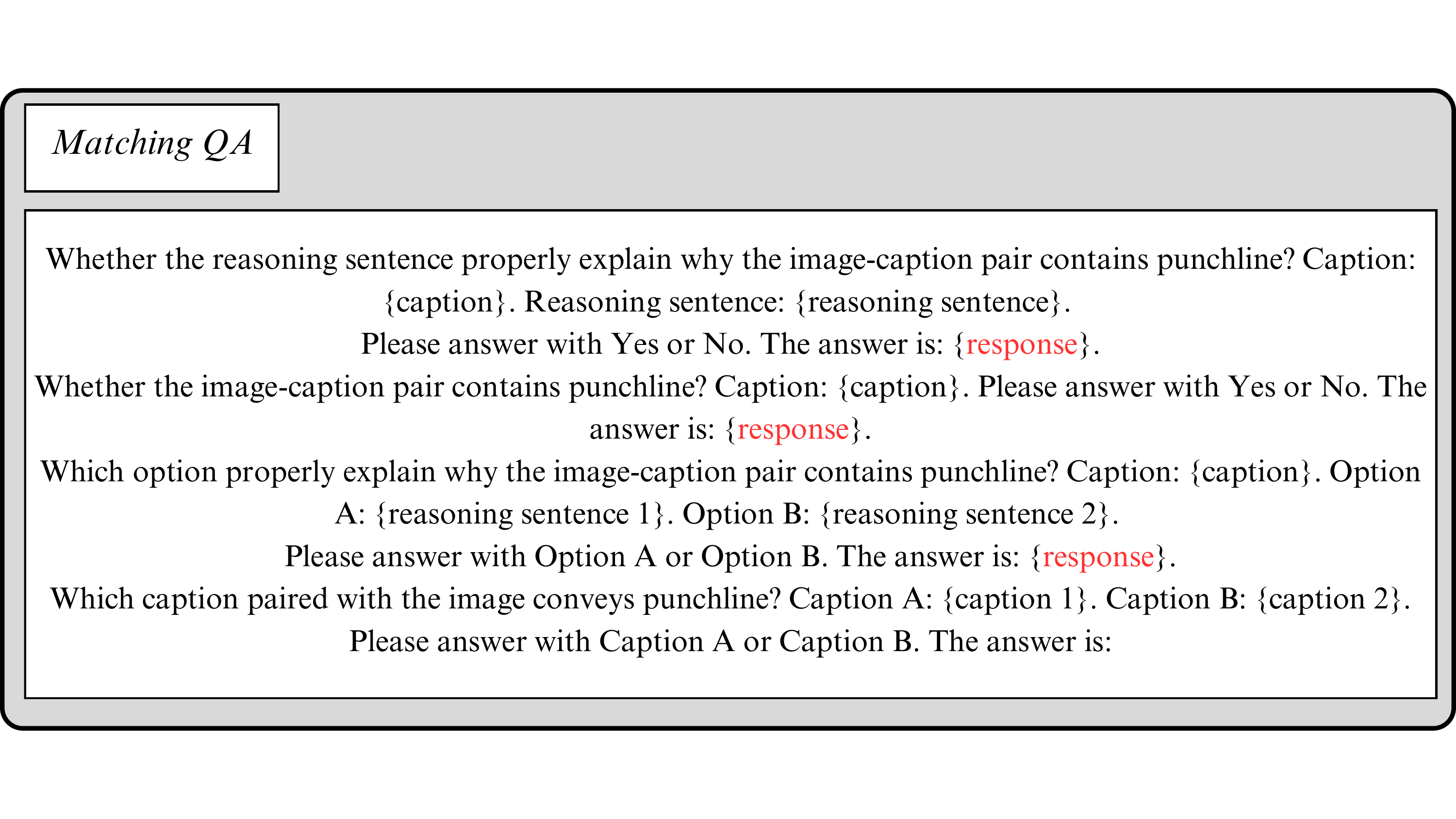}
     \caption{Prompt examples for \textit{Matching QA} of punchline perception using SC-CoQ.}
    \label{Fig:matching1_sccoq}
\end{figure*}

\begin{figure*}
    \centering
    \includegraphics[width=0.9\textwidth]{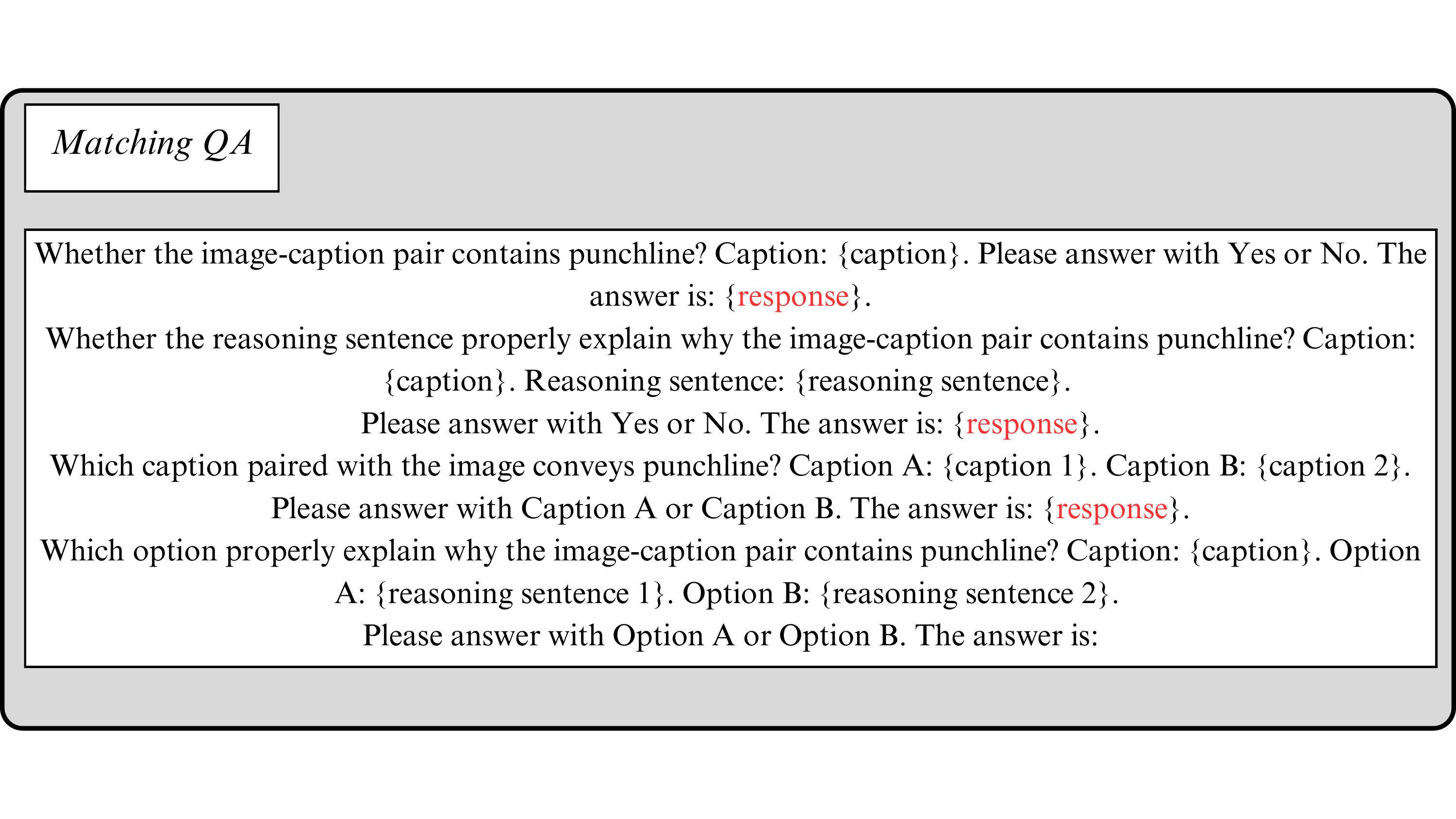}
     \caption{Prompt examples for \textit{Matching QA} of punchline reasoning using SC-CoQ.}
    \label{Fig:matching2_sccoq}
\end{figure*}

\begin{figure*}
    \centering
    \includegraphics[width=0.9\textwidth]{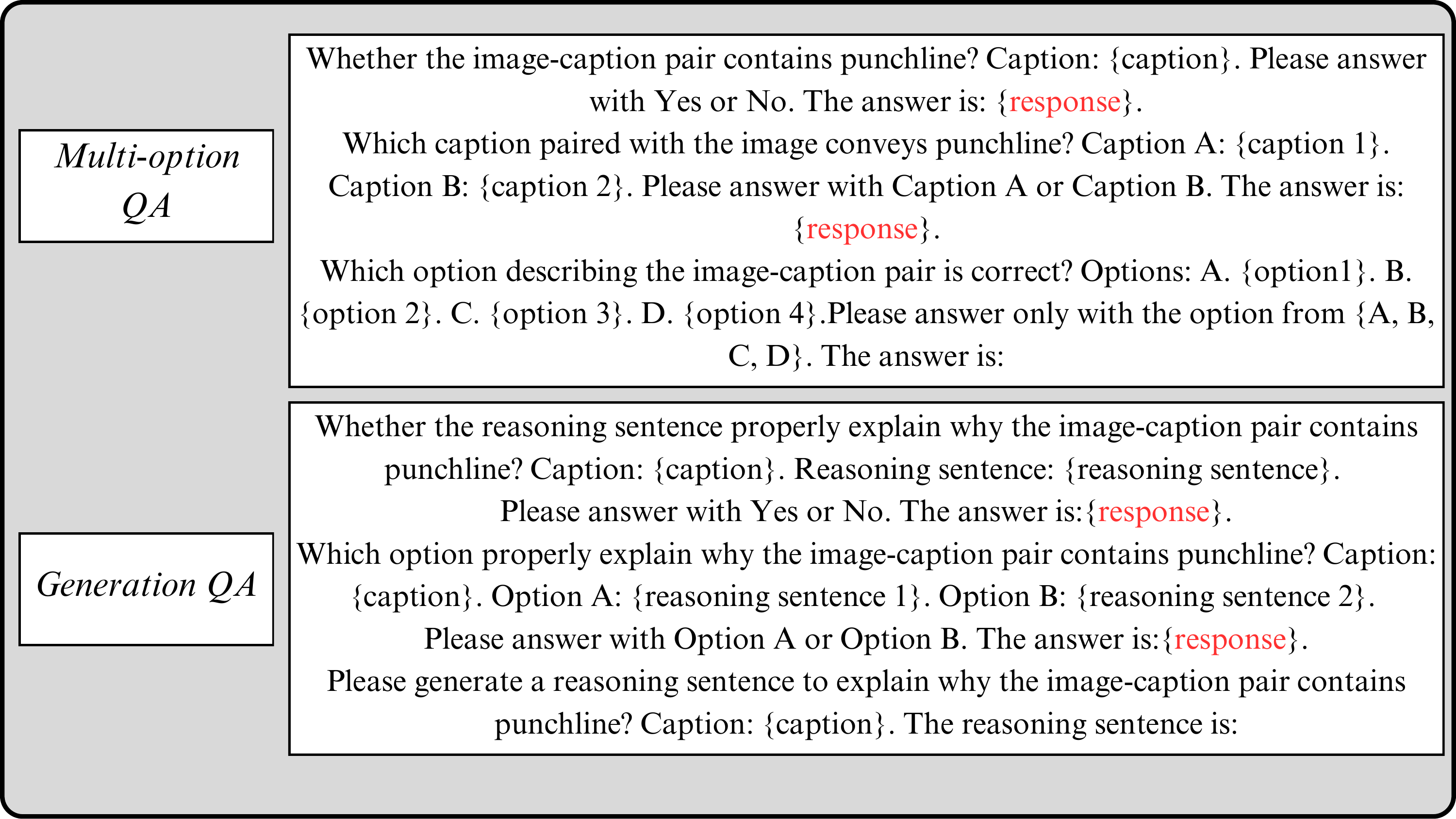}
     \caption{Prompt examples for \textit{Multi-option QA} and \textit{Generation QA} using SC-CoQ.}
    \label{Fig:gen_sccoq}
\end{figure*}

For the simplest question format \textit{Yes/No QA}, we construct Inter-task SC-CoQ, \ie <$\textit{Yes/No QA}_m$, $\textit{Yes/No QA}_n$>, <$\textit{Yes/No QA}_n$, $\textit{Yes/No QA}_m$>. $m$ denotes punchline perception and $n$ means punchline reasoning. Specifically, For a specific $\textit{Yes/No QA}_m$ in punchline perception task, $\textit{Yes/No QA}_n$> is filled by a randomly sampled {Yes/No QA} from punchline reasoning task. For a specific $\textit{Yes/No QA}_n$ in punchline reasoning task, $\textit{Yes/No QA}_m$> is implemented by the {Yes/No QA} from punchline reasoning task which shares the same image-caption pair. Notably, we integrate the response to the former question before the final question in the chain, as shown in n Figure~\ref{Fig:yes_no_sccoq}. Similarly, for \textit{Matching QA}, we adopt the same process. Then we can obtain SC-CoQ for \textit{Yes/No QA} and \textit{Matching QA}.
Additionally, we exhibit some prompt examples of $\textit{Matching QA}$ using SC-CoQ in Figure~\ref{Fig:matching1_sccoq} and Figure~\ref{Fig:matching2_sccoq}. 
For \textit{Multi-option QA} and \textit{Generation QA}, we implement <\textit{Yes/No QA}, \textit{Matching QA}, \textit{Multi-option QA} or \textit{Generation QA}> for a specific image-caption pair.
The prompt examples of $\textit{Multi-option QA}$ are shown in Figure~\ref{Fig:gen_sccoq}.
\section{More Details for Evaluation}
\label{Appendix:LLMs}
\noindent\textbf{Introduction for the MLLMs.}
\begin{itemize}
    \item \textbf{LLaVA}~\cite{liu2024llavanext}.
    We use llava-v1.6-mistral-7b in our experiment. It reuses the pre-trained connector of LLaVA-1.5~\cite{liu2023llava} and adopts Mistral~\cite{Mistral} as the base LLM.
    \item \textbf{GLM-4V}~\cite{glm2024chatglm}. It consists of GLMTransformer with $40$ GLM Blocks and an EVA2CLIP Model with $63$ Transformer Layers, along with a GLU mechanism.
    \item \textbf{Qwen2-VL}~\cite{Qwen2VL}. Qwen2-VL employs a $675M$ parameter ViT across various-sized LLMs, ensuring that the computational load of the ViT remains constant regardless of the scale of the LLM. In terms of language processing, we have opted for the more powerful Qwen2~\cite{DBLP:journals/corr/abs-2407-10671}.
    \item \textbf{CogVLM2}~\cite{hong2024cogvlm2}.
    It is a stronger version of CogVLM, which is an extension of Vicuna, incorporating ViT~\cite{DBLP:conf/iclr/DosovitskiyB0WZ21} as the vision encoder, a two-layer MLP~\cite{DBLP:journals/corr/abs-2002-05202} as adapter, and introducing Visual expert module.
    \item \textbf{LLaVA-OneVision}~\cite{li2024llava}. It integrates the Qwen2~\cite{qwen2.5} language backbone with the SigLIP~\cite{DBLP:conf/iccv/ZhaiM0B23} vision encoder, enhancing performance on tasks that demand fine-grained visual understanding.
    \item \textbf{InternVL2.5}~\cite{chen2024expanding}. This high-performing open-source MLLM integrates InternViT-300M-448px-V2\_5~\cite{chen2024internvl} as the vision encoder and internlm2\_5-7b-chat~\cite{DBLP:journals/corr/abs-2403-17297} as the language model backbone.
    \item \textbf{MiniCPM-o 2.6}~\cite{yao2024minicpm}. The model is built upon SigLIP-400M~\cite{DBLP:conf/iccv/ZhaiM0B23} and Qwen2.5-7B-Instruct~\cite{qwen2.5}, comprising a total of $8$B parameters.
    \item \textbf{Aria}~\cite{li2024aria}. The model features a fine-grained mixture-of-experts (MoE) decoder that activates $3.5$B of its $24.9$B total parameters per token, enabling faster and more efficient training and inference through expert specialization.
    \item \textbf{GPT-4V}~\cite{GPT-4V} and \textbf{GPT-4o}~\cite{GPT-4o}.
    They are the leading MLLMs proposed by OpenAI.
    \end{itemize}

\begin{table}[h]
\caption{Decoding strategy and parameters for the evaluated MLLMs.}
\label{Tab:para}
\centering
\setlength{\tabcolsep}{1mm}{
\begin{tabular}{c|cc}
\hline
\textbf{A}  & \textbf{Strategy}& \textbf{Parameters}\\ \hline
 \textbf{LLaVA}  & Random & $T$=0.7 \\
\textbf{GLM-4V} &Top-$k$  & $k$=3 \\
\textbf{Qwen2-VL} &Top-$p$ & $p$=0.7 \\
\textbf{CogVLM2}& Random & $T$=0.7 \\
\textbf{LLaVA-OneVision}&Greedy &-\\
\textbf{InternVL2.5}&Greedy &-\\
\textbf{MiniCPM-o 2.6}&Greedy&-\\
\textbf{Aria}&Greedy&-\\
\textbf{GPT-4V}& Greedy & -  \\
\textbf{GPT-4o}& Greedy &- \\\hline  
\end{tabular}
}
\end{table}
\noindent\textbf{Inference settings of the MLLMs.}
We present the inference settings, including decoding strategy and parameters of MLLMs
in Table~\ref{Tab:para}.

\noindent\textbf{Introduction for in-conext learning and chain-of-thought.}
1) In-context learning (ICL)~\cite{DBLP:conf/nips/BrownMRSKDNSSAA20}. ICL enables models to perform tasks without explicit parameter updates by conditioning on a sequence of input-output examples, often referred to as a prompt. The model implicitly learns the task by observing these examples within the context, leveraging its pre-trained knowledge to generate predictions for new inputs. In this work, we adopt 3-shot prompt as one of the baselines.
2) Chain-of-Thought (CoT)~\cite{DBLP:conf/nips/Wei0SBIXCLZ22}. CoT prompting encourages models to generate intermediate reasoning steps in natural language, leading to more accurate and interpretable outputs for complex problems. By including step-by-step explanations in the prompt, CoT facilitates the decomposition of multi-step tasks, such as arithmetic, logical reasoning, or commonsense inference, into manageable sub-tasks. This approach significantly improves performance on reasoning-heavy benchmarks and highlights the potential of leveraging language models for tasks requiring structured thought processes.

\section{Evaluation and Analysis}
\subsection{Performance Variations}
\begin{figure}
    \centering
    \includegraphics[width=\linewidth]{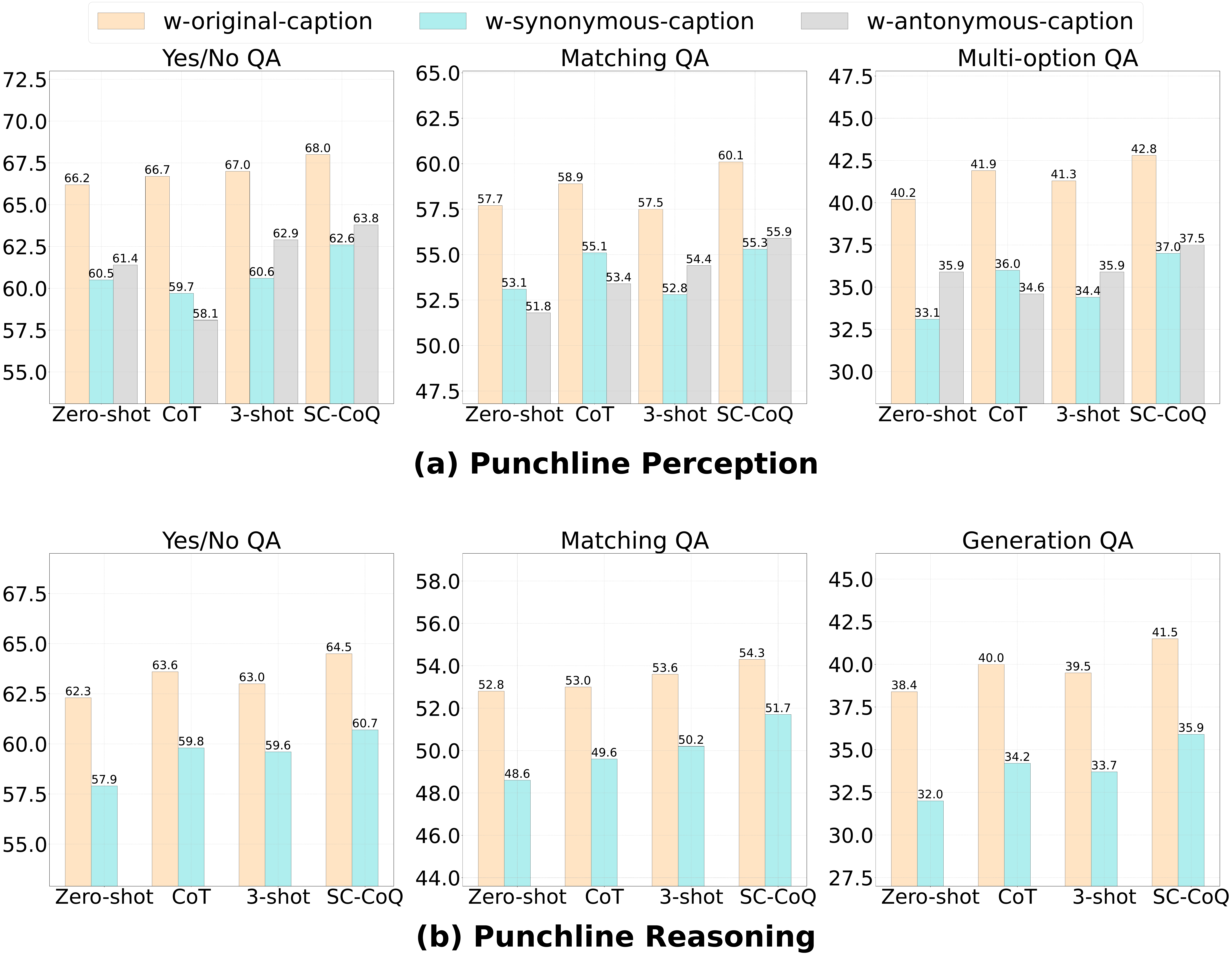}
     \caption{Performance comparison for LLaVA across original, synonymous and antonymous captions in zero-shot, 3-shot, CoT and our SC-CoQ.}
    \label{Fig:llava}
\end{figure}

\begin{figure}
    \centering
    \includegraphics[width=\linewidth]{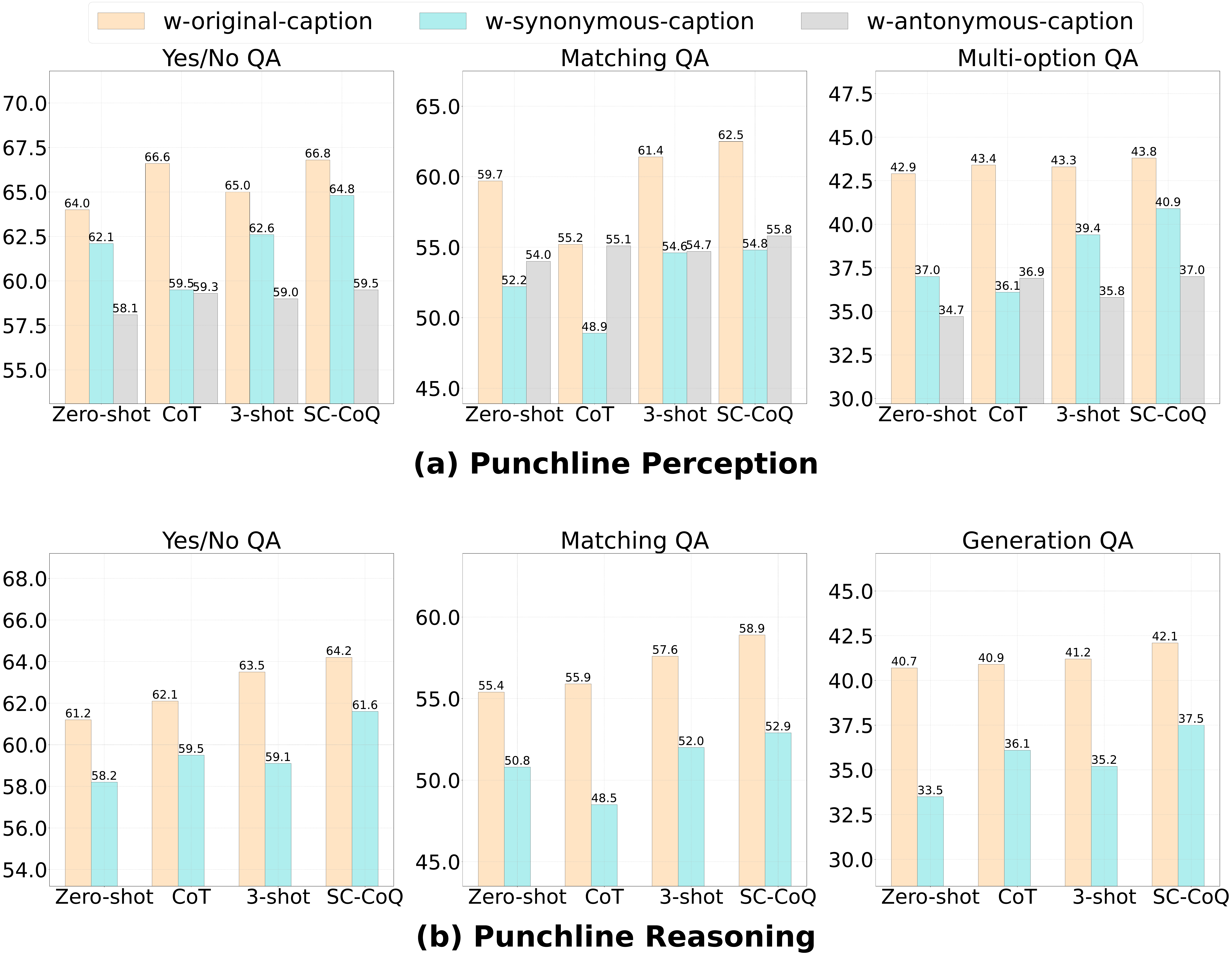}
     \caption{Performance comparison for GLM-4V across original, synonymous and antonymous captions in zero-shot, 3-shot, CoT and our SC-CoQ.}
    \label{Fig:glm4v}
\end{figure}

\begin{figure}
    \centering
    \includegraphics[width=\linewidth]{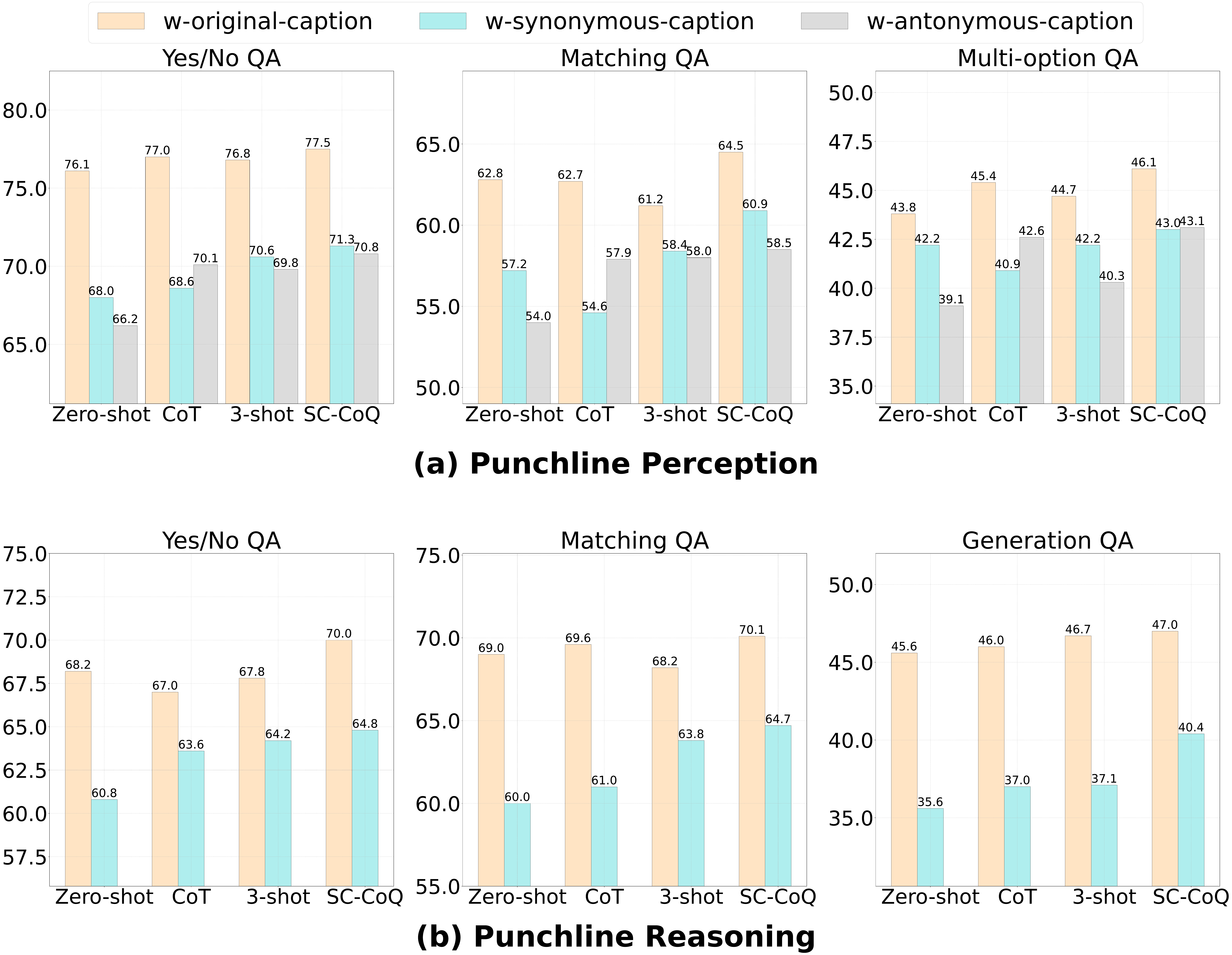}
     \caption{Performance comparison for Qwen2-VL across original, synonymous and antonymous captions in zero-shot, 3-shot, CoT and our SC-CoQ.}
    \label{Fig:qwen2vl}
\end{figure}

\begin{figure}
    \centering
    \includegraphics[width=\linewidth]{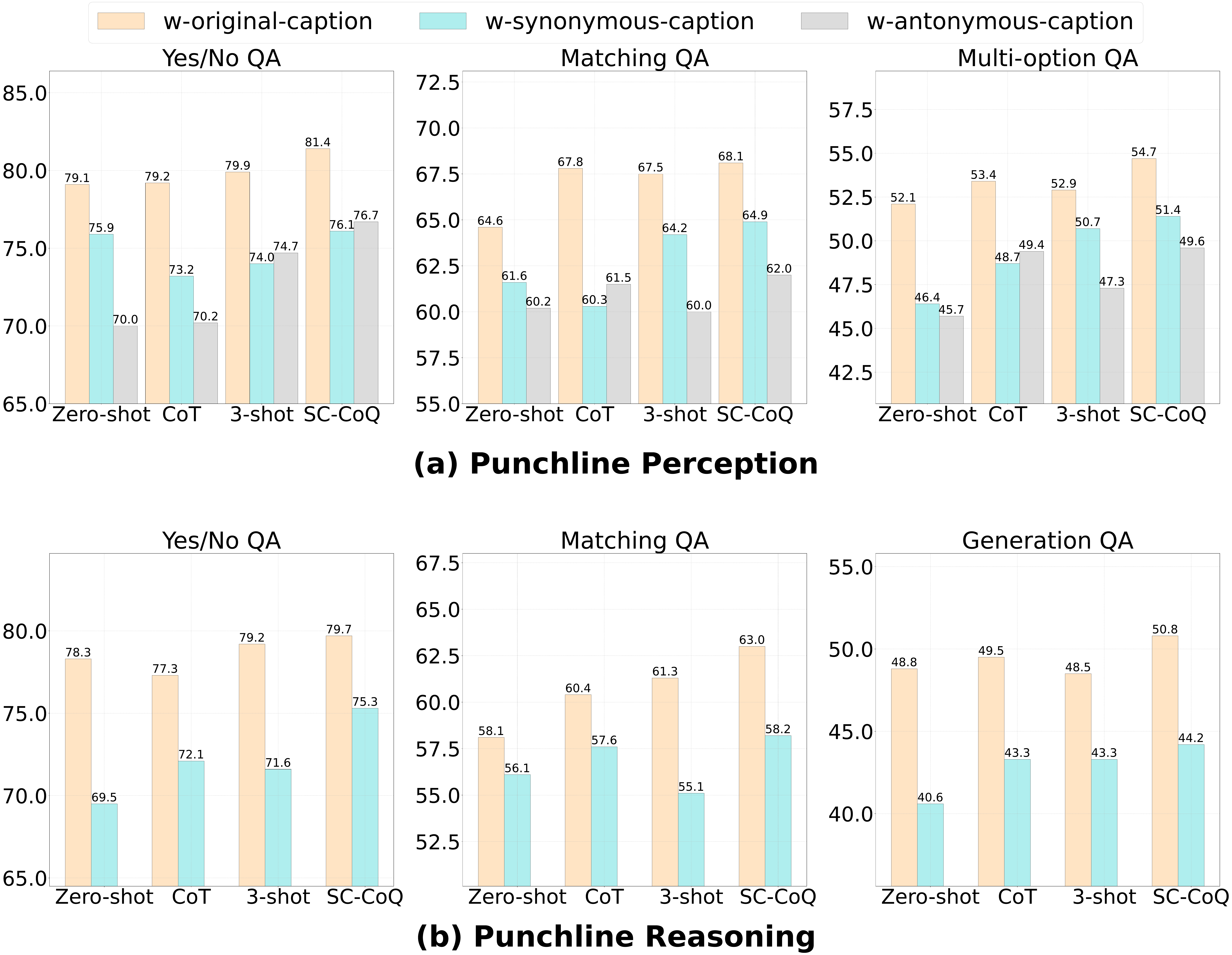}
     \caption{Performance comparison for GPT-4V across original, synonymous and antonymous captions in zero-shot, 3-shot, CoT and our SC-CoQ.}
    \label{Fig:gpt4v}
\end{figure}

\begin{figure}
    \centering
    \includegraphics[width=\linewidth]{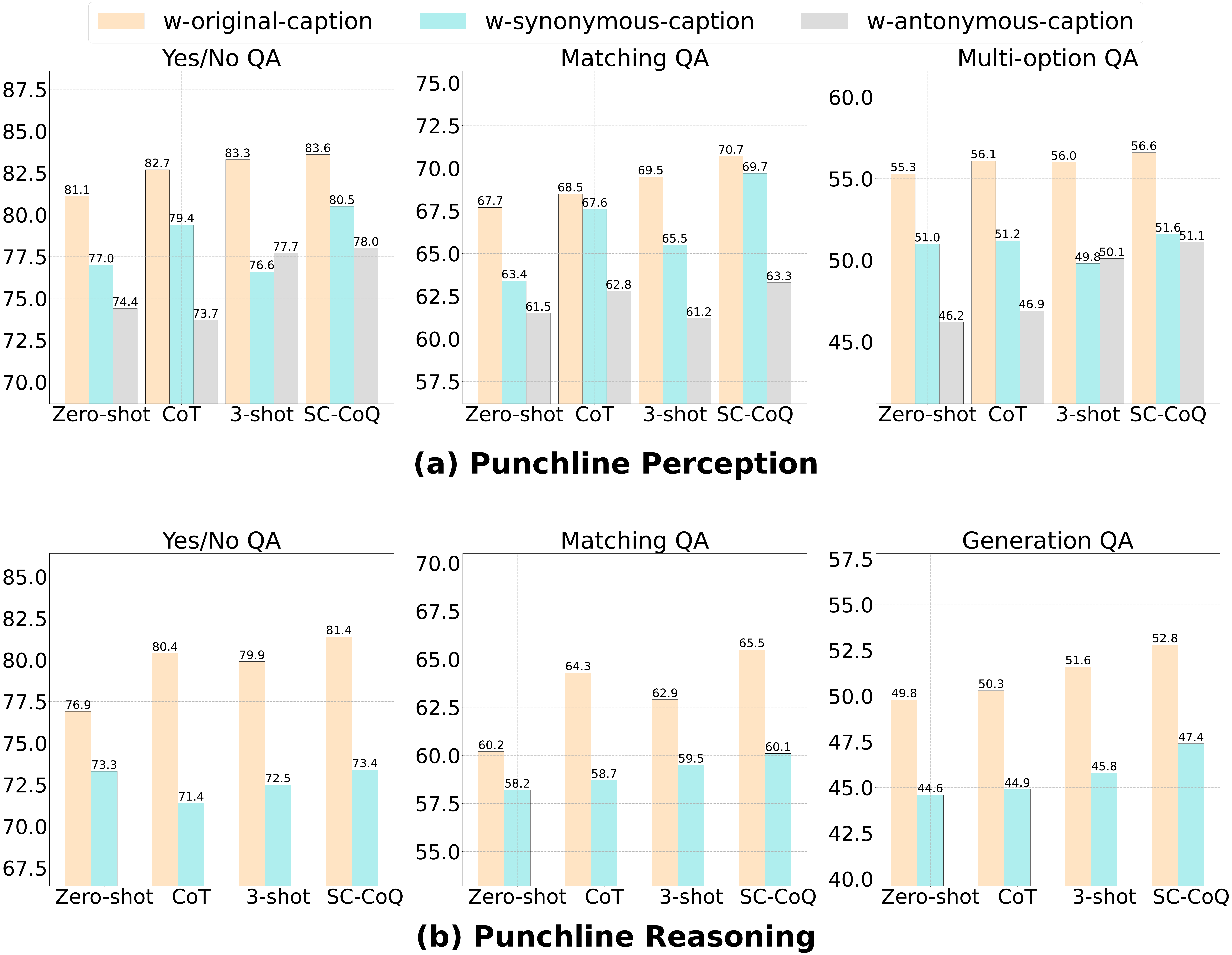}
     \caption{Performance comparison for GPT-4o across original, synonymous and antonymous captions in zero-shot, 3-shot, CoT and our SC-CoQ.}
    \label{Fig:gpt4o}
\end{figure}

We compare the results cross the original, synonymous, and antonymous captions for all the evaluated MLLMs. The results for LLaVA, GLM-4V, Qwen2-VL, GPT-4V and GPT-4o cross different captions are exhibited in Figure~\ref{Fig:llava}, Figure~\ref{Fig:glm4v}, Figure~\ref{Fig:qwen2vl}, Figure~\ref{Fig:gpt4v}, and Figure~\ref{Fig:gpt4o}. As can be seen, synonymous and antonymous captions effectively eliminate shortcuts in the original captions, challenging models to fully comprehend the image-caption pairs. This leads to a more comprehensive evaluation of punchline comprehension capabilities. When using 3-shot and CoT methods, model performance with synonymous and antonymous captions lags behind that with original captions. However, when applying SC-CoQ, models show significant improvement across all caption types. This demonstrates that SC-CoQ enhances the models' ability to grasp the semantics of image-caption pairs, leading to better punchline comprehension.
\begin{table*}[h]
\caption{Human estimation for \textit{Generation QA}. Inter-annotator agreement is emphasized by Gwet's $\gamma$~\cite{Gwet2014HandbookOI}, which is consistently larger than $70.0\%$, indicating substantial agreement.}
\label{Tab:human_eva}
\centering
\resizebox{\textwidth}{!}{
\begin{tabular}{cc|cccc}
\hline
\textbf{A} &\textbf{B} & \textbf{A Wins (\%)}& \textbf{A Draws B (\%)} & \textbf{B Wins (\%)}& \textbf{G-$\gamma$ (\%)}  \\ \hline
\textbf{GLM-4V}  &\textbf{Llava}  & 57.0 & 18.0 & 25.0  & 82.6 \\
\textbf{Qwen2-VL}  &\textbf{GLM-4V} &67.0  & 23.0 &10.0 & 77.4 \\
\textbf{CogVLM2}  &\textbf{Qwen2-VL} &41.0 & 37.0 &22.0  & 80.4 \\
\textbf{GPT-4V}  &\textbf{CogVLM2}& 59.0 & 20.0 & 21.0 & 78.1 \\
\textbf{GPT-4o}  &\textbf{GPT-4V}& 47.0 & 30.0 &23.0 & 74.6 \\\hline  
\textbf{GPT-4o (CoT)}  &\textbf{GPT-4o (Zero-shot)}& 31.0  &48.0  &21.0 & 71.2
\\
\textbf{GPT-4o (3-shot)}  &\textbf{GPT-4o (CoT)}&39.0 &38.0  &23.0 &76.3  \\
\textbf{GPT-4o (SC-CoQ)}  &\textbf{GPT-4o (3-shot)}&46.0  &32.0 &22.0 &82.7  \\\hline  
\end{tabular}
}
\end{table*}
\subsection{Human Evaluation}
\label{Appendix:human_evaluation}
To validate the reliability of automatic evaluation for \textit{Generation QA}, we conduct human evaluation through pairwise test.
Specifically, we first randomly sample $100$ pairs of reasoning sentences from two candidate models.
And we then involve three independent annotators (undergraduate students uninvolved in this work) to compare reasoning sentences generated by two models (A and B) for the same image-caption pair.
The annotators are supposed to choose one of three options: \ie ``A Wins'', ``A Draws B'' and ``B Wins''. Finally, the winner is determined by the ``Win'' votes. If both models receive an equal number of ``Win'' votes, the final result is recorded as ``A Draws B''. In addition, we calculate Gwet's $\gamma$~\cite{Gwet2014HandbookOI} to represent inter-annotator agreement.
The results for human evaluation of the generated reasoning sentences from evaluated models are shown in Table~\ref{Tab:human_eva}.
\begin{figure}[h]
    \centering
    \includegraphics[width=\textwidth]{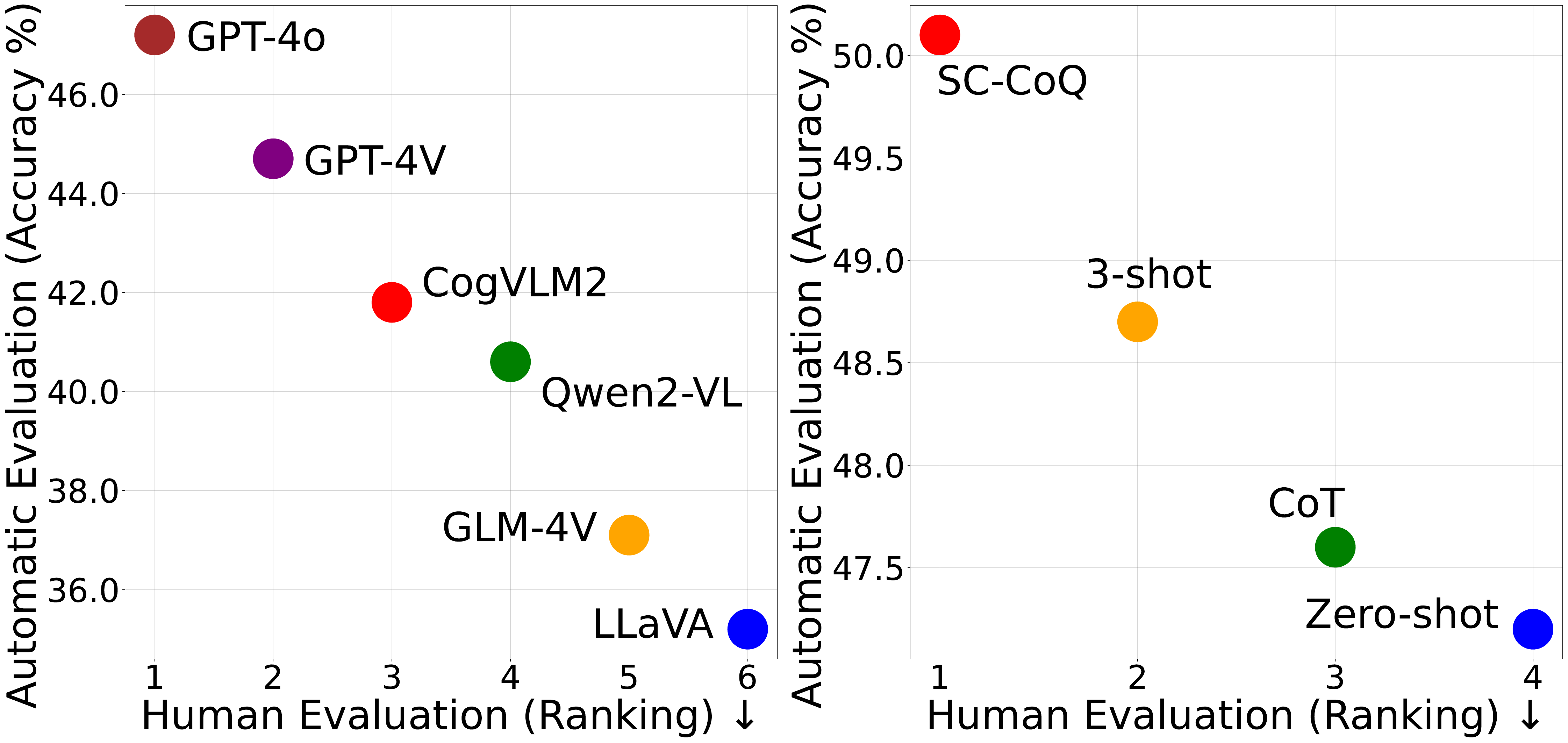}
     \caption{We show the relation between accuracy of automatic evaluation and ranking of human evaluation for evaluated MLLMs and different prompting methods.}
    \label{Fig:correlation}
\end{figure}
\subsection{Correlation between Automatic and Human Evaluation}
\label{Appendix:human_auto_correlation}
Human evaluation results, which are presented in Appendix~\ref{Appendix:human_evaluation}, show substantial agreement among annotators since Gwet's $\gamma$~\cite{Gwet2014HandbookOI} is consistently larger than $70\%$. And we exhibit the correlation between Automatic and Human evaluation in Figure~\ref{Fig:correlation} to emphasize the reliability of automatic evaluation for \textit{Generation QA}. As observed, the models or methods that rank higher in human evaluation also show better accuracy in automatic evaluation. And our SC-CoQ achieves the best performance in both automatic and human evaluation.
It not only verifies the credibility of the automatic evaluation results, but also further demonstrates the advantages of our SC-CoQ. 

\label{Appendix:Evaluation&Analysis}
\subsection{More Qualitative Results}
\begin{figure}
    \centering
    \includegraphics[width=\textwidth]{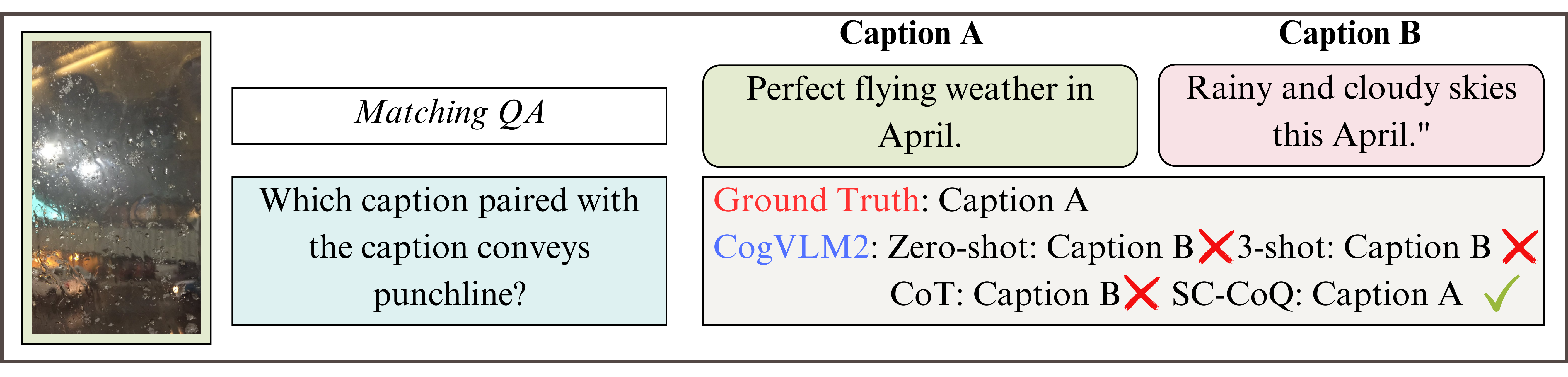}
     \caption{An example for qualitative analysis, where we show the responses from CogVLM2 to the \textit{Matching QA} with different settings (\ie zero-shot, 3-shot, CoT and SC-CoQ).}
    \label{Fig:Q_analysis_matching1}
\end{figure}

\begin{figure}
    \centering
    \includegraphics[width=\textwidth]{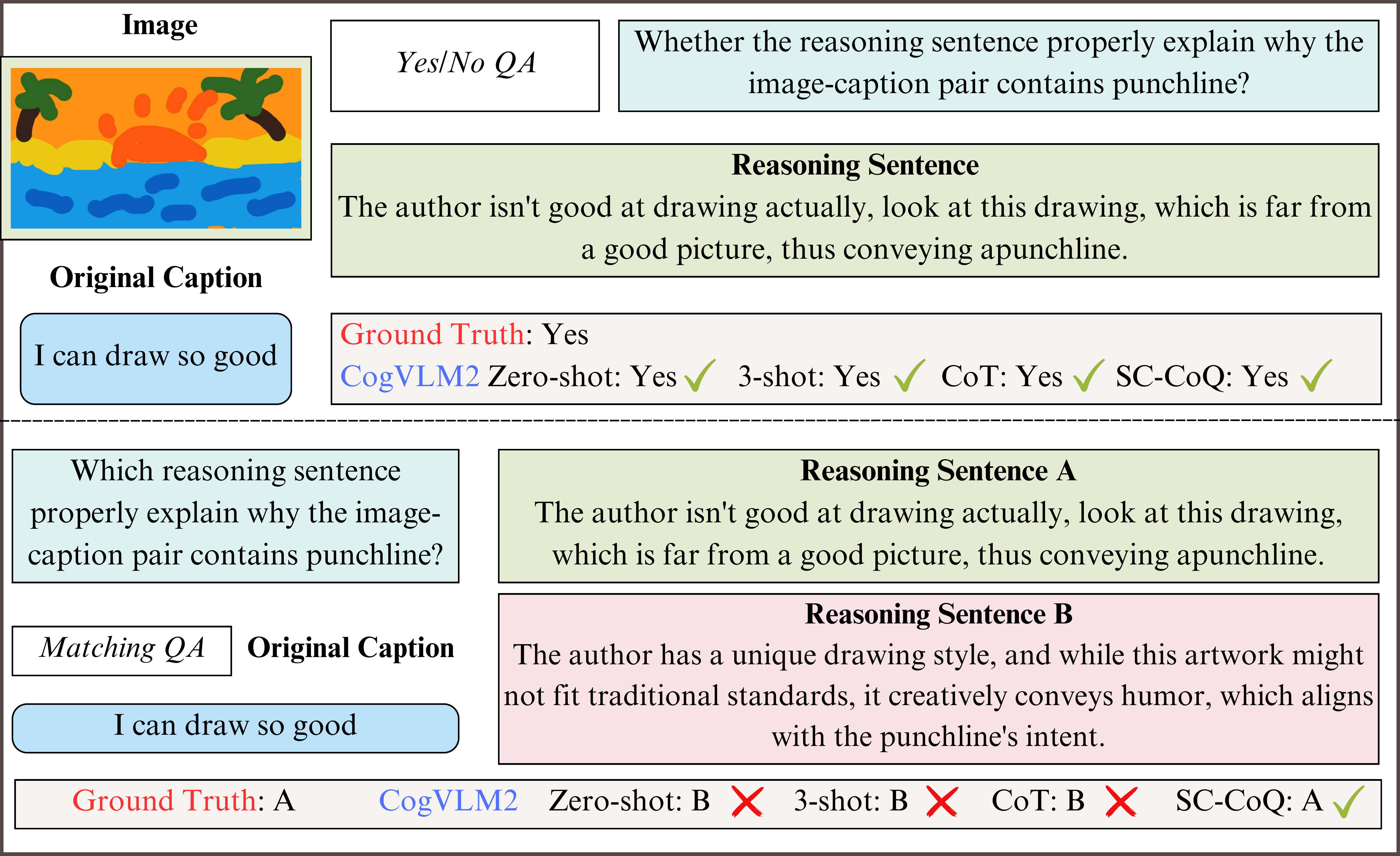}
     \caption{An example for qualitative analysis, where we show the responses from CogVLM2 to the \textit{Yes/No QA} and \textit{Matching QA} of punchline reasoning with different settings (\ie zero-shot, 3-shot, CoT and SC-CoQ).}
    \label{Fig:Q_analysis_reasoning}
\end{figure}

\begin{figure*}
    \centering
    \includegraphics[width=\textwidth]{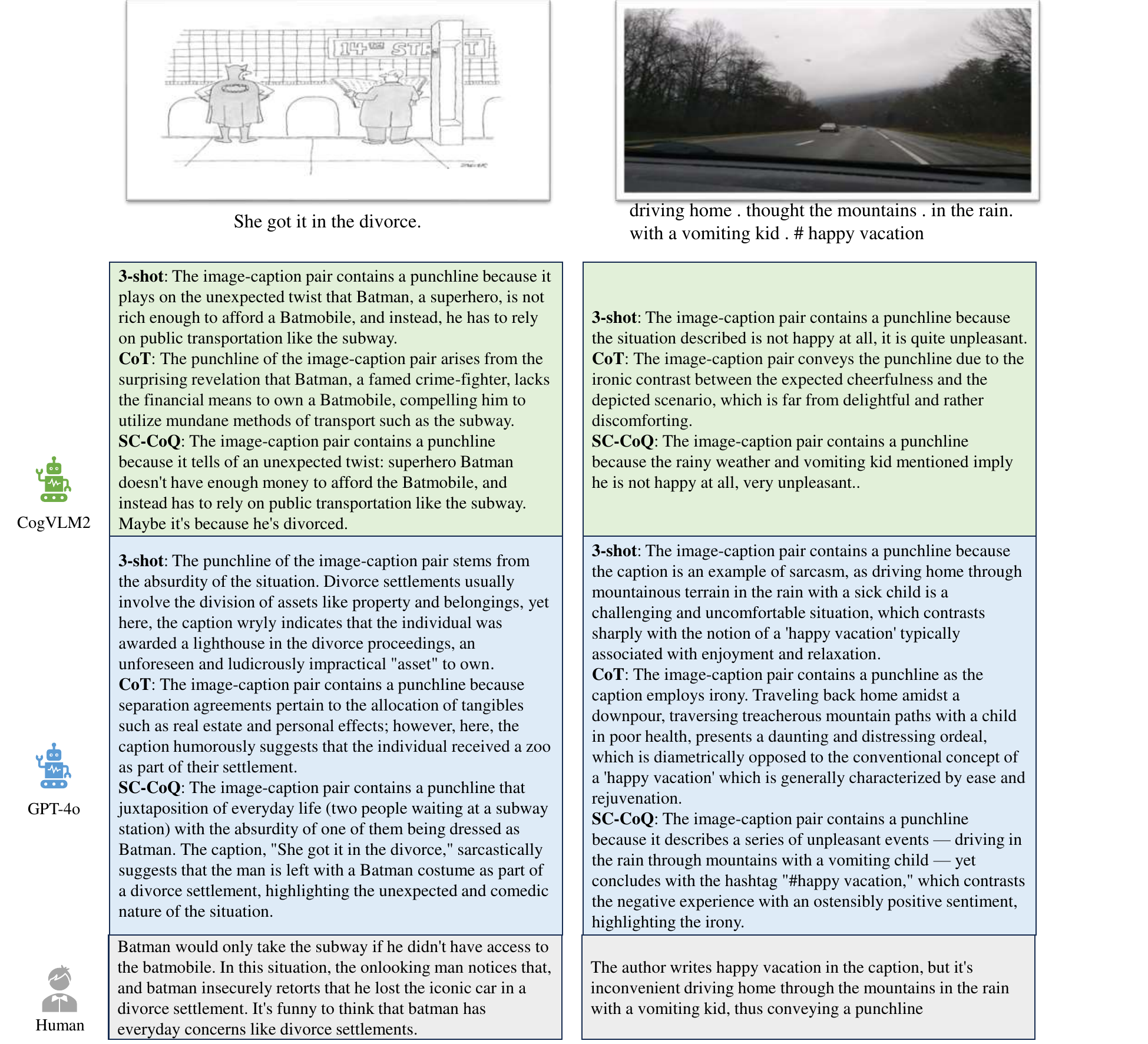}
    \caption{Two random samples of explanations generated by CogVLM2, GPT-4o, and human-written reasoning sentences. Notably, we present the generated reasoning sentences by CogVLM2 and GPT-4o prompted by 3-shot, CoT and SC-CoQ.}
    \label{Fig:example}
\end{figure*}

We provide result examples for \textit{Matching QA} of punchline perception in Figure~\ref{Fig:Q_analysis_matching1}. As we can see, when using SC-CoQ, the model correctly answers the question, while failing when utilizing other prompting methods.
For punchline reasoning task, we supply result examples for \textit{Yes/No QA} and \textit{Matching QA} in Figure~\ref{Fig:Q_analysis_reasoning}. In addition, we present result examples for \textit{Generation QA} in Figure~\ref{Fig:example}.
\label{subsec:qualitative analysis}
\input{Table/benchmark}
\section{Documentation, Licensing, Potential risk and Intended Use of PunchBench}
 PunchBench encompasses $6,000$ image-caption pairs and $54,000$ question-answer pairs for multimodal punchline comprehension. We evaluate punchline comprehension in two levels: shallow-level punchline perception and deep-level punchline reasoning. We introduce three question formats for each task.
We release the dataset without ground truth answers, along with a validation set that includes ground truth annotations, under the CC BY-NC 4.0 license\footnote{\url{https://creativecommons.org/licenses/by-nc/4.0/}}.
 Notably, there may be some offensive information in the images, despite we have made efforts to exclude the potential offensive information in the collection and filtering process. Furthermore, PunchBench should only be used for research purpose only.
 \section{Annotators Recruitment and Multimedia Platforms}
 \label{Appendix:Platform}
 For human baseline, we employed three undergraduates outside of the work as the annotators. For human evaluation, we asked another three undergraduate students to evaluate the quality of generated reasoning sentences. 
 The information about the multimedia platforms we used is listed as follows. The social media platforms X\footnote{\url{https://x.com/}.}, Instagram\footnote{\url{https://www.instagram.com/}.}, and YouTube\footnote{\url{https://www.youtube.com/}.}. Additionally, we include image-caption pairs from the cartoon websites like CartoonMovement\footnote{\url{https://www.cartoonmovement.com/}.} and CartoonStock\footnote{\url{https://www.cartoonstock.com/}.}.

%% file: Table/benchmark.tex
\begin{table*}[]
\centering
\caption{Comparison between our PunchBench and previous benchmarks.
}
\resizebox{\textwidth}{!}{
\begin{tabular}{cccccccc}
\hline
\textbf{Benchmarks} & \textbf{Domain} & \textbf{Task}    & \makecell{\textbf{Question}\\ \textbf{Format}}   &\makecell{\textbf{Punchline}\\ \textbf{Type}}  
& \makecell{\textbf{\#Num of }\\ \textbf{Image-caption Pairs}} & \makecell{\textbf{\#Num of} \\ \textbf{Question-answer Pairs}} 
\\ \hline
\textbf{MTSD}~\cite{DBLP:conf/acl/CaiCW19}       & Post       & Sarcasm Classification      &Single  &Sarcasm   &19,816 &19,816\\
\textbf{MORE}~\cite{DBLP:conf/aaai/Desai0A22}       & Post       & Sarcasm Explanation    &Single      &Sarcasm  &3,510 &3,510 \\ 
\textbf{HUB}~\cite{DBLP:conf/acl/HesselMHLDZM023}        & Cartoon  & \makecell{Matching,\\Ranking and Explanation}  &Single &Humor &704 &5,973        \\\hline
\textbf{PunchBech}  & \makecell{Cartoon, Post,\\Comment, Meme.}     & \makecell{Punchline Perception,\\Punchline Reasoning} &\makecell{\textit{Yes/No QA},\\ \textit{Matching QA},\\ \textit{Multi-option QA},\\ \textit{Generation QA}.} &\makecell{Humor, \\Sarcasm}     &6,000 &54,000
 \\ \hline
\end{tabular}
}
\label{Tab:Benchmark}
\end{table*}

%% file: main.bbl
\begin{thebibliography}{64}
\expandafter\ifx\csname natexlab\endcsname\relax\def\natexlab#1{#1}\fi

\bibitem[{AI@Meta(2024)}]{llama3modelcard}
AI@Meta. 2024.
\newblock \href {https://github.com/meta-llama/llama3/blob/main/MODEL_CARD.md} {Llama 3 model card}.

\bibitem[{Antol et~al.(2015)Antol, Agrawal, Lu, Mitchell, Batra, Zitnick, and Parikh}]{vqaiccv}
Stanislaw Antol, Aishwarya Agrawal, Jiasen Lu, Margaret Mitchell, Dhruv Batra, C.~Lawrence Zitnick, and Devi Parikh. 2015.
\newblock {VQA:} visual question answering.
\newblock In \emph{{ICCV}}, pages 2425--2433. {IEEE} Computer Society.

\bibitem[{Bai et~al.(2023{\natexlab{a}})Bai, Bai, Yang, Wang, Tan, Wang, Lin, Zhou, and Zhou}]{DBLP:journals/corr/abs-2308-12966}
Jinze Bai, Shuai Bai, Shusheng Yang, Shijie Wang, Sinan Tan, Peng Wang, Junyang Lin, Chang Zhou, and Jingren Zhou. 2023{\natexlab{a}}.
\newblock \href {https://doi.org/10.48550/ARXIV.2308.12966} {Qwen-vl: {A} frontier large vision-language model with versatile abilities}.
\newblock \emph{CoRR}, abs/2308.12966.

\bibitem[{Bai et~al.(2023{\natexlab{b}})Bai, Bai, Yang, Wang, Tan, Wang, Lin, Zhou, and Zhou}]{Qwen-VL}
Jinze Bai, Shuai Bai, Shusheng Yang, Shijie Wang, Sinan Tan, Peng Wang, Junyang Lin, Chang Zhou, and Jingren Zhou. 2023{\natexlab{b}}.
\newblock Qwen-vl: A versatile vision-language model for understanding, localization, text reading, and beyond.
\newblock \emph{arXiv preprint arXiv:2308.12966}.

\bibitem[{Brown et~al.(2020)Brown, Mann, Ryder, Subbiah, Kaplan, Dhariwal, Neelakantan, Shyam, Sastry, Askell, Agarwal, Herbert{-}Voss, Krueger, Henighan, Child, Ramesh, Ziegler, Wu, Winter, Hesse, Chen, Sigler, Litwin, Gray, Chess, Clark, Berner, McCandlish, Radford, Sutskever, and Amodei}]{DBLP:conf/nips/BrownMRSKDNSSAA20}
Tom~B. Brown, Benjamin Mann, Nick Ryder, Melanie Subbiah, Jared Kaplan, Prafulla Dhariwal, Arvind Neelakantan, Pranav Shyam, Girish Sastry, Amanda Askell, Sandhini Agarwal, Ariel Herbert{-}Voss, Gretchen Krueger, Tom Henighan, Rewon Child, Aditya Ramesh, Daniel~M. Ziegler, Jeffrey Wu, Clemens Winter, Christopher Hesse, Mark Chen, Eric Sigler, Mateusz Litwin, Scott Gray, Benjamin Chess, Jack Clark, Christopher Berner, Sam McCandlish, Alec Radford, Ilya Sutskever, and Dario Amodei. 2020.
\newblock \href {https://proceedings.neurips.cc/paper/2020/hash/1457c0d6bfcb4967418bfb8ac142f64a-Abstract.html} {Language models are few-shot learners}.
\newblock In \emph{Advances in Neural Information Processing Systems 33: Annual Conference on Neural Information Processing Systems 2020, NeurIPS 2020, December 6-12, 2020, virtual}.

\bibitem[{Cai et~al.(2019)Cai, Cai, and Wan}]{DBLP:conf/acl/CaiCW19}
Yitao Cai, Huiyu Cai, and Xiaojun Wan. 2019.
\newblock \href {https://doi.org/10.18653/V1/P19-1239} {Multi-modal sarcasm detection in twitter with hierarchical fusion model}.
\newblock In \emph{Proceedings of the 57th Conference of the Association for Computational Linguistics, {ACL} 2019, Florence, Italy, July 28- August 2, 2019, Volume 1: Long Papers}, pages 2506--2515. Association for Computational Linguistics.

\bibitem[{Cai et~al.(2024)Cai, Cao, Chen, Chen, Chen, Chen, Chen, Chen, Chen, Chu et~al.}]{DBLP:journals/corr/abs-2403-17297}
Zheng Cai, Maosong Cao, Haojiong Chen, Kai Chen, Keyu Chen, Xin Chen, Xun Chen, Zehui Chen, Zhi Chen, Pei Chu, et~al. 2024.
\newblock Internlm2 technical report.
\newblock \emph{arXiv preprint arXiv:2403.17297}.

\bibitem[{Castro et~al.(2019)Castro, Hazarika, P{\'e}rez-Rosas, Zimmermann, Mihalcea, and Poria}]{castro-etal-2019-towards}
Santiago Castro, Devamanyu Hazarika, Ver{\'o}nica P{\'e}rez-Rosas, Roger Zimmermann, Rada Mihalcea, and Soujanya Poria. 2019.
\newblock \href {https://doi.org/10.18653/v1/P19-1455} {Towards multimodal sarcasm detection (an {\_}{O}bviously{\_} perfect paper)}.
\newblock In \emph{Proceedings of the 57th Annual Meeting of the Association for Computational Linguistics}, pages 4619--4629, Florence, Italy. Association for Computational Linguistics.

\bibitem[{Chen et~al.(2024{\natexlab{a}})Chen, Wang, Cao, Liu, Gao, Cui, Zhu, Ye, Tian, Liu et~al.}]{chen2024expanding}
Zhe Chen, Weiyun Wang, Yue Cao, Yangzhou Liu, Zhangwei Gao, Erfei Cui, Jinguo Zhu, Shenglong Ye, Hao Tian, Zhaoyang Liu, et~al. 2024{\natexlab{a}}.
\newblock Expanding performance boundaries of open-source multimodal models with model, data, and test-time scaling.
\newblock \emph{arXiv preprint arXiv:2412.05271}.

\bibitem[{Chen et~al.(2024{\natexlab{b}})Chen, Wu, Wang, Su, Chen, Xing, Zhong, Zhang, Zhu, Lu et~al.}]{chen2024internvl}
Zhe Chen, Jiannan Wu, Wenhai Wang, Weijie Su, Guo Chen, Sen Xing, Muyan Zhong, Qinglong Zhang, Xizhou Zhu, Lewei Lu, et~al. 2024{\natexlab{b}}.
\newblock Internvl: Scaling up vision foundation models and aligning for generic visual-linguistic tasks.
\newblock In \emph{Proceedings of the IEEE/CVF conference on computer vision and pattern recognition}, pages 24185--24198.

\bibitem[{Cheng et~al.(2024)Cheng, Leng, Zhang, Xin, Li, Chen, Zhu, Zhang, Luo, and Bing}]{damonlpsg2024videollama2}
Zesen Cheng, Sicong Leng, Hang Zhang, Yifei Xin, Xin Li, Guanzheng Chen, Yongxin Zhu, Wenqi Zhang, Ziyang Luo, and Lidong Bing. 2024.
\newblock \href {https://arxiv.org/abs/2406.07476} {Videollama 2: Advancing spatial-temporal modeling and audio understanding in video-llms}.
\newblock \emph{arXiv preprint arXiv:2406.07476}.

\bibitem[{Desai et~al.(2022)Desai, Chakraborty, and Akhtar}]{DBLP:conf/aaai/Desai0A22}
Poorav Desai, Tanmoy Chakraborty, and Md.~Shad Akhtar. 2022.
\newblock Nice perfume. how long did you marinate in it? multimodal sarcasm explanation.
\newblock In \emph{Thirty-Sixth {AAAI} Conference on Artificial Intelligence, {AAAI}}, pages 10563--10571. AAAI.

\bibitem[{Dosovitskiy et~al.(2021)Dosovitskiy, Beyer, Kolesnikov, Weissenborn, Zhai, Unterthiner, Dehghani, Minderer, Heigold, Gelly, Uszkoreit, and Houlsby}]{DBLP:conf/iclr/DosovitskiyB0WZ21}
Alexey Dosovitskiy, Lucas Beyer, Alexander Kolesnikov, Dirk Weissenborn, Xiaohua Zhai, Thomas Unterthiner, Mostafa Dehghani, Matthias Minderer, Georg Heigold, Sylvain Gelly, Jakob Uszkoreit, and Neil Houlsby. 2021.
\newblock \href {https://openreview.net/forum?id=YicbFdNTTy} {An image is worth 16x16 words: Transformers for image recognition at scale}.
\newblock In \emph{9th International Conference on Learning Representations, {ICLR} 2021, Virtual Event, Austria, May 3-7, 2021}. OpenReview.net.

\bibitem[{GLM et~al.(2024)GLM, Zeng, Xu, Wang, Zhang, Yin, Rojas, Feng, Zhao, Lai, Yu, Wang, Sun, Zhang, Cheng, Gui, Tang, Zhang, Li, Zhao, Wu, Zhong, Liu, Huang, Zhang, Zheng, Lu, Duan, Zhang, Cao, Yang, Tam, Zhao, Liu, Xia, Zhang, Gu, Lv, Liu, Liu, Yang, Song, Zhang, An, Xu, Niu, Yang, Li, Bai, Dong, Qi, Wang, Yang, Du, Hou, and Wang}]{glm2024chatglm}
Team GLM, Aohan Zeng, Bin Xu, Bowen Wang, Chenhui Zhang, Da~Yin, Diego Rojas, Guanyu Feng, Hanlin Zhao, Hanyu Lai, Hao Yu, Hongning Wang, Jiadai Sun, Jiajie Zhang, Jiale Cheng, Jiayi Gui, Jie Tang, Jing Zhang, Juanzi Li, Lei Zhao, Lindong Wu, Lucen Zhong, Mingdao Liu, Minlie Huang, Peng Zhang, Qinkai Zheng, Rui Lu, Shuaiqi Duan, Shudan Zhang, Shulin Cao, Shuxun Yang, Weng~Lam Tam, Wenyi Zhao, Xiao Liu, Xiao Xia, Xiaohan Zhang, Xiaotao Gu, Xin Lv, Xinghan Liu, Xinyi Liu, Xinyue Yang, Xixuan Song, Xunkai Zhang, Yifan An, Yifan Xu, Yilin Niu, Yuantao Yang, Yueyan Li, Yushi Bai, Yuxiao Dong, Zehan Qi, Zhaoyu Wang, Zhen Yang, Zhengxiao Du, Zhenyu Hou, and Zihan Wang. 2024.
\newblock \href {http://arxiv.org/abs/2406.12793} {Chatglm: A family of large language models from glm-130b to glm-4 all tools}.

\bibitem[{Gwet(2014)}]{Gwet2014HandbookOI}
Kilem~L. Gwet. 2014.
\newblock Handbook of inter-rater reliability: The definitive guide to measuring the extent of agreement among raters.
\newblock In \emph{4th edition edition}, pages 1--38. Advanced Analytics, LLC.

\bibitem[{Hempelmann and Petrenko(2015)}]{DBLP:conf/hci/HempelmannP15}
Christian~F. Hempelmann and Max Petrenko. 2015.
\newblock \href {https://doi.org/10.1007/978-3-319-20804-6\_59} {An {AI} for humorously reframing interaction narratives with human users}.
\newblock In \emph{Distributed, Ambient, and Pervasive Interactions - Third International Conference, {DAPI} 2015, Held as Part of {HCI} International 2015, Los Angeles, CA, USA, August 2-7, 2015, Proceedings}, volume 9189 of \emph{Lecture Notes in Computer Science}, pages 651--658. Springer.

\bibitem[{Hessel et~al.(2023)Hessel, Marasovic, Hwang, Lee, Da, Zellers, Mankoff, and Choi}]{DBLP:conf/acl/HesselMHLDZM023}
Jack Hessel, Ana Marasovic, Jena~D. Hwang, Lillian Lee, Jeff Da, Rowan Zellers, Robert Mankoff, and Yejin Choi. 2023.
\newblock \href {https://doi.org/10.18653/V1/2023.ACL-LONG.41} {Do androids laugh at electric sheep? humor "understanding" benchmarks from the new yorker caption contest}.
\newblock In \emph{Proceedings of the 61st Annual Meeting of the Association for Computational Linguistics (Volume 1: Long Papers), {ACL} 2023, Toronto, Canada, July 9-14, 2023}, pages 688--714. Association for Computational Linguistics.

\bibitem[{Hong et~al.(2024)Hong, Wang, Ding, Yu, Lv, Wang, Cheng, Huang, Ji, Xue et~al.}]{hong2024cogvlm2}
Wenyi Hong, Weihan Wang, Ming Ding, Wenmeng Yu, Qingsong Lv, Yan Wang, Yean Cheng, Shiyu Huang, Junhui Ji, Zhao Xue, et~al. 2024.
\newblock Cogvlm2: Visual language models for image and video understanding.
\newblock \emph{arXiv preprint arXiv:2408.16500}.

\bibitem[{Islam et~al.(2017)Islam, Islam, and Noor}]{Islam2017ASO}
Noman Islam, Zeeshan Islam, and Nazia Noor. 2017.
\newblock \href {https://api.semanticscholar.org/CorpusID:1040908} {A survey on optical character recognition system}.
\newblock \emph{ArXiv}, abs/1710.05703.

\bibitem[{Jentzsch and Kersting(2023)}]{DBLP:conf/wassa/JentzschK23}
Sophie~F. Jentzsch and Kristian Kersting. 2023.
\newblock \href {https://doi.org/10.18653/V1/2023.WASSA-1.29} {Chatgpt is fun, but it is not funny! humor is still challenging large language models}.
\newblock In \emph{Proceedings of the 13th Workshop on Computational Approaches to Subjectivity, Sentiment, {\&} Social Media Analysis, WASSA@ACL 2023, Toronto, Canada, July 14, 2023}, pages 325--340. Association for Computational Linguistics.

\bibitem[{Jian et~al.(2024)Jian, Yu, and Zhang}]{DBLP:conf/emnlp/JianY024}
Pu~Jian, Donglei Yu, and Jiajun Zhang. 2024.
\newblock \href {https://aclanthology.org/2024.emnlp-main.613} {Large language models know what is key visual entity: An llm-assisted multimodal retrieval for {VQA}}.
\newblock In \emph{Proceedings of the 2024 Conference on Empirical Methods in Natural Language Processing, {EMNLP} 2024, Miami, FL, USA, November 12-16, 2024}, pages 10939--10956. Association for Computational Linguistics.

\bibitem[{Jing et~al.(2023)Jing, Song, Ouyang, Jia, and Nie}]{DBLP:conf/acl/JingSOJN23}
Liqiang Jing, Xuemeng Song, Kun Ouyang, Mengzhao Jia, and Liqiang Nie. 2023.
\newblock \href {https://doi.org/10.18653/V1/2023.ACL-LONG.635} {Multi-source semantic graph-based multimodal sarcasm explanation generation}.
\newblock In \emph{Proceedings of the 61st Annual Meeting of the Association for Computational Linguistics (Volume 1: Long Papers), {ACL} 2023, Toronto, Canada, July 9-14, 2023}, pages 11349--11361. Association for Computational Linguistics.

\bibitem[{Johnson et~al.(2016)Johnson, Karpathy, and Fei{-}Fei}]{DBLP:conf/cvpr/JohnsonKF16}
Justin Johnson, Andrej Karpathy, and Li~Fei{-}Fei. 2016.
\newblock \href {https://doi.org/10.1109/CVPR.2016.494} {Densecap: Fully convolutional localization networks for dense captioning}.
\newblock In \emph{2016 {IEEE} Conference on Computer Vision and Pattern Recognition, {CVPR} 2016, Las Vegas, NV, USA, June 27-30, 2016}, pages 4565--4574. {IEEE} Computer Society.

\bibitem[{Ko et~al.(2023)Ko, Lee, and Kim}]{DBLP:conf/emnlp/KoLK23}
Dayoon Ko, Sangho Lee, and Gunhee Kim. 2023.
\newblock \href {https://doi.org/10.18653/V1/2023.EMNLP-MAIN.176} {Can language models laugh at youtube short-form videos?}
\newblock In \emph{Proceedings of the 2023 Conference on Empirical Methods in Natural Language Processing, {EMNLP} 2023, Singapore, December 6-10, 2023}, pages 2897--2916. Association for Computational Linguistics.

\bibitem[{Kumar et~al.(2022)Kumar, Kulkarni, Akhtar, and Chakraborty}]{DBLP:conf/acl/KumarKA022}
Shivani Kumar, Atharva Kulkarni, Md.~Shad Akhtar, and Tanmoy Chakraborty. 2022.
\newblock \href {https://doi.org/10.18653/V1/2022.ACL-LONG.411} {When did you become so smart, oh wise one?! sarcasm explanation in multi-modal multi-party dialogues}.
\newblock In \emph{Proceedings of the 60th Annual Meeting of the Association for Computational Linguistics (Volume 1: Long Papers), {ACL} 2022, Dublin, Ireland, May 22-27, 2022}, pages 5956--5968. Association for Computational Linguistics.

\bibitem[{Li et~al.(2024{\natexlab{a}})Li, Zhang, Guo, Zhang, Li, Zhang, Zhang, Zhang, Li, Liu et~al.}]{li2024llava}
Bo~Li, Yuanhan Zhang, Dong Guo, Renrui Zhang, Feng Li, Hao Zhang, Kaichen Zhang, Peiyuan Zhang, Yanwei Li, Ziwei Liu, et~al. 2024{\natexlab{a}}.
\newblock Llava-onevision: Easy visual task transfer.
\newblock \emph{arXiv preprint arXiv:2408.03326}.

\bibitem[{Li et~al.(2024{\natexlab{b}})Li, Liu, Wu, Wang, Shen, Qu, Niu, Zhou, Huang, Li et~al.}]{li2024aria}
Dongxu Li, Yudong Liu, Haoning Wu, Yue Wang, Zhiqi Shen, Bowen Qu, Xinyao Niu, Fan Zhou, Chengen Huang, Yanpeng Li, et~al. 2024{\natexlab{b}}.
\newblock Aria: An open multimodal native mixture-of-experts model.
\newblock \emph{arXiv preprint arXiv:2410.05993}.

\bibitem[{Li et~al.(2023{\natexlab{a}})Li, Li, Savarese, and Hoi}]{DBLP:conf/icml/0008LSH23}
Junnan Li, Dongxu Li, Silvio Savarese, and Steven C.~H. Hoi. 2023{\natexlab{a}}.
\newblock \href {https://proceedings.mlr.press/v202/li23q.html} {{BLIP-2:} bootstrapping language-image pre-training with frozen image encoders and large language models}.
\newblock In \emph{International Conference on Machine Learning, {ICML} 2023, 23-29 July 2023, Honolulu, Hawaii, {USA}}, volume 202 of \emph{Proceedings of Machine Learning Research}, pages 19730--19742. {PMLR}.

\bibitem[{Li et~al.(2023{\natexlab{b}})Li, He, Wang, Li, Wang, Luo, Wang, Wang, and Qiao}]{2023videochat}
Kunchang Li, Yinan He, Yi~Wang, Yizhuo Li, Wenhai Wang, Ping Luo, Yali Wang, Limin Wang, and Yu~Qiao. 2023{\natexlab{b}}.
\newblock Videochat: Chat-centric video understanding.
\newblock \emph{arXiv preprint arXiv:2305.06355}.

\bibitem[{Lin et~al.(2023)Lin, Zhu, Ye, Ning, Jin, and Yuan}]{lin2023video}
Bin Lin, Bin Zhu, Yang Ye, Munan Ning, Peng Jin, and Li~Yuan. 2023.
\newblock Video-llava: Learning united visual representation by alignment before projection.
\newblock \emph{arXiv preprint arXiv:2311.10122}.

\bibitem[{Lin(2004)}]{lin-2004-rouge}
Chin-Yew Lin. 2004.
\newblock \href {https://aclanthology.org/W04-1013} {{ROUGE}: A package for automatic evaluation of summaries}.
\newblock In \emph{Text Summarization Branches Out}, pages 74--81. Association for Computational Linguistics.

\bibitem[{Liu et~al.(2024{\natexlab{a}})Liu, Li, Li, and Lee}]{DBLP:conf/cvpr/LiuLLL24}
Haotian Liu, Chunyuan Li, Yuheng Li, and Yong~Jae Lee. 2024{\natexlab{a}}.
\newblock \href {https://doi.org/10.1109/CVPR52733.2024.02484} {Improved baselines with visual instruction tuning}.
\newblock In \emph{{IEEE/CVF} Conference on Computer Vision and Pattern Recognition, {CVPR} 2024, Seattle, WA, USA, June 16-22, 2024}, pages 26286--26296. {IEEE}.

\bibitem[{Liu et~al.(2024{\natexlab{b}})Liu, Li, Li, Li, Zhang, Shen, and Lee}]{liu2024llavanext}
Haotian Liu, Chunyuan Li, Yuheng Li, Bo~Li, Yuanhan Zhang, Sheng Shen, and Yong~Jae Lee. 2024{\natexlab{b}}.
\newblock \href {https://llava-vl.github.io/blog/2024-01-30-llava-next/} {Llava-next: Improved reasoning, ocr, and world knowledge}.

\bibitem[{Liu et~al.(2023{\natexlab{a}})Liu, Li, Wu, and Lee}]{DBLP:conf/nips/LiuLWL23a}
Haotian Liu, Chunyuan Li, Qingyang Wu, and Yong~Jae Lee. 2023{\natexlab{a}}.
\newblock \href {http://papers.nips.cc/paper\_files/paper/2023/hash/6dcf277ea32ce3288914faf369fe6de0-Abstract-Conference.html} {Visual instruction tuning}.
\newblock In \emph{Advances in Neural Information Processing Systems 36: Annual Conference on Neural Information Processing Systems 2023, NeurIPS 2023, New Orleans, LA, USA, December 10 - 16, 2023}.

\bibitem[{Liu et~al.(2023{\natexlab{b}})Liu, Li, Wu, and Lee}]{liu2023llava}
Haotian Liu, Chunyuan Li, Qingyang Wu, and Yong~Jae Lee. 2023{\natexlab{b}}.
\newblock Visual instruction tuning.

\bibitem[{Liu et~al.(2024{\natexlab{c}})Liu, Li, Liu, Wang, Ren, Li, Chen, Sun, and Hou}]{DBLP:conf/acl/LiuLLWRLCSH24}
Yuanxin Liu, Shicheng Li, Yi~Liu, Yuxiang Wang, Shuhuai Ren, Lei Li, Sishuo Chen, Xu~Sun, and Lu~Hou. 2024{\natexlab{c}}.
\newblock \href {https://doi.org/10.18653/V1/2024.FINDINGS-ACL.517} {Tempcompass: Do video llms really understand videos?}
\newblock In \emph{Findings of the Association for Computational Linguistics, {ACL} 2024, Bangkok, Thailand and virtual meeting, August 11-16, 2024}, pages 8731--8772. Association for Computational Linguistics.

\bibitem[{Liu et~al.(2023{\natexlab{c}})Liu, Li, Li, Yu, Huang, Peng, Liu, Chen, Li, Jin, and Bai}]{ocrbench}
Yuliang Liu, Zhang Li, Hongliang Li, Wenwen Yu, Mingxin Huang, Dezhi Peng, Mingyu Liu, Mingrui Chen, Chunyuan Li, Lianwen Jin, and Xiang Bai. 2023{\natexlab{c}}.
\newblock On the hidden mystery of {OCR} in large multimodal models.
\newblock \emph{CoRR}, abs/2305.07895.

\bibitem[{Long et~al.(2023)Long, Wen, Han, Xu, Ren, Zhang, Zhao, and Liang}]{DBLP:conf/cvpr/LongWHXRZZL23}
Yanxin Long, Youpeng Wen, Jianhua Han, Hang Xu, Pengzhen Ren, Wei Zhang, Shen Zhao, and Xiaodan Liang. 2023.
\newblock \href {https://doi.org/10.1109/CVPR52729.2023.01462} {Capdet: Unifying dense captioning and open-world detection pretraining}.
\newblock In \emph{{IEEE/CVF} Conference on Computer Vision and Pattern Recognition, {CVPR} 2023, Vancouver, BC, Canada, June 17-24, 2023}, pages 15233--15243. {IEEE}.

\bibitem[{Lu et~al.(2024)Lu, Bansal, Xia, Liu, Li, Hajishirzi, Cheng, Chang, Galley, and Gao}]{DBLP:conf/iclr/LuBX0LH0CG024}
Pan Lu, Hritik Bansal, Tony Xia, Jiacheng Liu, Chunyuan Li, Hannaneh Hajishirzi, Hao Cheng, Kai{-}Wei Chang, Michel Galley, and Jianfeng Gao. 2024.
\newblock \href {https://openreview.net/forum?id=KUNzEQMWU7} {Mathvista: Evaluating mathematical reasoning of foundation models in visual contexts}.
\newblock In \emph{The Twelfth International Conference on Learning Representations, {ICLR} 2024, Vienna, Austria, May 7-11, 2024}. OpenReview.net.

\bibitem[{Luo et~al.(2023)Luo, Zhao, Yang, Dong, Qiu, Lu, Wang, and Wei}]{luo2023valley}
Ruipu Luo, Ziwang Zhao, Min Yang, Junwei Dong, Minghui Qiu, Pengcheng Lu, Tao Wang, and Zhongyu Wei. 2023.
\newblock \href {http://arxiv.org/abs/2306.07207} {Valley: Video assistant with large language model enhanced ability}.

\bibitem[{Maaz et~al.(2024)Maaz, Rasheed, Khan, and Khan}]{Maaz2023VideoChatGPT}
Muhammad Maaz, Hanoona Rasheed, Salman Khan, and Fahad~Shahbaz Khan. 2024.
\newblock Video-chatgpt: Towards detailed video understanding via large vision and language models.
\newblock In \emph{Proceedings of the 62nd Annual Meeting of the Association for Computational Linguistics (ACL 2024)}.

\bibitem[{Mahdaouy et~al.(2021)Mahdaouy, Mekki, Essefar, Mamoun, Berrada, and Khoumsi}]{DBLP:conf/wanlp/MahdaouyMEMBK21}
Abdelkader~El Mahdaouy, Abdellah~El Mekki, Kabil Essefar, Nabil~El Mamoun, Ismail Berrada, and Ahmed Khoumsi. 2021.
\newblock \href {https://www.aclweb.org/anthology/2021.wanlp-1.42/} {Deep multi-task model for sarcasm detection and sentiment analysis in arabic language}.
\newblock In \emph{Proceedings of the Sixth Arabic Natural Language Processing Workshop, {WANLP} 2021, Kyiv, Ukraine (Virtual), April 9, 2021}, pages 334--339. Association for Computational Linguistics.

\bibitem[{Mathew et~al.(2021)Mathew, Karatzas, and Jawahar}]{DBLP:conf/wacv/MathewKJ21}
Minesh Mathew, Dimosthenis Karatzas, and C.~V. Jawahar. 2021.
\newblock \href {https://doi.org/10.1109/WACV48630.2021.00225} {Docvqa: {A} dataset for {VQA} on document images}.
\newblock In \emph{{IEEE} Winter Conference on Applications of Computer Vision, {WACV} 2021, Waikoloa, HI, USA, January 3-8, 2021}, pages 2199--2208. {IEEE}.

\bibitem[{OpenAI(2022)}]{chatgpt}
OpenAI. 2022.
\newblock \href {https://openai.com/blog/chatgpt} {Introducing chatgpt}.
\newblock \emph{CoRR}.

\bibitem[{OpenAI(2023a)}]{GPT-4V}
OpenAI. 2023a.
\newblock \href {https://cdn.openai.com/papers/GPTV_System_Card.pdf} {Gpt-4v(ision) system card}.

\bibitem[{OpenAI(2024)}]{GPT-4o}
OpenAI. 2024.
\newblock \href {https://openai.com/index/hello-gpt-4o/} {Gpt-4o}.

\bibitem[{OpenAI et~al.(2024)OpenAI, Achiam, Adler, Agarwal, Ahmad, Akkaya, Aleman, Almeida, Altenschmidt, Altman, Anadkat, Avila, Babuschkin, Balaji, Balcom, Baltescu, Bao, Bavarian, Belgum, Bello, Berdine, Bernadett-Shapiro, Berner, Bogdonoff, Boiko, Boyd, Brakman, Brockman, Brooks, Brundage, Button, Cai, Campbell, Cann, Carey, Carlson, Carmichael, Chan, Chang, Chantzis, Chen, Chen, Chen, Chen, Chen, Chess, Cho, Chu, Chung, Cummings, Currier, Dai, Decareaux, Degry, Deutsch, Deville, Dhar, Dohan, Dowling, Dunning, Ecoffet, Eleti, Eloundou, Farhi, Fedus, Felix, Fishman, Forte, Fulford, Gao, Georges, Gibson, Goel, Gogineni, Goh, Gontijo-Lopes, Gordon, Grafstein, Gray, Greene, Gross, Gu, Guo, Hallacy, Han, Harris, He, Heaton, Heidecke, Hesse, Hickey, Hickey, Hoeschele, Houghton, Hsu, Hu, Hu, Huizinga, Jain, Jain, Jang, Jiang, Jiang, Jin, Jin, Jomoto, Jonn, Jun, Kaftan, Łukasz Kaiser, Kamali, Kanitscheider, Keskar, Khan, Kilpatrick, Kim, Kim, Kim, Kirchner, Kiros, Knight, Kokotajlo, Łukasz Kondraciuk,
  Kondrich, Konstantinidis, Kosic, Krueger, Kuo, Lampe, Lan, Lee, Leike, Leung, Levy, Li, Lim, Lin, Lin, Litwin, Lopez, Lowe, Lue, Makanju, Malfacini, Manning, Markov, Markovski, Martin, Mayer, Mayne, McGrew, McKinney, McLeavey, McMillan, McNeil, Medina, Mehta, Menick, Metz, Mishchenko, Mishkin, Monaco, Morikawa, Mossing, Mu, Murati, Murk, Mély, Nair, Nakano, Nayak, Neelakantan, Ngo, Noh, Ouyang, O'Keefe, Pachocki, Paino, Palermo, Pantuliano, Parascandolo, Parish, Parparita, Passos, Pavlov, Peng, Perelman, de~Avila Belbute~Peres, Petrov, de~Oliveira~Pinto, Michael, Pokorny, Pokrass, Pong, Powell, Power, Power, Proehl, Puri, Radford, Rae, Ramesh, Raymond, Real, Rimbach, Ross, Rotsted, Roussez, Ryder, Saltarelli, Sanders, Santurkar, Sastry, Schmidt, Schnurr, Schulman, Selsam, Sheppard, Sherbakov, Shieh, Shoker, Shyam, Sidor, Sigler, Simens, Sitkin, Slama, Sohl, Sokolowsky, Song, Staudacher, Such, Summers, Sutskever, Tang, Tezak, Thompson, Tillet, Tootoonchian, Tseng, Tuggle, Turley, Tworek, Uribe, Vallone,
  Vijayvergiya, Voss, Wainwright, Wang, Wang, Wang, Ward, Wei, Weinmann, Welihinda, Welinder, Weng, Weng, Wiethoff, Willner, Winter, Wolrich, Wong, Workman, Wu, Wu, Wu, Xiao, Xu, Yoo, Yu, Yuan, Zaremba, Zellers, Zhang, Zhang, Zhao, Zheng, Zhuang, Zhuk, and Zoph}]{openai2024gpt4}
OpenAI, Josh Achiam, Steven Adler, Sandhini Agarwal, Lama Ahmad, Ilge Akkaya, Florencia~Leoni Aleman, Diogo Almeida, Janko Altenschmidt, Sam Altman, Shyamal Anadkat, Red Avila, Igor Babuschkin, Suchir Balaji, Valerie Balcom, Paul Baltescu, Haiming Bao, Mohammad Bavarian, Jeff Belgum, Irwan Bello, Jake Berdine, Gabriel Bernadett-Shapiro, Christopher Berner, Lenny Bogdonoff, Oleg Boiko, Madelaine Boyd, Anna-Luisa Brakman, Greg Brockman, Tim Brooks, Miles Brundage, Kevin Button, Trevor Cai, Rosie Campbell, Andrew Cann, Brittany Carey, Chelsea Carlson, Rory Carmichael, Brooke Chan, Che Chang, Fotis Chantzis, Derek Chen, Sully Chen, Ruby Chen, Jason Chen, Mark Chen, Ben Chess, Chester Cho, Casey Chu, Hyung~Won Chung, Dave Cummings, Jeremiah Currier, Yunxing Dai, Cory Decareaux, Thomas Degry, Noah Deutsch, Damien Deville, Arka Dhar, David Dohan, Steve Dowling, Sheila Dunning, Adrien Ecoffet, Atty Eleti, Tyna Eloundou, David Farhi, Liam Fedus, Niko Felix, Simón~Posada Fishman, Juston Forte, Isabella Fulford, Leo
  Gao, Elie Georges, Christian Gibson, Vik Goel, Tarun Gogineni, Gabriel Goh, Rapha Gontijo-Lopes, Jonathan Gordon, Morgan Grafstein, Scott Gray, Ryan Greene, Joshua Gross, Shixiang~Shane Gu, Yufei Guo, Chris Hallacy, Jesse Han, Jeff Harris, Yuchen He, Mike Heaton, Johannes Heidecke, Chris Hesse, Alan Hickey, Wade Hickey, Peter Hoeschele, Brandon Houghton, Kenny Hsu, Shengli Hu, Xin Hu, Joost Huizinga, Shantanu Jain, Shawn Jain, Joanne Jang, Angela Jiang, Roger Jiang, Haozhun Jin, Denny Jin, Shino Jomoto, Billie Jonn, Heewoo Jun, Tomer Kaftan, Łukasz Kaiser, Ali Kamali, Ingmar Kanitscheider, Nitish~Shirish Keskar, Tabarak Khan, Logan Kilpatrick, Jong~Wook Kim, Christina Kim, Yongjik Kim, Jan~Hendrik Kirchner, Jamie Kiros, Matt Knight, Daniel Kokotajlo, Łukasz Kondraciuk, Andrew Kondrich, Aris Konstantinidis, Kyle Kosic, Gretchen Krueger, Vishal Kuo, Michael Lampe, Ikai Lan, Teddy Lee, Jan Leike, Jade Leung, Daniel Levy, Chak~Ming Li, Rachel Lim, Molly Lin, Stephanie Lin, Mateusz Litwin, Theresa Lopez, Ryan
  Lowe, Patricia Lue, Anna Makanju, Kim Malfacini, Sam Manning, Todor Markov, Yaniv Markovski, Bianca Martin, Katie Mayer, Andrew Mayne, Bob McGrew, Scott~Mayer McKinney, Christine McLeavey, Paul McMillan, Jake McNeil, David Medina, Aalok Mehta, Jacob Menick, Luke Metz, Andrey Mishchenko, Pamela Mishkin, Vinnie Monaco, Evan Morikawa, Daniel Mossing, Tong Mu, Mira Murati, Oleg Murk, David Mély, Ashvin Nair, Reiichiro Nakano, Rajeev Nayak, Arvind Neelakantan, Richard Ngo, Hyeonwoo Noh, Long Ouyang, Cullen O'Keefe, Jakub Pachocki, Alex Paino, Joe Palermo, Ashley Pantuliano, Giambattista Parascandolo, Joel Parish, Emy Parparita, Alex Passos, Mikhail Pavlov, Andrew Peng, Adam Perelman, Filipe de~Avila Belbute~Peres, Michael Petrov, Henrique~Ponde de~Oliveira~Pinto, Michael, Pokorny, Michelle Pokrass, Vitchyr~H. Pong, Tolly Powell, Alethea Power, Boris Power, Elizabeth Proehl, Raul Puri, Alec Radford, Jack Rae, Aditya Ramesh, Cameron Raymond, Francis Real, Kendra Rimbach, Carl Ross, Bob Rotsted, Henri Roussez,
  Nick Ryder, Mario Saltarelli, Ted Sanders, Shibani Santurkar, Girish Sastry, Heather Schmidt, David Schnurr, John Schulman, Daniel Selsam, Kyla Sheppard, Toki Sherbakov, Jessica Shieh, Sarah Shoker, Pranav Shyam, Szymon Sidor, Eric Sigler, Maddie Simens, Jordan Sitkin, Katarina Slama, Ian Sohl, Benjamin Sokolowsky, Yang Song, Natalie Staudacher, Felipe~Petroski Such, Natalie Summers, Ilya Sutskever, Jie Tang, Nikolas Tezak, Madeleine~B. Thompson, Phil Tillet, Amin Tootoonchian, Elizabeth Tseng, Preston Tuggle, Nick Turley, Jerry Tworek, Juan Felipe~Cerón Uribe, Andrea Vallone, Arun Vijayvergiya, Chelsea Voss, Carroll Wainwright, Justin~Jay Wang, Alvin Wang, Ben Wang, Jonathan Ward, Jason Wei, CJ~Weinmann, Akila Welihinda, Peter Welinder, Jiayi Weng, Lilian Weng, Matt Wiethoff, Dave Willner, Clemens Winter, Samuel Wolrich, Hannah Wong, Lauren Workman, Sherwin Wu, Jeff Wu, Michael Wu, Kai Xiao, Tao Xu, Sarah Yoo, Kevin Yu, Qiming Yuan, Wojciech Zaremba, Rowan Zellers, Chong Zhang, Marvin Zhang, Shengjia
  Zhao, Tianhao Zheng, Juntang Zhuang, William Zhuk, and Barret Zoph. 2024.
\newblock \href {http://arxiv.org/abs/2303.08774} {Gpt-4 technical report}.

\bibitem[{Ouyang et~al.(2024)Ouyang, Jing, Song, Liu, Hu, and Nie}]{DBLP:journals/corr/abs-2402-03658}
Kun Ouyang, Liqiang Jing, Xuemeng Song, Meng Liu, Yupeng Hu, and Liqiang Nie. 2024.
\newblock \href {https://doi.org/10.48550/ARXIV.2402.03658} {Sentiment-enhanced graph-based sarcasm explanation in dialogue}.
\newblock \emph{CoRR}, abs/2402.03658.

\bibitem[{Papineni et~al.(2002)Papineni, Roukos, Ward, and Zhu}]{DBLP:conf/acl/PapineniRWZ02}
Kishore Papineni, Salim Roukos, Todd Ward, and Wei{-}Jing Zhu. 2002.
\newblock \href {https://doi.org/10.3115/1073083.1073135} {Bleu: a method for automatic evaluation of machine translation}.
\newblock In \emph{Proceedings of the 40th Annual Meeting of the Association for Computational Linguistics, July 6-12, 2002, Philadelphia, PA, {USA}}, pages 311--318. {ACL}.

\bibitem[{Qiao et~al.(2023)Qiao, Jing, Song, Chen, Zhu, and Nie}]{DBLP:conf/aaai/QiaoJSCZN23}
Yang Qiao, Liqiang Jing, Xuemeng Song, Xiaolin Chen, Lei Zhu, and Liqiang Nie. 2023.
\newblock \href {https://doi.org/10.1609/AAAI.V37I8.26138} {Mutual-enhanced incongruity learning network for multi-modal sarcasm detection}.
\newblock In \emph{Thirty-Seventh {AAAI} Conference on Artificial Intelligence, {AAAI} 2023, Thirty-Fifth Conference on Innovative Applications of Artificial Intelligence, {IAAI} 2023, Thirteenth Symposium on Educational Advances in Artificial Intelligence, {EAAI} 2023, Washington, DC, USA, February 7-14, 2023}, pages 9507--9515. {AAAI} Press.

\bibitem[{Shazeer(2020)}]{DBLP:journals/corr/abs-2002-05202}
Noam Shazeer. 2020.
\newblock \href {http://arxiv.org/abs/2002.05202} {{GLU} variants improve transformer}.
\newblock \emph{CoRR}, abs/2002.05202.

\bibitem[{Team(2023)}]{Mistral}
The Mistral~AI Team. 2023.
\newblock \href {https://huggingface.co/mistralai/Mistral-7B-Instruct-v0.2} {Mistral-7b-instruct-v0.2}.

\bibitem[{Touvron et~al.(2023{\natexlab{a}})Touvron, Lavril, Izacard, Martinet, Lachaux, Lacroix, Rozi{\`{e}}re, Goyal, Hambro, Azhar, Rodriguez, Joulin, Grave, and Lample}]{DBLP:journals/corr/abs-2302-13971}
Hugo Touvron, Thibaut Lavril, Gautier Izacard, Xavier Martinet, Marie{-}Anne Lachaux, Timoth{\'{e}}e Lacroix, Baptiste Rozi{\`{e}}re, Naman Goyal, Eric Hambro, Faisal Azhar, Aur{\'{e}}lien Rodriguez, Armand Joulin, Edouard Grave, and Guillaume Lample. 2023{\natexlab{a}}.
\newblock \href {https://doi.org/10.48550/ARXIV.2302.13971} {Llama: Open and efficient foundation language models}.
\newblock \emph{CoRR}, abs/2302.13971.

\bibitem[{Touvron et~al.(2023{\natexlab{b}})Touvron, Martin, Stone, Albert, Almahairi, Babaei, Bashlykov, Batra, Bhargava, Bhosale, Bikel, Blecher, Canton{-}Ferrer, Chen, Cucurull, Esiobu, Fernandes, Fu, Fu, Fuller, Gao, Goswami, Goyal, Hartshorn, Hosseini, Hou, Inan, Kardas, Kerkez, Khabsa, Kloumann, Korenev, Koura, Lachaux, Lavril, Lee, Liskovich, Lu, Mao, Martinet, Mihaylov, Mishra, Molybog, Nie, Poulton, Reizenstein, Rungta, Saladi, Schelten, Silva, Smith, Subramanian, Tan, Tang, Taylor, Williams, Kuan, Xu, Yan, Zarov, Zhang, Fan, Kambadur, Narang, Rodriguez, Stojnic, Edunov, and Scialom}]{DBLP:journals/corr/abs-2307-09288}
Hugo Touvron, Louis Martin, Kevin Stone, Peter Albert, Amjad Almahairi, Yasmine Babaei, Nikolay Bashlykov, Soumya Batra, Prajjwal Bhargava, Shruti Bhosale, Dan Bikel, Lukas Blecher, Cristian Canton{-}Ferrer, Moya Chen, Guillem Cucurull, David Esiobu, Jude Fernandes, Jeremy Fu, Wenyin Fu, Brian Fuller, Cynthia Gao, Vedanuj Goswami, Naman Goyal, Anthony Hartshorn, Saghar Hosseini, Rui Hou, Hakan Inan, Marcin Kardas, Viktor Kerkez, Madian Khabsa, Isabel Kloumann, Artem Korenev, Punit~Singh Koura, Marie{-}Anne Lachaux, Thibaut Lavril, Jenya Lee, Diana Liskovich, Yinghai Lu, Yuning Mao, Xavier Martinet, Todor Mihaylov, Pushkar Mishra, Igor Molybog, Yixin Nie, Andrew Poulton, Jeremy Reizenstein, Rashi Rungta, Kalyan Saladi, Alan Schelten, Ruan Silva, Eric~Michael Smith, Ranjan Subramanian, Xiaoqing~Ellen Tan, Binh Tang, Ross Taylor, Adina Williams, Jian~Xiang Kuan, Puxin Xu, Zheng Yan, Iliyan Zarov, Yuchen Zhang, Angela Fan, Melanie Kambadur, Sharan Narang, Aur{\'{e}}lien Rodriguez, Robert Stojnic, Sergey Edunov,
  and Thomas Scialom. 2023{\natexlab{b}}.
\newblock \href {https://doi.org/10.48550/ARXIV.2307.09288} {Llama 2: Open foundation and fine-tuned chat models}.
\newblock \emph{CoRR}, abs/2307.09288.

\bibitem[{Wang et~al.(2024)Wang, Bai, Tan, Wang, Fan, Bai, Chen, Liu, Wang, Ge, Fan, Dang, Du, Ren, Men, Liu, Zhou, Zhou, and Lin}]{Qwen2VL}
Peng Wang, Shuai Bai, Sinan Tan, Shijie Wang, Zhihao Fan, Jinze Bai, Keqin Chen, Xuejing Liu, Jialin Wang, Wenbin Ge, Yang Fan, Kai Dang, Mengfei Du, Xuancheng Ren, Rui Men, Dayiheng Liu, Chang Zhou, Jingren Zhou, and Junyang Lin. 2024.
\newblock Qwen2-vl: Enhancing vision-language model's perception of the world at any resolution.
\newblock \emph{arXiv preprint arXiv:2409.12191}.

\bibitem[{Wang et~al.(2023)Wang, Lv, Yu, Hong, Qi, Wang, Ji, Yang, Zhao, Song, Xu, Xu, Li, Dong, Ding, and Tang}]{wang2023cogvlm}
Weihan Wang, Qingsong Lv, Wenmeng Yu, Wenyi Hong, Ji~Qi, Yan Wang, Junhui Ji, Zhuoyi Yang, Lei Zhao, Xixuan Song, Jiazheng Xu, Bin Xu, Juanzi Li, Yuxiao Dong, Ming Ding, and Jie Tang. 2023.
\newblock \href {http://arxiv.org/abs/2311.03079} {Cogvlm: Visual expert for pretrained language models}.

\bibitem[{Wei et~al.(2022)Wei, Wang, Schuurmans, Bosma, Ichter, Xia, Chi, Le, and Zhou}]{DBLP:conf/nips/Wei0SBIXCLZ22}
Jason Wei, Xuezhi Wang, Dale Schuurmans, Maarten Bosma, Brian Ichter, Fei Xia, Ed~H. Chi, Quoc~V. Le, and Denny Zhou. 2022.
\newblock \href {http://papers.nips.cc/paper\_files/paper/2022/hash/9d5609613524ecf4f15af0f7b31abca4-Abstract-Conference.html} {Chain-of-thought prompting elicits reasoning in large language models}.
\newblock In \emph{Advances in Neural Information Processing Systems 35: Annual Conference on Neural Information Processing Systems 2022, NeurIPS 2022, New Orleans, LA, USA, November 28 - December 9, 2022}.

\bibitem[{Xie et~al.(2021)Xie, Li, and Pu}]{DBLP:conf/acl/XieLP20}
Yubo Xie, Junze Li, and Pearl Pu. 2021.
\newblock \href {https://doi.org/10.18653/V1/2021.ACL-SHORT.6} {Uncertainty and surprisal jointly deliver the punchline: Exploiting incongruity-based features for humor recognition}.
\newblock In \emph{Proceedings of the 59th Annual Meeting of the Association for Computational Linguistics and the 11th International Joint Conference on Natural Language Processing, {ACL/IJCNLP} 2021, (Volume 2: Short Papers), Virtual Event, August 1-6, 2021}, pages 33--39. Association for Computational Linguistics.

\bibitem[{Yang et~al.(2024{\natexlab{a}})Yang, Yang, Hui, Zheng, Yu, Zhou, Li, Li, Liu, Huang, Dong, Wei, Lin, Tang, Wang, Yang, Tu, Zhang, Ma, Xu, Zhou, Bai, He, Lin, Dang, Lu, Chen, Yang, Li, Xue, Ni, Zhang, Wang, Peng, Men, Gao, Lin, Wang, Bai, Tan, Zhu, Li, Liu, Ge, Deng, Zhou, Ren, Zhang, Wei, Ren, Fan, Yao, Zhang, Wan, Chu, Liu, Cui, Zhang, and Fan}]{qwen2}
An~Yang, Baosong Yang, Binyuan Hui, Bo~Zheng, Bowen Yu, Chang Zhou, Chengpeng Li, Chengyuan Li, Dayiheng Liu, Fei Huang, Guanting Dong, Haoran Wei, Huan Lin, Jialong Tang, Jialin Wang, Jian Yang, Jianhong Tu, Jianwei Zhang, Jianxin Ma, Jin Xu, Jingren Zhou, Jinze Bai, Jinzheng He, Junyang Lin, Kai Dang, Keming Lu, Keqin Chen, Kexin Yang, Mei Li, Mingfeng Xue, Na~Ni, Pei Zhang, Peng Wang, Ru~Peng, Rui Men, Ruize Gao, Runji Lin, Shijie Wang, Shuai Bai, Sinan Tan, Tianhang Zhu, Tianhao Li, Tianyu Liu, Wenbin Ge, Xiaodong Deng, Xiaohuan Zhou, Xingzhang Ren, Xinyu Zhang, Xipin Wei, Xuancheng Ren, Yang Fan, Yang Yao, Yichang Zhang, Yu~Wan, Yunfei Chu, Yuqiong Liu, Zeyu Cui, Zhenru Zhang, and Zhihao Fan. 2024{\natexlab{a}}.
\newblock Qwen2 technical report.
\newblock \emph{arXiv preprint arXiv:2407.10671}.

\bibitem[{Yang et~al.(2024{\natexlab{b}})Yang, Yang, Hui, Zheng, Yu, Zhou, Li, Li, Liu, Huang, Dong, Wei, Lin, Tang, Wang, Yang, Tu, Zhang, Ma, Yang, Xu, Zhou, Bai, He, Lin, Dang, Lu, Chen, Yang, Li, Xue, Ni, Zhang, Wang, Peng, Men, Gao, Lin, Wang, Bai, Tan, Zhu, Li, Liu, Ge, Deng, Zhou, Ren, Zhang, Wei, Ren, Liu, Fan, Yao, Zhang, Wan, Chu, Liu, Cui, Zhang, Guo, and Fan}]{DBLP:journals/corr/abs-2407-10671}
An~Yang, Baosong Yang, Binyuan Hui, Bo~Zheng, Bowen Yu, Chang Zhou, Chengpeng Li, Chengyuan Li, Dayiheng Liu, Fei Huang, Guanting Dong, Haoran Wei, Huan Lin, Jialong Tang, Jialin Wang, Jian Yang, Jianhong Tu, Jianwei Zhang, Jianxin Ma, Jianxin Yang, Jin Xu, Jingren Zhou, Jinze Bai, Jinzheng He, Junyang Lin, Kai Dang, Keming Lu, Keqin Chen, Kexin Yang, Mei Li, Mingfeng Xue, Na~Ni, Pei Zhang, Peng Wang, Ru~Peng, Rui Men, Ruize Gao, Runji Lin, Shijie Wang, Shuai Bai, Sinan Tan, Tianhang Zhu, Tianhao Li, Tianyu Liu, Wenbin Ge, Xiaodong Deng, Xiaohuan Zhou, Xingzhang Ren, Xinyu Zhang, Xipin Wei, Xuancheng Ren, Xuejing Liu, Yang Fan, Yang Yao, Yichang Zhang, Yu~Wan, Yunfei Chu, Yuqiong Liu, Zeyu Cui, Zhenru Zhang, Zhifang Guo, and Zhihao Fan. 2024{\natexlab{b}}.
\newblock \href {https://doi.org/10.48550/ARXIV.2407.10671} {Qwen2 technical report}.
\newblock \emph{CoRR}, abs/2407.10671.

\bibitem[{Yang et~al.(2024{\natexlab{c}})Yang, Yang, Zhang, Hui, Zheng, Yu, Li, Liu, Huang, Wei et~al.}]{qwen2.5}
An~Yang, Baosong Yang, Beichen Zhang, Binyuan Hui, Bo~Zheng, Bowen Yu, Chengyuan Li, Dayiheng Liu, Fei Huang, Haoran Wei, et~al. 2024{\natexlab{c}}.
\newblock Qwen2. 5 technical report.
\newblock \emph{arXiv preprint arXiv:2412.15115}.

\bibitem[{Yang et~al.(2024{\natexlab{d}})Yang, Li, Dong, Xia, and Sui}]{DBLP:journals/corr/abs-2402-11281}
Yixin Yang, Zheng Li, Qingxiu Dong, Heming Xia, and Zhifang Sui. 2024{\natexlab{d}}.
\newblock \href {https://doi.org/10.48550/ARXIV.2402.11281} {Can large multimodal models uncover deep semantics behind images?}
\newblock \emph{CoRR}, abs/2402.11281.

\bibitem[{Yao et~al.(2024)Yao, Yu, Zhang, Wang, Cui, Zhu, Cai, Li, Zhao, He et~al.}]{yao2024minicpm}
Yuan Yao, Tianyu Yu, Ao~Zhang, Chongyi Wang, Junbo Cui, Hongji Zhu, Tianchi Cai, Haoyu Li, Weilin Zhao, Zhihui He, et~al. 2024.
\newblock Minicpm-v: A gpt-4v level mllm on your phone.
\newblock \emph{arXiv preprint arXiv:2408.01800}.

\bibitem[{Zhai et~al.(2023)Zhai, Mustafa, Kolesnikov, and Beyer}]{DBLP:conf/iccv/ZhaiM0B23}
Xiaohua Zhai, Basil Mustafa, Alexander Kolesnikov, and Lucas Beyer. 2023.
\newblock Sigmoid loss for language image pre-training.
\newblock In \emph{Proceedings of the IEEE/CVF international conference on computer vision}, pages 11975--11986.

\end{thebibliography}
